# Mimesis, Poiesis, and Imagination: Exploring Text-to-Image Generation of Biblical Narratives

Willem Th. van Peursen and Samuel E. Entsua-Mensah

Eep Talstra Centre for Bible and Computer (ETCBC), School for Religion and Theology, Vrije Universiteit Amsterdam
w.t.van.peursen@vu.nl; s.e.entsua-mensah@student.vu.nl

**Abstract** This study explores the intersection of artificial intelligence and the visualization of Biblical narratives by analyzing AI-generated images of Exodus 2:5–9 (Moses found in River Nile) using MidJourney. Drawing on the classical concepts of *mimesis* (imitation) and *poiesis* (creative generation), the authors investigate how text-to-image (T2I) models reproduce or reimagine sacred narratives. Through comparative visual analysis—including Google image results and classical paintings—the research evaluates the stylistic, theological, and cultural dimensions of AI-generated depictions. Findings show that while AI excels in producing aesthetically rich and imaginative visuals, it also reflects the biases and limitations of its training data. The study highlights AI's potential to augment human imagination but questions its capacity for genuine creativity, authorial intent, and theological depth. It concludes by suggesting that AI can serve as a creative partner in reinterpreting biblical texts, though its role in sacred art remains complex and contested.

# 1. Introduction

The capabilities of Text-to-Image (T2I) models are advancing rapidly, enabling the generation of highly accurate visuals from textual descriptions. For instance, when a short story is input into a model like Midjourney, the resulting images often achieve remarkable fidelity to the narrative. However, this progress brings a potential trade-off: the generated images can become increasingly predictable. For example, using a biblical story as a prompt frequently produces results resembling widely available depictions of the same story found online. This phenomenon is unsurprising given that Midjourney trains on vast datasets scraped from the internet.

| | | |
|---|---|---|
| 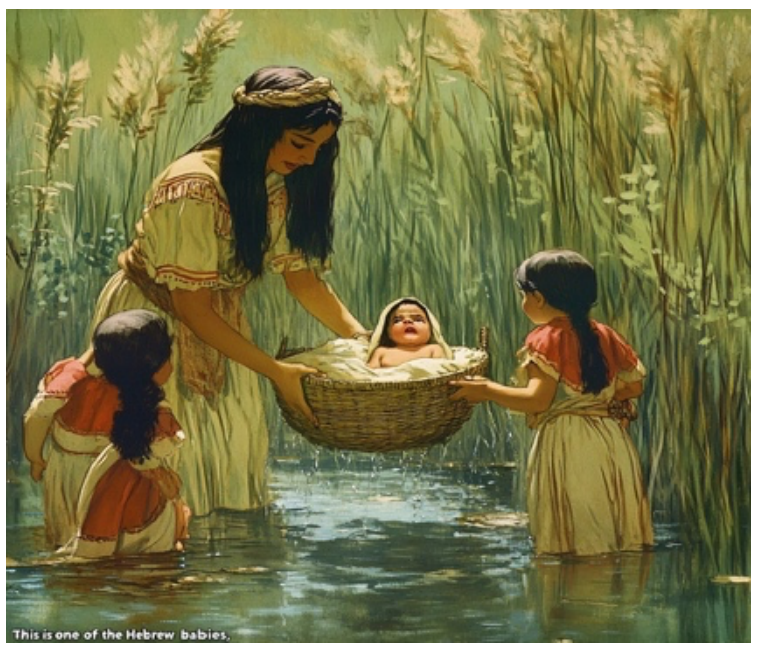<br> | | 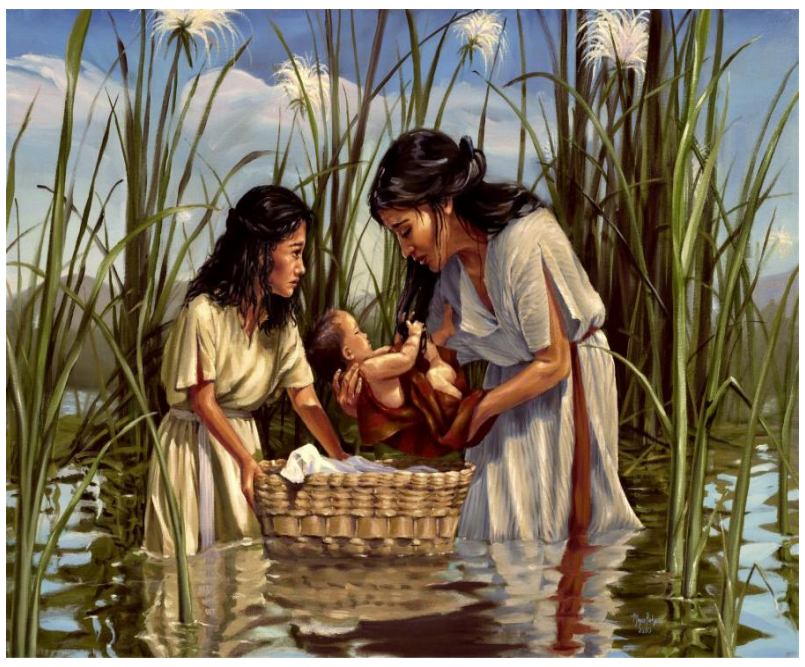 |
| Prompt: KJV | | https://www.womeninthescriptures.com/2015/08/the-women-who-delivered-moses.html |

What are the implications of such tendencies in T2I models? How does it affect their potential to reinforce visual stereotypes or, conversely, generate novel artistic interpretations? Starting from the distinction between *mimesis* (imitation) and *poiesis* (creation), we can ask: Do T2I models possess the creative potential for *poiesis*, or are they increasingly confined to *mimesis*? The use of the terms *mimesis* and *poesis* can be traced back to Aristotle, who considered them complementary, we rather use them here in the way in which they have been framed by Romantic authors such as Friedrich Schiller (1759–1805)[1] see Samuel Taylor Coleridge (1772–1834),[2] and Percy Bysshe Shelley (1792–1822),[3] who distinguish between mechanical imitation,[4] and organic or poetic creativity, in which art finds its true destination.[5]

This question is related, on the one hand, to the issue of the potential and limitations of computational creativity, and, on the other hand, to the question how we can define creativity informed by AI tools.

As to the question how to define creation and creativity, let's just quote one of the many definitions one finds in the dictionaries. According to *Collins Cobuild English Language* Dictionary, creativity is:

> The ability to invent and develop new and original ideas, especially in an artistic way.[6]

Or, in a more recent online version of Collins dictionary:

> The ability to transcend traditional ideas, rules, patterns, relationships, or the like, and to create meaningful new ideas, forms, methods, interpretations, etc.; originality, progressiveness, or imagination.[7]

These definitions agree to a large extent with what people may intuitively think of when talking about creativity, but a scholarly definition of what is creativity is much more complex.[8] Defining creativity is an interdisciplinary endeavour, involving psychological, philosophical and theological disciplines.[9] In the Western tradition the concept of creativity is closely related to the concept of divine inspiration and the belief in God as the Creator.[10] The question as to the relation between creativity and creator is relevant to our research because it has been claimed that in the case of computational creativity the notion of authorial intent or an 'experienced creator' is hard to maintain.

---

[1] Especially his letters *Über die ästhetische Erziehung des Menschen*.
[2] See especially his *Biographia Literaria*.
[3] In his essay *Defence of Poetry*.
[4] Cf. Schopenhauer, *Welt als Wille und Vorstellung*, 268–269 on "Nachahmung der Natur".
[5] Since our main focus is on the distinction between imitation and creativity we will not elaborate on the further development of these concepts in the works of Heidegger or Gadamer, although also with these authors we find a positive evaluation of "poiesis" or "creativity" as against a more pejorative use of "imitation".
[6] Collins Cobuild English Language Dictionary, 331b.
[7] https://www.collinsdictionary.com/dictionary/english/creativity
[8] See, e.g., Brown, "Creativity. What Are We to Measure?"; Mumford, "Where Have We Been, Where Are We Going?"
[9] On Margaret Boden's distinction between combinational, exploratory and transformational creativity, see below, section 4.2.1.
[10] Niu, Weihua; Sternberg, Robert J. (2006). "The Philosophical Roots of Western and Eastern Conceptions of Creativity" (PDF). Journal of Theoretical and Philosophical Psychology. 26 (1–2): 18–38. doi:10.1037/h0091265.

David Holz, the founder of MidJourney, defined the role of T2I not so much in terms of creativity or art, but rather in terms of imagination:

> It's important that we don't think of this as an AI 'artist.' We think of it more like using AI to augment our imagination. It's not necessarily about art but about imagining. We are asking, 'what if.' The AI sort of increases the power of our imagination.[11]

The idea of technology augmenting the imagination is even more interesting when we recall that in the past the appearance of "new media" such as TV and cinema was sometimes described as a threat to imagination and raised a concern that they might diminish the imagination.[12] In the case of AI, on the one hand computational creativity seems to have the potential of the generation of an endless number of images with all kinds of variations, but on the other hand the homogenization effects of AI tools seem somehow to limit the creative potential,[13] Anna Caterina Dalmasso and Sofia Pirandello argue that "augmented imagination" should be defined as a process where technology does not just extend existing capabilities but also reconfigures imagination itself through the environment.[14]

The distinction between creativity and imagination is important, but it should be noted that the two concepts are interrelated and that imagination, too, in many of its definitions goes beyond merely imitating or mirroring. Thus we find the following definition in Collins Cobuild dictionary:

> The ability that you have to think of and form pictures of ideas in your mind of things that are different, interesting, or exciting.[15]

This also means that questions about *mimesis* and *poiesis*, about human and artificial creativity, and about imagination are all related. In response to David Holz' remark quoted above, we can ask the question as to whether T2I technology can truly fulfill the aspiration to "augment our imagination".

To explore these questions, we examined the story of Moses being found in the Nile (Exodus 2:5–9), employing diverse prompts to generate images.[16] We compared these outputs to historical artistic renditions of the same scene as well as modern images found on the internet.

## 2. Methodology

This study employs a comparative visual analysis approach to examine AI-generated biblical imagery with classical paintings and Google search images. The research focuses on representations of Moses being found, evaluating how different sources—MidJourney-generated images, classical paintings, and images retrieved from Google—depict the biblical narrative. The study explores mimesis (imitation of artistic traditions) and poiesis (creative reinterpretation) to assess how AI expands, distorts, or enhances traditional biblical storytelling.

[11] See https://www.openbible.info/blog/2022/08/exploring-ai-assisted-bible-study/ and https://archive.ph/dGoP3.
[12] Dalmasso and Pirandello. "Augmented Imagination: Thinking Technology Beyond Extension", 34.
[13] Cf. Kreminski, "Endless Forms Most Similar", 1–2.
[14] Dalmasso and Pirandello. "Augmented Imagination: Thinking Technology Beyond Extension", 34.
[15] Collins Cobuild English Language Dictionary, 723a.
[16] Most of our prompts were the plain text of the Bible. It should be noted however, that prompt engineering is a rapidly evolving field of research that deserves further exploration. See, e.g., Liu and Chilton, "Design Guidelines for Prompt Engineering Text-to-Image Generative Models".

## 2.1 Data collection

Currently, various T2I generative AI models are available. In the current project we focus on MidJourney. In a previous project we experimented with DALL-E, Midjourney and Stable Diffusion.[17] In that project Midjourney performed best in terms of accuracy. MidJourney, like other T2I generation platforms based on AI, is developed on the foundation of a large dataset comprising publicly available images, licensed material, and sets collected from online archives. Through this, MidJourney can create images that convey a broad range of artistic styles, from classical paintings to contemporary digital artwork and photographic renderings. In contrast to the conventional datasets such as Art and the Bible,[18] Iconclass,[19] or the online archive of the Netherlands Institute for Art History,[20] the precise data used by Midjourney is under wraps, pointing to the secretive and proprietary nature of its training. This is in sharp contrast with other T2I models like DALL.E, which draws upon massive collections of text-image pairs, and Stable Diffusion, which trains on the publicly available LAION-5B dataset comprising over 5 billion image-text pairings.[21]

A significant distinction between MidJourney and other models is its emphasis on artistic stylization and creativity. While DALL.E is known for producing realistic images and Stable Diffusion offers customizable outputs through its open-source nature, MidJourney excels in creating visually rich, dreamlike images that often have a romanticized or fantastical quality.

Moreover, MidJourney's approach to image generation emphasizes creativity that goes beyond simple replication. Its ability to integrate stylistic elements from different cultural and artistic traditions allows for a more expansive and imaginative visual output, which is not just a reflection of pre-existing images but an augmentation of visual storytelling itself. This aligns with the concept of augmented reality, where technology reshapes how imagination is externalized and interacted with in real-time.[22]

## 2.2 Data analysis

The collected images were examined through qualitative visual inspection and comparative evaluation. AI-generated images were analyzed alongside historical paintings and Google search results to assess their stylistic patterns, accuracy, and artistic effectiveness. A key focus was identifying recurring stylistic patterns in AI-generated outputs and how they are compared to traditional biblical paintings.

## 2.3 Analytical framework

The study applies a structured comparative visual analysis based on the following criteria:

---

[17] Makimei, Wang and Van Peursen, "Seeing The Words: Evaluating AI-generated Biblical Art".
[18] https://www.artbible.info/
[19] https://iconclass.org/
[20] https://www.rkd.nl/
[21] University of Michigan. Painting or Scraping? MidJourney and Training AI to Make Art" https://sites.lsa.umich.edu/qmss/2024/04/03/painting-or-scraping-midjourney-and-training-ai-to-make-art.
[22] Dalmasso and Pirandello. "Augmented Imagination: Thinking Technology Beyond Extension".

- Mimesis (Imitation & Accuracy): Examining how closely AI-generated images, Google search results, and historical paintings replicate the biblical scene, particularly in terms of composition, setting, and cultural representation.
- Poiesis (Creativity & Reinterpretation): Analyzing the creative liberties taken by AI in reimagining the scene, assessing whether AI-generated outputs introduce new artistic perspectives beyond conventional depictions.
- Composition & Stylistic Elements: Comparing color palettes, lighting, realism, and symbolic elements across different sources to evaluate the artistic coherence of AI-generated works versus traditional art.
- Cultural & Theological Representation: Investigating how different sources portray race, ethnicity, clothing, and religious symbols, assessing whether AI-generated images challenge or reinforce traditional iconography.

## 2.4 Visual Elements in Exodus 2:5–9

A comparison of Exodus 2:5–9 across the KJV, NIV, and NRSV highlights key recurring motifs central to Moses' discovery. Using Erich Auerbach's mimesis framework, this section identifies the passage's key visual elements and their narrative significance.

- Pharaoh's Daughter at the River: The act of a royal figure descending to bathe suggests movement, luxury, and a contrast between the Egyptian court and nature.
- The Basket in the Reeds: A small, fragile vessel hidden among reeds symbolizes both concealment and divine providence.
- Maidens/Attendants: Their presence adds courtly grandeur, contrasting with the child's vulnerability.
- Opening the Basket: A dramatic, intimate act, revealing the child's helplessness.
- The Crying Child: His weeping humanizes the scene, evoking compassion and emotional intensity.
- Dialogue & Interaction: The sister's sudden appearance shifts the focus to human agency, emphasizing familial ties.
- The River & Reeds as Setting: Symbolizing life, danger, and escape, the Nile reinforces the story's dramatic stakes.

The KJV retains the term "ark," which, at least for contemporary readers, evokes parallels with Noah's Ark and reinforces themes of divine protection and salvation,[23] while the NIV and NRSV modernize the language, replacing it with "basket," emphasizing the object's function rather than its symbolic weight. Additionally, NIV and NRSV introduce emotional nuance, such as "she felt sorry for him," making Pharaoh's daughter's compassion more explicit compared to the KJV's straightforward "she took pity on him." Variations in terminology, such as "maidens" (KJV) vs. "attendants" (NIV/NRSV), subtly influence the perception of Pharaoh's daughter's social standing and the formal structure of her entourage, shaping the scene's cultural and hierarchical undertones.

[23] Compare the definitions given in https://www.collinsdictionary.com/dictionary/english/ark.

# 3. Data

## 3.1 Images generated with Midjourney

Throughout the process of generating AI-based visual interpretations of Exodus 2:5–9, a total of 417 images were created using MidJourney. These images represent diverse artistic approaches, reflecting different Bible translations, artistic traditions, and creative prompt variations. Given the scope and depth of this project, it was necessary to curate a representative subset of the generated images to ensure a balanced and meaningful presentation. As a result, 50 images were carefully selected for inclusion in this table. This curated set not only illustrates AI's ability to visualize sacred narratives but also highlights the dynamic interplay between mimesis and poiesis in the generation of Biblical imagery. The selected images can be found in Appendix 1.

## 3.2 Random images from Google Search

Appendix 2 contains a list of images gathered through Google search. While these images contained a mix of ancient paintings, modern Christian art, children's illustrations, and cinematic adaptations, they followed a pattern similar to those generated by MidJourney.

## 3.3 Works from art history

As a starting point for the comparison with AI-generated images, we focused on pictures from the Renaissance and Baroque periods because of their naturalistic style, which replaced the more symbolic and static style of the Middle Ages.[24] The following paintings were included in the comparison:

1. Moses is found by the daughter of Pharaoh; Moses throws Pharaoh's crown to the ground, dated between 1190-1200 anonymously.
2. Moses in his basket is found by the daughter of Pharaoh dated 1430–45 by Alexander Master (draughtsman).[25]
3. Finding of Moses dated between 1560–90 by Jacob de Backer (1550–1600)
4. Moses found by Pharaoh's daughter (Exodus 2:5–6) between 1593–1654 by Pieter Vromans (II) (1578–1654)
5. Moses found by pharaoh's daughter in the rush basket (Exodus 2:5–6) dated 1625 by Moyses van Wtenbrouck (1595–ca. 1647).
6. Moses Abandoned dated between 1628–29 by Nicolas Poussin (1594 - 1665)
7. Moses is found dated 1635 by Rembrandt Harmensz. van Rijn (1606–1669)
8. Moses found by Pharaoh's daughter (Exodus 2:5–6) dated 1640 by Peter Lely (1618-1680)
9. Moses found by Pharaoh's daughter (Exodus 2:5–6), dated 1645, by Hans Bollongier (1600 – 1645).
10. Moses is found by Pharaoh's daughter dated between 1645–80 by Toussaint Gelton (1630–1680)
11. Moses found by the pharao's daughter around 1665 by Jan van Haensbergen (1642–1705)
12. Moses found by the daughter of Pharaoh dated 1705 by Gerard Hoet (I) (1648–1733).

[24] For details see Makimei, Wang and Van Peursen, "Seeing The Words: Evaluating AI-generated Biblical Art", 8–9 (section 4.1).

[25] Alexander Master was active around 1430 in Utrecht as part of the "Bible Masters of the First Generation." His exact birth and death dates are not documented. His contributions are recognized through manuscript studies https://www.mythfolklore.net/2003frametales/weeks/week03/images/alex_virgins.htm

13. The Finding of Moses dated between 1651 by Nicolas Poussin (1594 - 1665)
14. Moses found by Pharaoh's daughter (Exodus 2:5–6) dated 1722 by Adriaen van der Werff (1659–1722) and Pieter van der Werff (1665–1772).

# 4. Evaluation

## 4.1 Visual elements from the biblical text passage

In section 2.4 we identified visual elements in Exodus 2:5–9. We will analyse here how these are represented in the pictures. Note that in most prompts used we included the Biblical text, leaving it to the model to identify the visual elements. In some cases (indicated by "Prompt: visual elements") we made the visual elements explicit.

### 4.1.1 Recurring Themes and Patterns

The elements that appear most frequently across the AI-generated images show which aspects of the narrative AI tends to highlight. The following elements are commonly present:

- A woman (Pharaoh's daughter) appears in most of the images, but her ethnicity and attire vary significantly.
- A boy (Moses) in a basket is consistently depicted but he appears to be in an open basket, in contrast to the prompt text describing Pharaoh's daughter as "opening the basket".
- Water (Nile, reeds, or reflections) is almost universal.

The following elements show high variability:

- The number of maidens fluctuates, with some images aligning with traditional iconography (2–4 attendants), while others depict Pharaoh's daughter alone.
- Moses' age is inconsistent—some images show a newborn, while others depict a toddler.[26]
- Egyptian architectural elements are present in some images but absent in others, revealing MidJourney's selective incorporation of historical details.

### 4.1.2 Elements from the history of interpretation

The short biblical text is compact. The missing details are filled in by tradition. Thus, according to some sources the princess' attendants protected the princess's privacy, and only her maidservant entered the water.[27] However, according to Ezekiel the tragedian, all women went into the water.[28] Josephus describes Pharaoh's daughter as playing on the riverbank rather than bathing, having swimmers retrieve Moses' basket.[29] Islamic traditions mention seven daughters of Pharaoh being present, guided by angelic intervention.[30] Nachmanides specifies that Pharaoh's daughter stood on the lowest step of a staircase into the Nile.[31]

[26] This may be due to the prompts used—Exodus 2:5–9—which does not include the explicit reference to Moses' age (cf. Exodus 2:2)

[27] Cornelis Houtman, *Exodus*, p. 281

[28] Ibid., p. 281, note 24.

[29] Ibid. p. 282

Since these traditions are less represented in existing images of the narrative and less widespread in textual evidence than the short biblical account, it cannot be expected that the AI-generated visuals agree with these traditions. Rather, they alternate between these portrayals, sometimes depicting solitary maidservants aligned with biblical traditions or larger groups, rarely matching the Islamic narrative of seven daughters explicitly. Likewise, some images partially depict Pharao's daughter standing in shallow water (cf. Nachmanides), while others place her directly on the riverbank.

| | | | | |
|---|---|---|---|---|
| 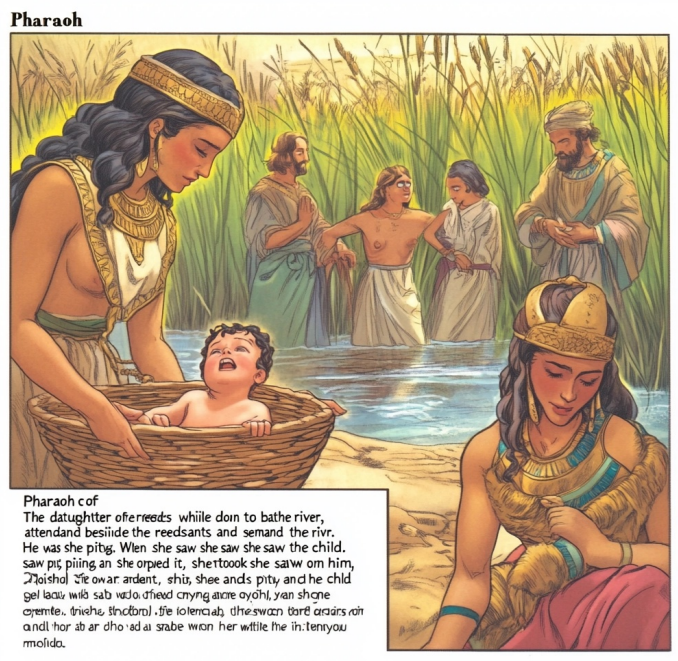  | | 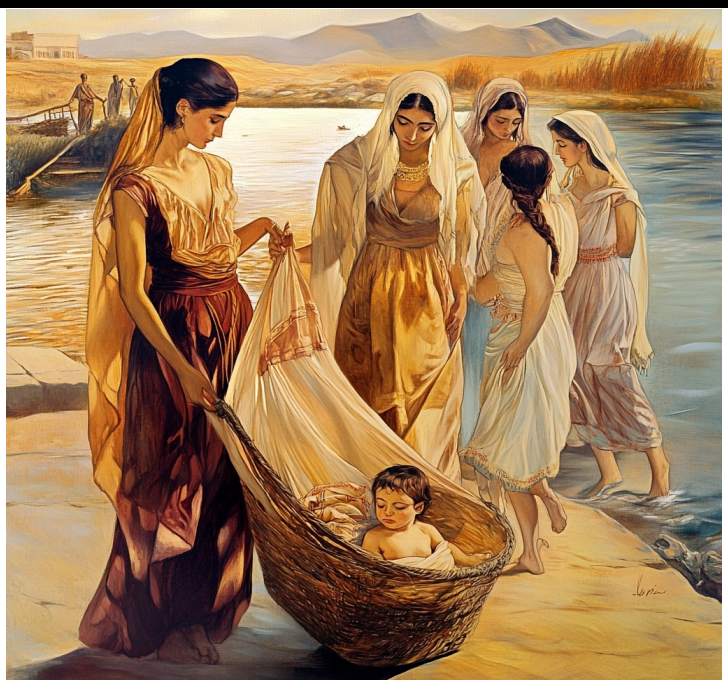 | | 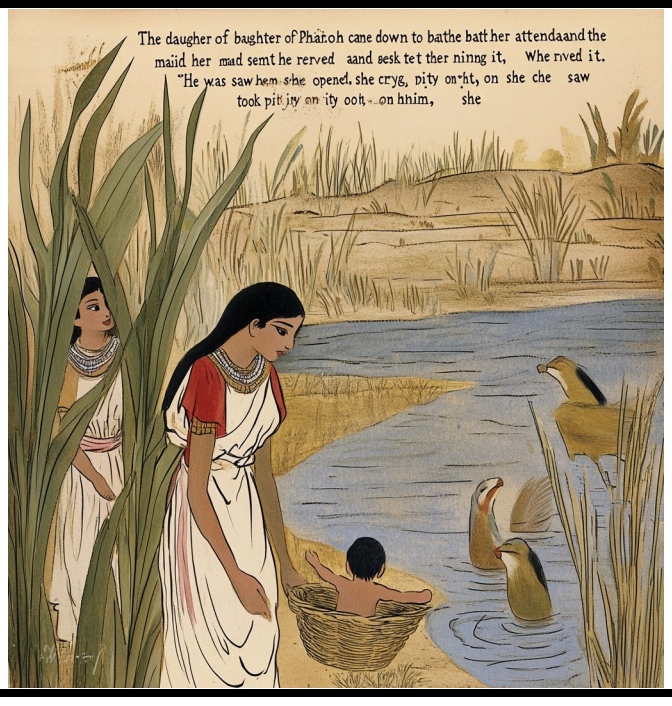 |
| Prompt: NIV | | Prompt: In Dali style | | Prompt: NRSV |
| 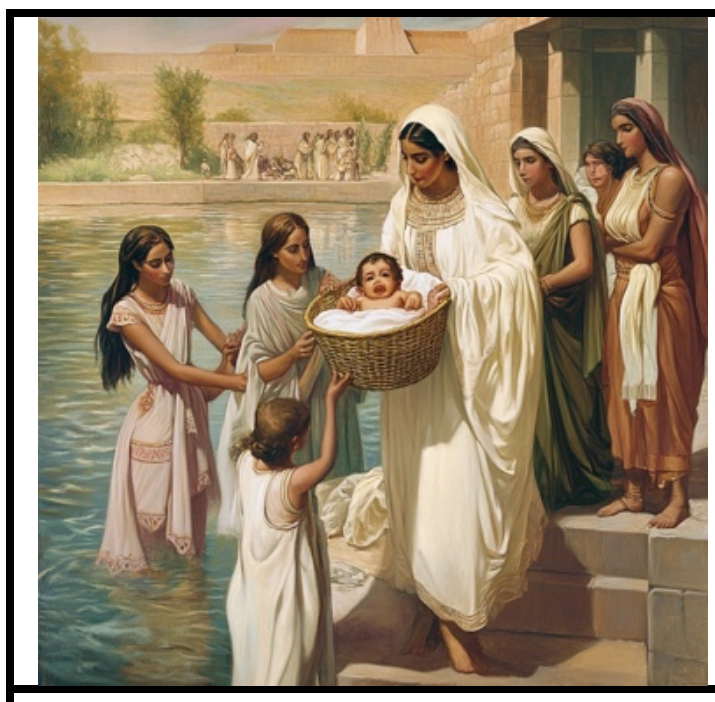 | | 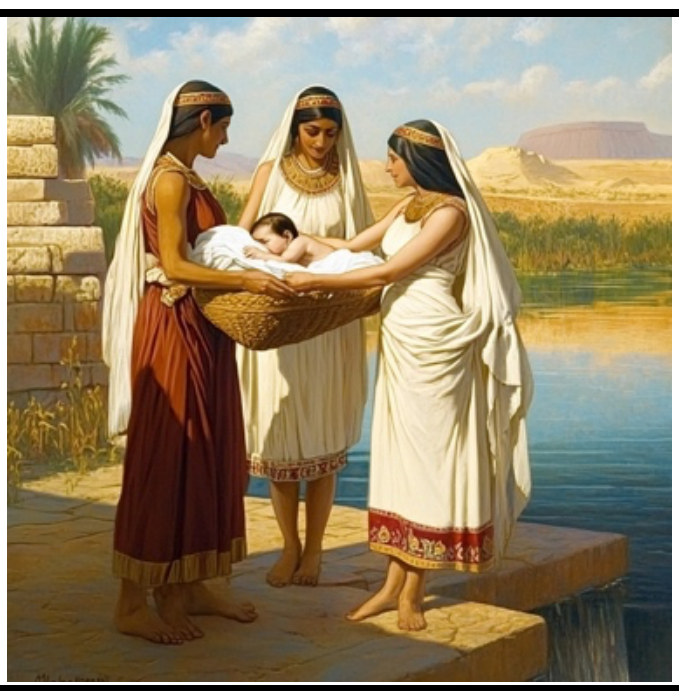 | | 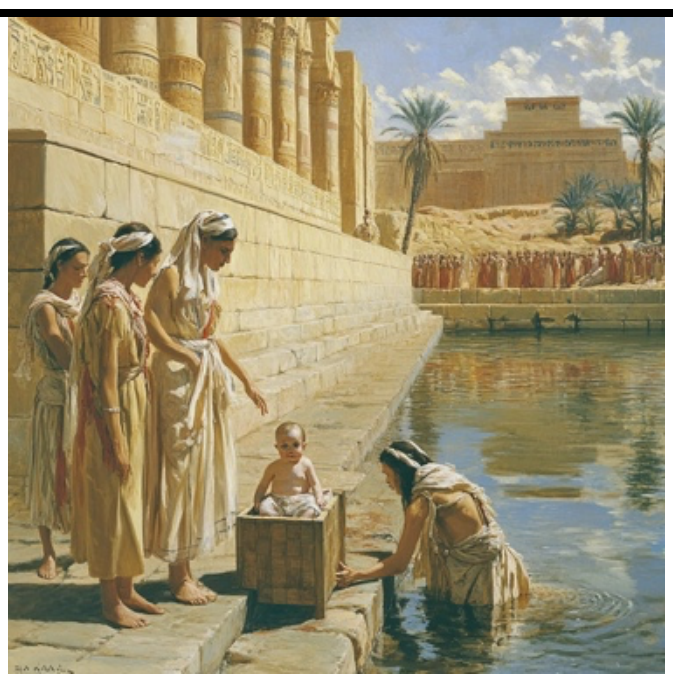 |
| Prompt: in a catholic saint art style | | Prompt:<br>in a Michael Angelo stone art style | | Prompt: KJV |

[30] Ibid. p. 282 citing Josephus, *Antiquities of the Jews* 2.9.5.

[31] Ibid, citing Moses ben Nahman (Nachmanides), *Commentary on the Torah* (Lisbon: 1489)

| | | | | |
|---|---|---|---|---|
| 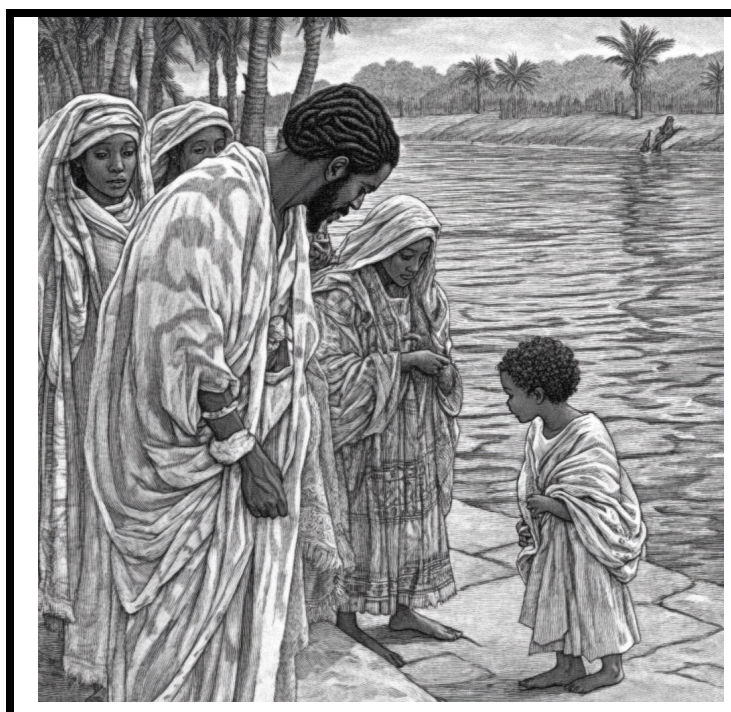 | | 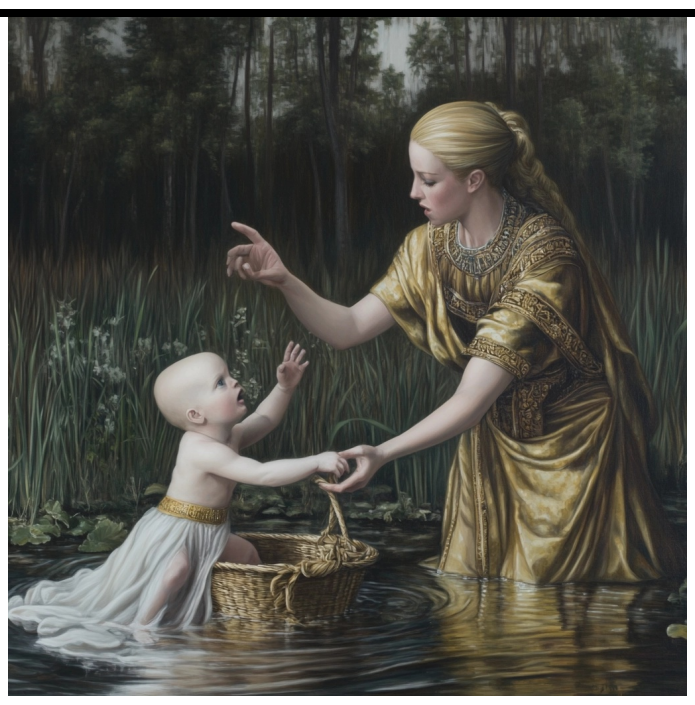 | | 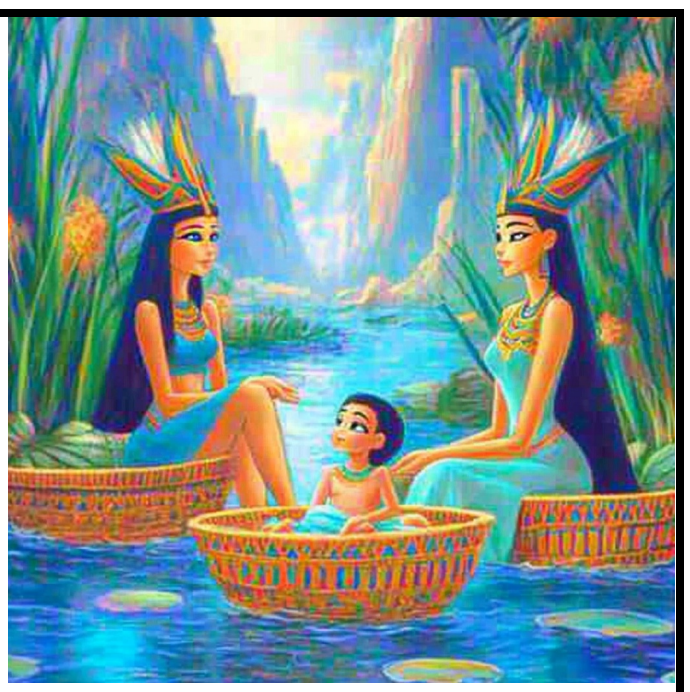 |
| Prompt: From an African point of view | | Prompt: an image of albino baby Moses being found in a basket at the Nile River by the royal daughter of the Pharaoh and her maidens [variety: 40, stylize: 250, weirdness: 2400] | | Prompt: a modern-day image of baby Moses being found in a basket at the Nile River by the Gen Z Disney princess daughter of the pharaoh and her friends. [Variety: 40, stylize: 0, weirdness: 3000] |

## 4.2. Creativity

The preceding section 4.1 largely deals with the accuracy of MidJourney in visualizing elements of the biblical text (*mimesis*). The question we asked in the introduction is whether models such as MidJourney can go beyond reinforcing visual stereotypes and, conversely, generate novel artistic interpretations.

In the context of AI-generated images depicting the Biblical story of Moses being found, *poiesis* refers to the act of creation that transforms textual descriptions into visual representations. Poiesis is not merely imitation but the dynamic process of bringing something new into existence—a synthesis of preexisting elements into a novel, meaningful form. When applied to AI-generated art, it embodies the interplay between human intent, technological capability, and interpretative creativity.

### 4.2.1 The Nature of AI Creativity in Biblical Art

T2I generation concerns the transformation of text to image. The AI takes textual prompts based on biblical descriptions and interprets these words into a visual form. This process involves extracting key elements (e.g., the basket, the Nile River, Pharaoh's attendants) and contextualizing them within specific styles or themes, such as realism, abstraction, or iconography. This process has parallels in the creative process of a human artist who creates an image based on a biblical passage and in the way in which human imagination is triggered by a text to visualize the world of the text.

AI creativity is distinct from traditional artistic creation because it uses algorithms trained on vast art, text, and cultural symbols datasets. This has homogenization effects (cf above, section 1), but the huge datasets on which they are trained also enables the generating of new visual forms based on textual prompts, blending historical styles, cultural motifs, and symbolic elements. AI-generated biblical art frequently merges diverse artistic traditions, drawing inspiration from Renaissance biblical depictions, Baroque dramatism, and modern minimalist aesthetics. For instance, a prompt such as "The finding of Moses in Baroque style" typically produces images featuring chiaroscuro lighting, elaborate costumes, and expressive human figures. In contrast, a "modern interpretation" may favor simplified forms and symbolic abstraction. Notably, prompts that align with established artistic traditions tend to yield more refined outputs compared to experimental styles like "Finding Moses in Karel Appel", as evidenced by the AI-generated images collected in our dataset. This raises an intriguing question: Has MidJourney been predominantly a bias toward these classical styles, perhaps because they are more broadly represented on the Internet?[32]

| | | | | |
|---|---|---|---|---|
| 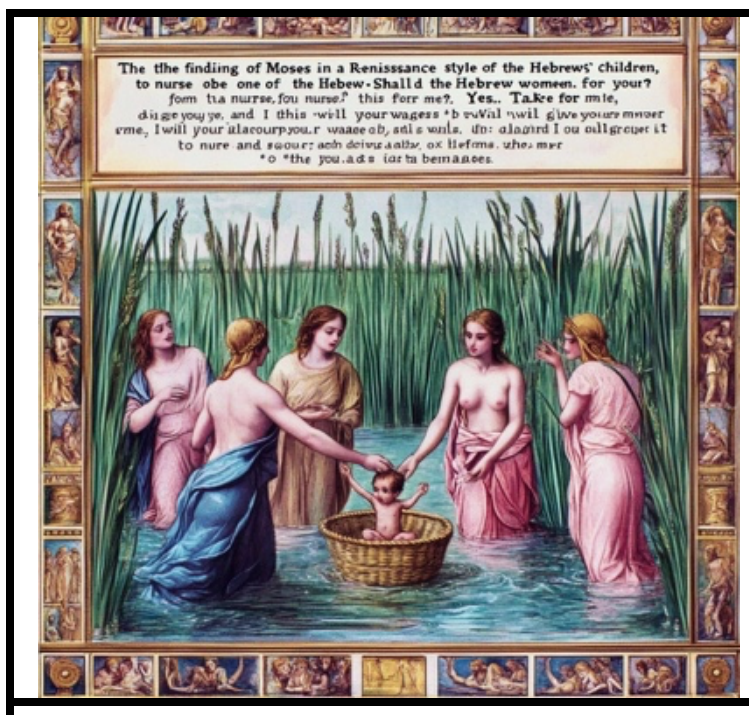  | | 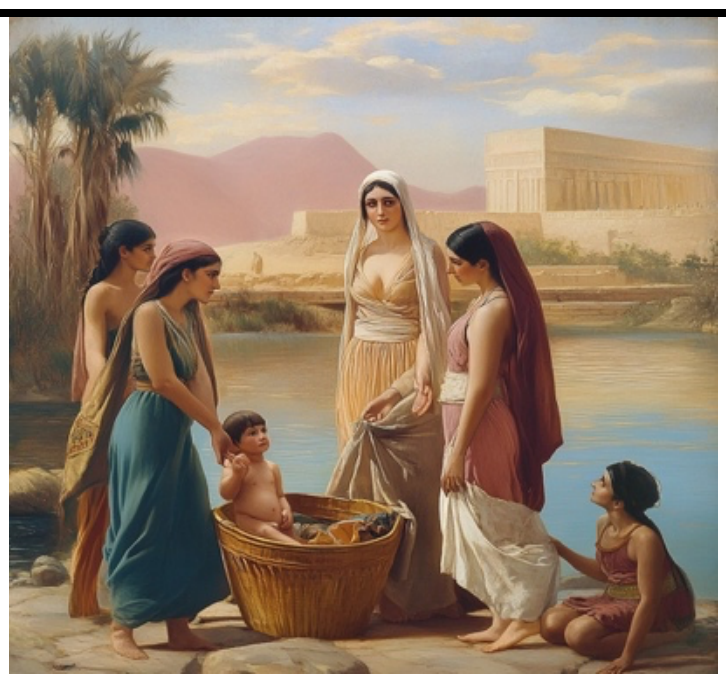 | | 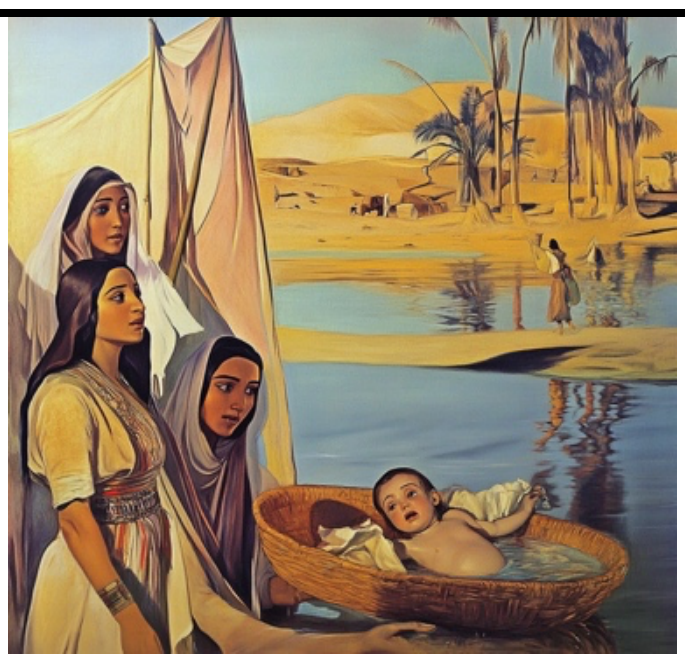 |
| Prompt: the finding of Moses in a Renaissance style" + NRSV | | Prompt: in a Michaelangelo style | | Prompt: in Dali style |

When creativity is defined in terms of novelty, we can point to novel Interpretations beyond the text. AI-generated images are not bound by traditional interpretations, allowing for innovative visualizations.

While there is a general concern that the output of generative AI reflects the biases of their training data, such as the dominance of European artistic traditions and the underrepresentation of non-Western biblical imagery, we noticed in our dataset the AI-generated images demonstrated a degree of diversity, particularly in variations of skin tone, even when such diversity was not explicitly specified in the prompt.

[32] On the complications involved in style transfer in T2I generation, see, e.g., Shadid, Koch and Schneider, "Paint it Black".

| | | | | |
|---|---|---|---|---|
| 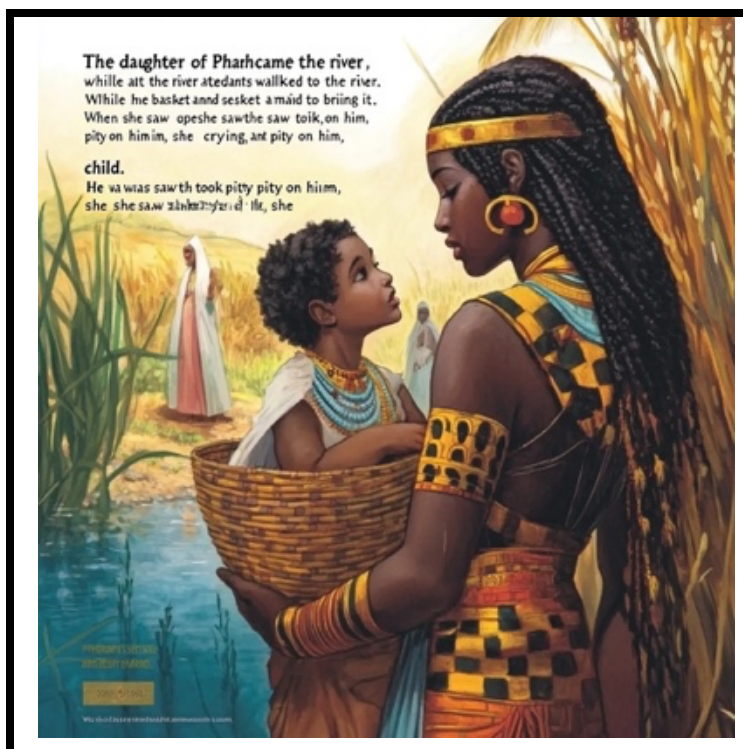 | | 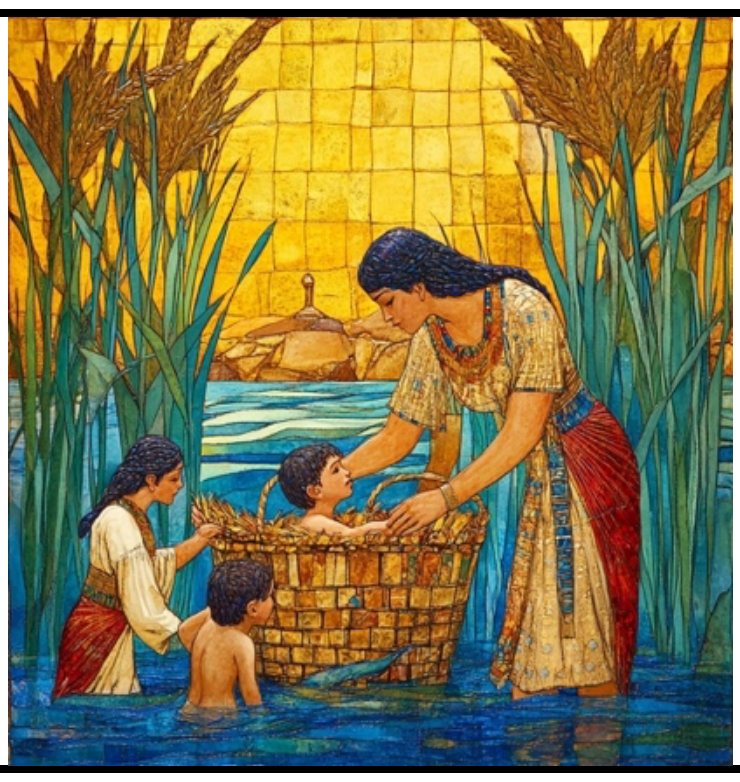 | | 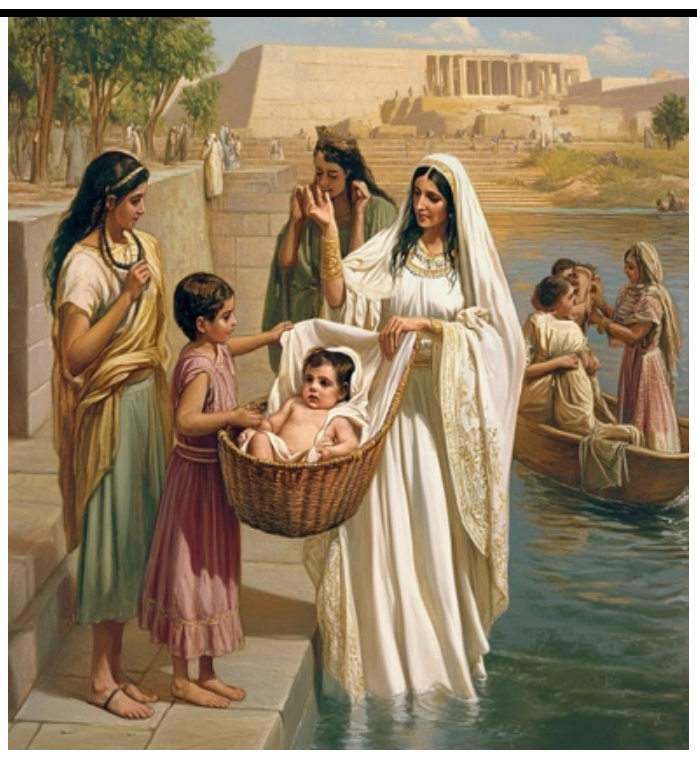 |
| Prompt: NSRV, 3000 scale weirdness, rerun | | Prompt: Generate using the visual elements. | | Prompt: in a catholic saint art style |

An interesting example of MidJourney's creative divergence is the LEGO-based depiction of the finding of Moses, substituting human figures with LEGO minifigures. This occurred through prompts like "generate Lego style images." Although MidJourney typically produces painterly, photorealistic, or stylized images, this result represented a notable deviation.[33] It introduces an interesting intersection between playfulness and religious storytelling. While LEGO-based biblical retellings exist in educational and entertainment contexts, their inclusion in AI-generated religious art suggests a broader cultural influence on AI's outputs. This prompts further reflection on how AI interprets sacred narratives—whether it prioritizes theological depth or defaults to recognizable, contemporary representations found in digital media.

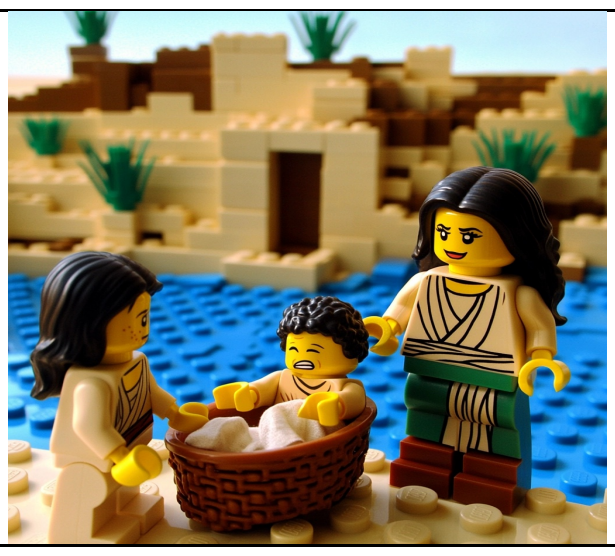

Prompt: using LEGO style

Obviously, a challenging question is when we call certain elements creative and when we call them inaccurate. For example, when Moses is depicted as a young child, rather than a baby (cf. above, section 4.1.1), we may be inclined to consider this inaccurate rather than creative. AI lacks the nuanced theological and cultural understanding that human artists bring to their work. This can result in superficial or inaccurate interpretations (e.g., misplacing the setting or omitting key characters).

[33]A similar observation was made by Makimei, Wang, and Van Peursen, who noted AI-generated images unexpectedly including LEGO figures on a construction site, likely influenced by LEGO-based Bible story videos on YouTube; see Makimei, Wang and Van Peursen, "Seeing The Words: Evaluating AI-generated Biblical Art", 31–32.

| | | |
|---|---|---|
| 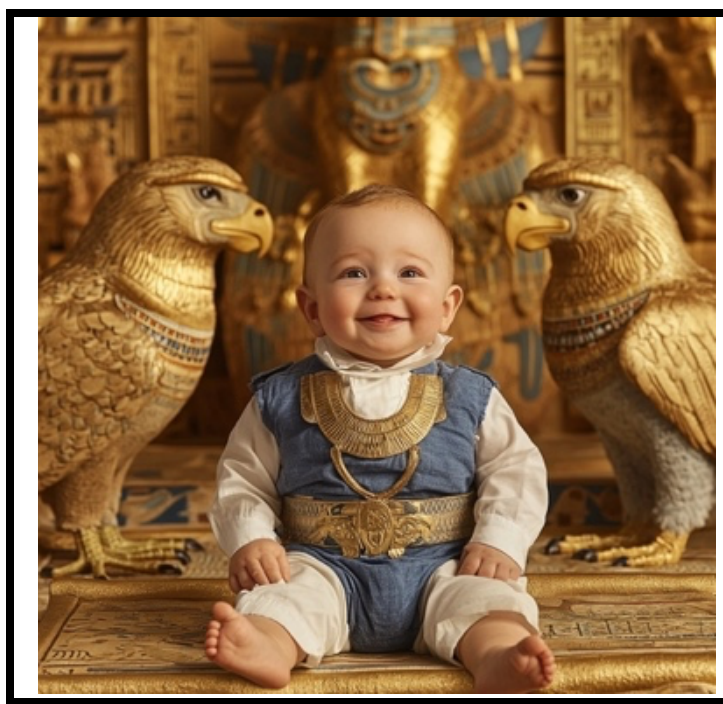 | | 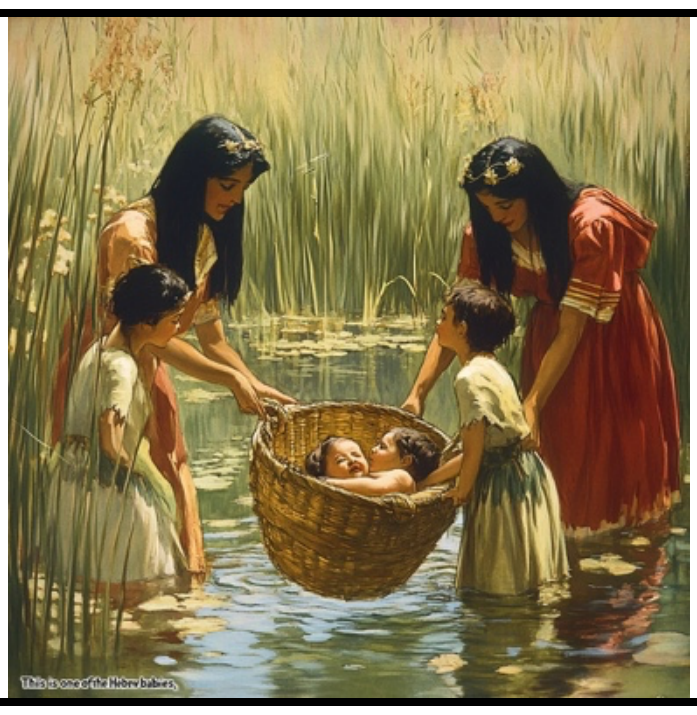 |
| Prompt: realistic images of the child Moses found by the daughter of Pharoah in the river Nile based on the bible story Exodus 2:5–9 | | Prompt: NIV |

| | | | | |
|---|---|---|---|---|
| 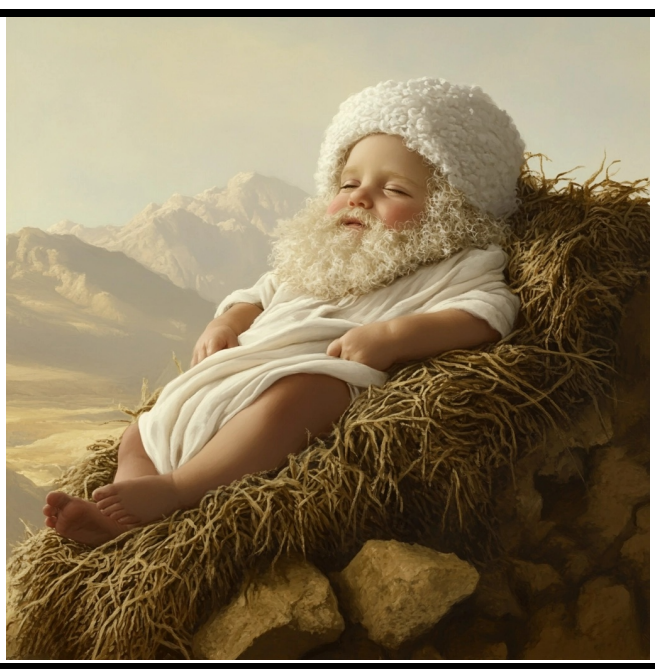 | | 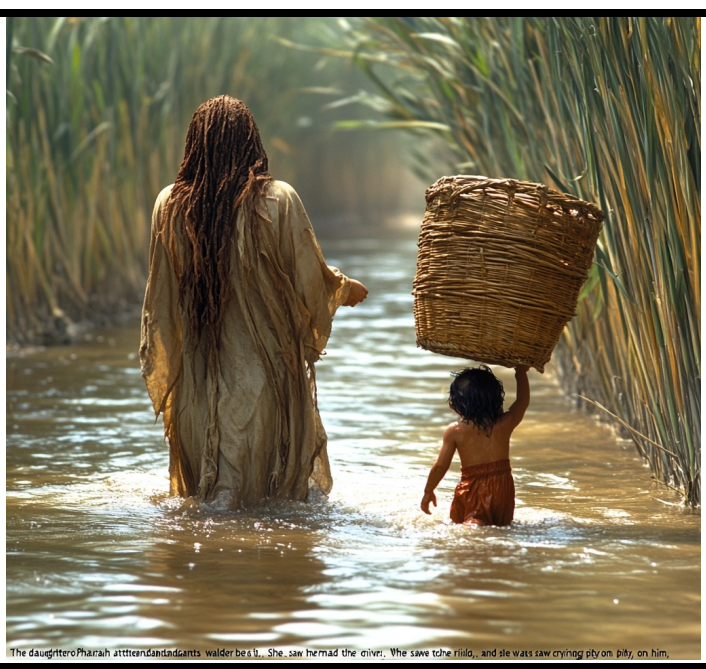 | | 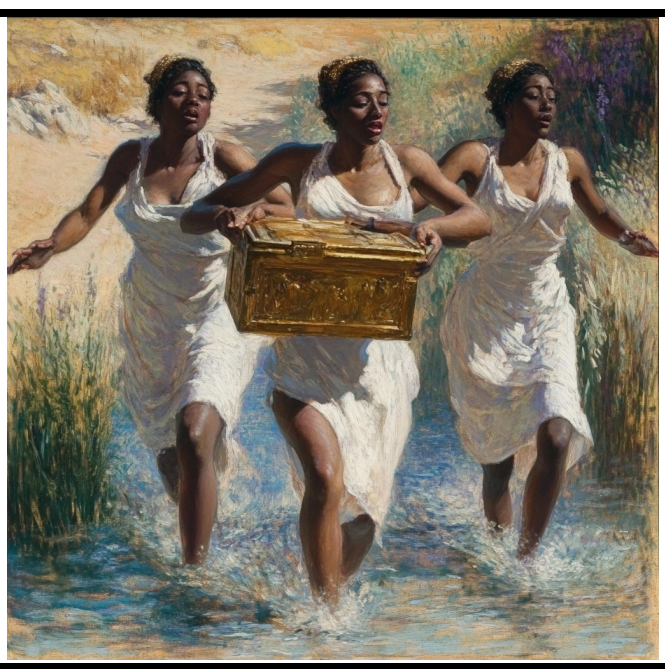 |
| Prompt: Images of baby Moses found according to Exodus 2:5–9 | | Prompt: NIV | | Prompt: from an African point of view [stylize: 550, weirdness: 2900, chaos 40, style: raw] |

Another surprising element in the generated images that can be regarded as inaccuracies rather than innovations is the occurrence of men. This inclusion subtly reinforces male oversight, even in moments where biblical women are the central figures. Int reminds us of Cheryl Exum's critique of how male figures are inserted into biblical women's narratives without textual basis and how biblical women's agency is often minimized or reinterpreted in visual culture.[34] Traditionally, the scene of Moses' discovery is an all-female narrative, emphasizing Pharaoh's daughter, her maidens, and Moses' sister as key actors.

[34] Exum, Plotted, Shot, and Painted.

| | | | | | |
|---|---|---|---|---|---|
| 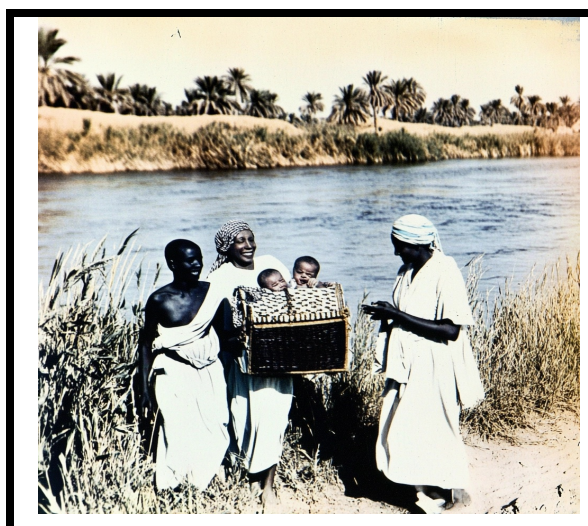 | | 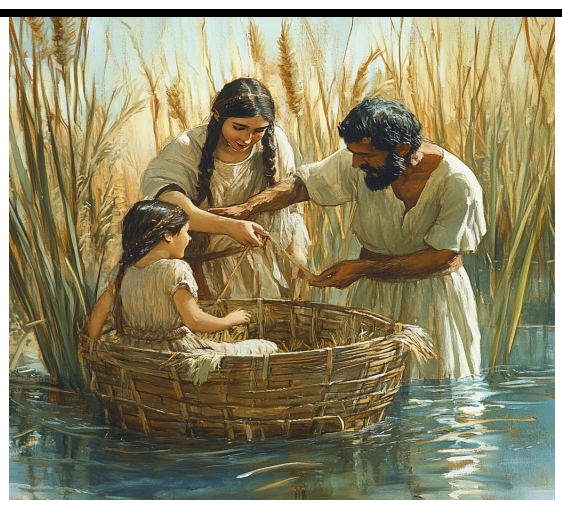 | | 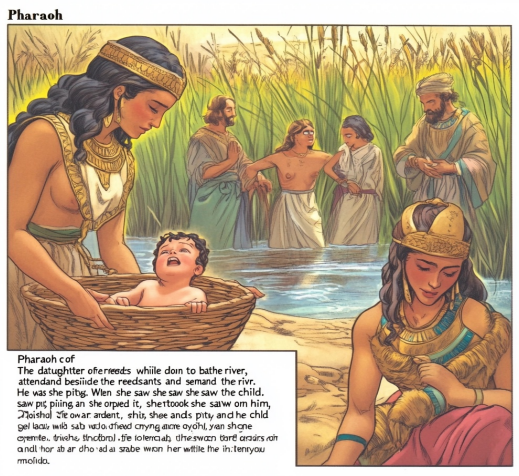 | 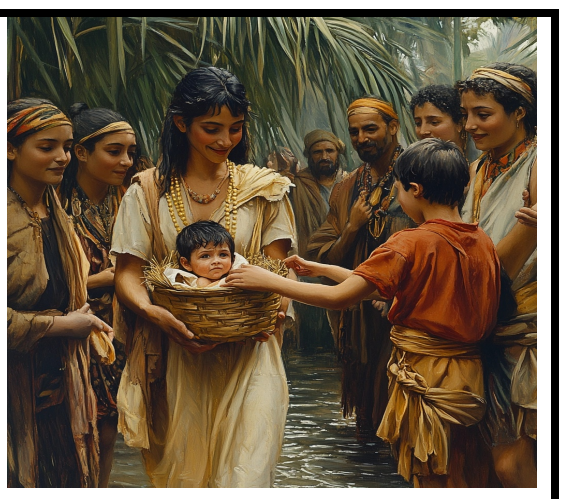 |
| Prompt: Surrealism + increased weirdness | | Prompt: using visual elements | | Prompt: NRSV | Prompt: NIV |

Although we focus in the present contribution on computational creativity, the Human-AI collaboration should be mentioned here as well: The creative process is a partnership where the user defines the prompt and refines outputs, guiding the AI's generative process. For instance, we might adjust prompts to emphasize certain details, such as the reeds symbolizing concealment or the basket as a vessel of salvation.[35]

| | | | | |
|---|---|---|---|---|
| 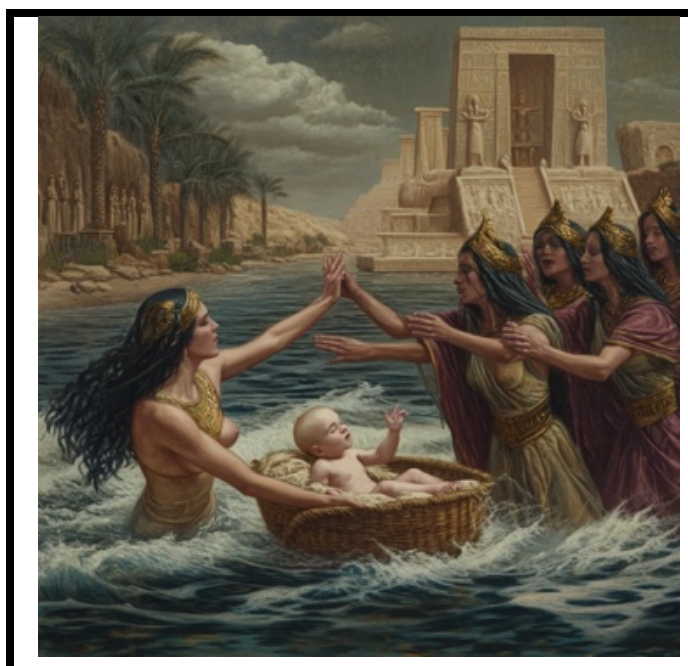 | | 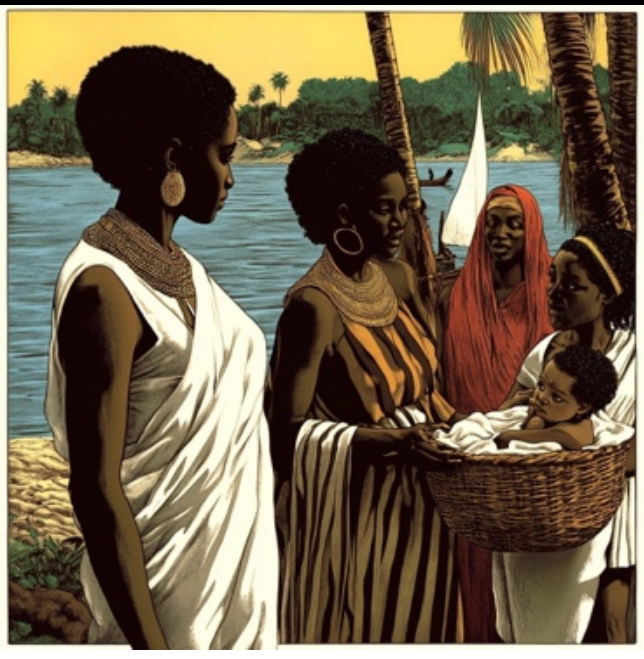 | | 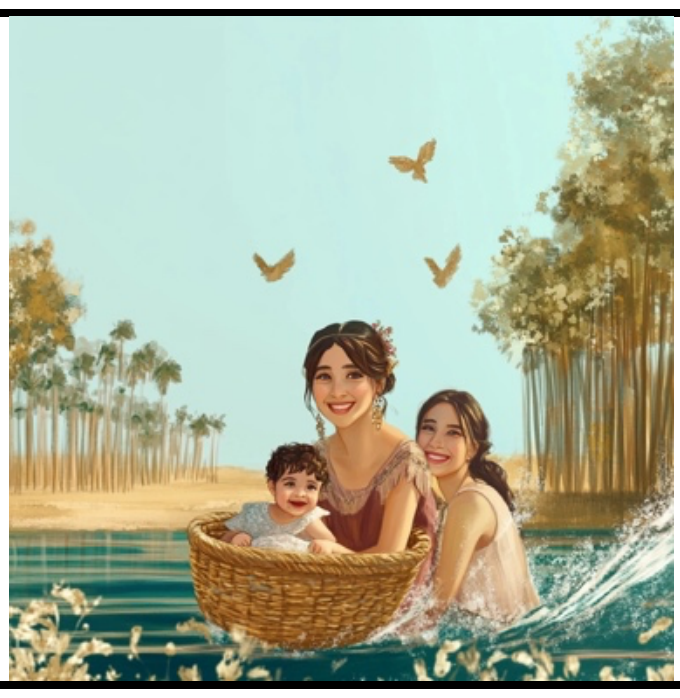 |
| Prompt: an image of albino baby Moses being found in a basket at the Nile River by the royal daughter of Pharaoh and her maidens.<br>[Weirdness: 2000, stylize: 250, chaos: 40] | | Prompt: in postcolonial African style | | Prompt: a modern-day image of baby Moses being found in a basket at the Nile River by the Gen Z daughter of Pharaoh and her friends<br>[Weirdness: 3000, chaos 40] |

Referring to Margaret Boden's distinction between three forms of creativity, Mark Graves evaluates the limitation of artificial creativity. Boden distinguishes between combinational creativity (combining old ideas in new ways), exploratory creativity (navigating within existing conceptual spaces) and transformational creativity (changing the conceptual space itself).[36] Graves observes that AI performs well with the first two forms of creativity, but has still difficulty with the third one.[37]

[35] Cf. above, footnote 6, on prompt engineering.

[36] Boden, *The Creative Mind*.

[37] Mark Graves, personal communication (contribution to panel discussion on "The Limits of Technological Creativity ", at the GoneDigiTal 2025 conference).

This observation is confirmed by our observations. The examples of creativity we have seen above confirm the observation. Especially in te combinational creativity, Midjourney excels. Boden's framework helps to divine both the scope and the limitaitons of Midjourney's creativity.

### 4.2.2 Theological and Artistic Innovation

Artistic dimensions that convey the theological significance of the biblical narrative can also be found in AI-generated images. To formulate it more carefully: there are elements in the images that to the human observer reflects symbolic and theological aspects. For example:

- An exaggeration of the smallness of the basket against the vastness of the Nile highlights themes of vulnerability and protection.
- The reeds may be interpreted as a metaphor for divine concealment.
- A depiction of Moses' sister with expressions of strategic courage as she approaches Pharaoh's daughter highlights here role of the story and the theme of courage.
- An interplay of light and dark may be interpreted as representing danger and salvation.
- The globalized perspective on the Moses story, moving beyond Western-centric depictions (cf. above, section 4.2.1) can make the narrative more relatable to different communities.
- The blending of historical and futuristic visions can be described as AI's ability to incorporate futuristic or anachronistic elements (e.g., surreal landscapes or modernized Nile settings) reimagines the narrative in fresh, thought-provoking ways.[38]

Whereas the modernistic or globalized images and the blending of styles and periods may be criticized for lacking historical and geographical accuracy—although the same reproach can be made to the Google search images or major works from art history— they are significant from a theological perspective. They align with the view of the Dutch reformed theologian Oepke Noordmans that biblical stories should be read in our own context.[39] This can be considered as a form of incarnation or accommodation.[40]

## 4.3 Comparison with random Google images

The starting point of our research (see section 1) was that using a biblical story as a prompt frequently produces results resembling widely available depictions of the same story found online. This implies that the generated images can become increasingly predictable. They "improve" by being more successful in imitating human visualisations. When we compare the AI-generated images and the results of a random Google search, we can, however, also note some interesting differences.

There are some creative aspects that we observed in section 4.2.1 that mark a difference with the online pictures. Firstly, the online images are often in one particular style, most of them rather popular and naturalistic, whereas MiidJourney blends historical accuracy with stylistic creativity. This results in unique renderings incorporating different cultural influences, alternative artistic traditions, and even surrealist or post-colonial reimaginings. This effect can be reinforced by prompting. For example, prompts requesting "Moses found in a Renaissance style" or "in Egyptian style" yield images that fuse historical aesthetic conventions with contemporary artistic liberties. This capacity to generate fresh perspectives extends beyond mere replication; MidJourney operates as a creative partner, allowing users to refine and adjust prompts to influence outcomes.

[38] Obviously, the question as to whether and how we should distinguish between creativity and inaccuracy (cf. above, section 4.2.1) applies here as well.
[39] Noordmans, "Gogoltha".
[40] I thank my colleague Prof. Henk van de Belt (oral communication) for this suggestion.

Secondly, T2I generation excels in photorealistic pictures. Combined with artistic abstraction this expands the possibilities of digital narrative approaches. This can be further reinforced in the prompting, resulting in the representation of various global perspectives of biblical personalities.

## 4.4 Comparison with works from art history

### 4.4.1 Mimesis and Poiesis

Not bound, as it seems, by artistic, religious or cultural conventions, Midjourney mixes various art styles, may include exaggerated details, and even introduces non-conventional elements without the need to meet religious and cultural requirements, AI art can be allowed to experiment, thus a way of reimagining biblical visual narratives. It exerts a certain freedom in how the figures are positioned and is not bound by, for example, a convention to give Moses a central place. At times it shows the potential to fresh reinterpretations that extend the visual horizons of human art. the freedom for alternative aesthetics, ethnic representations, and fantasy creatures that were challenging for classical painters to attempt.

The dependency of T2I generation on its vast amounts of training data, which is often used as an argument against its status as true art, has parallels in the power of artistic conventions in art history. Not only do we see both common elements and variations in various depictions of the same narrative by one author, but also the indebtedness of one artist on another. Even famous painters which are celebrated as innovative geniuses built upon the work of others, partly imitating them, partly giving their own personal touch to it.[41]

Another argument that is often put forward to show the excellence of human art over AI-generated art, namely the inaccuracies in the latter, is not convincing. Even a masterpiece such as Poussin's painting of Moses Found contains many historical and geographical inaccuracies (see below, section 4.4.2). As we argued in an earlier contribution: It is hard to apply a fundamentally different evaluation to, say, skyscrapers in the background of the Last Supper in a AI-generated image and mediaeval castles as the backdrop of the same scene in a Renaissance painting.[42]

All these considerations highlight the similarities between human art and AI-generated art as an expression of accuracy (related to *mimesis*) and novelty (related to poiesis). There are, however, some aspects in human artworks that are unparalleled in the computational artwork, which are related to careful and meaningful composition and symbolic significance and emotional depth. Our discussion of these aspects starts with an exploration of one of the paintings listed above (section 3.3), namely Nicolas Poussin's painting of the Moses Found episode.

---

[41] E.g., although in some cases a famous painter may also have used another picture (cf. the example of Rembrandt's Balaam (1626), based on a painting by Pieter Lastman 1622).
[42] Makimei, Wang and Van Peursen, "Seeing The Words: Evaluating AI-generated Biblical Art", 12.

### 4.4.2 Case study: Nicolas Poussin

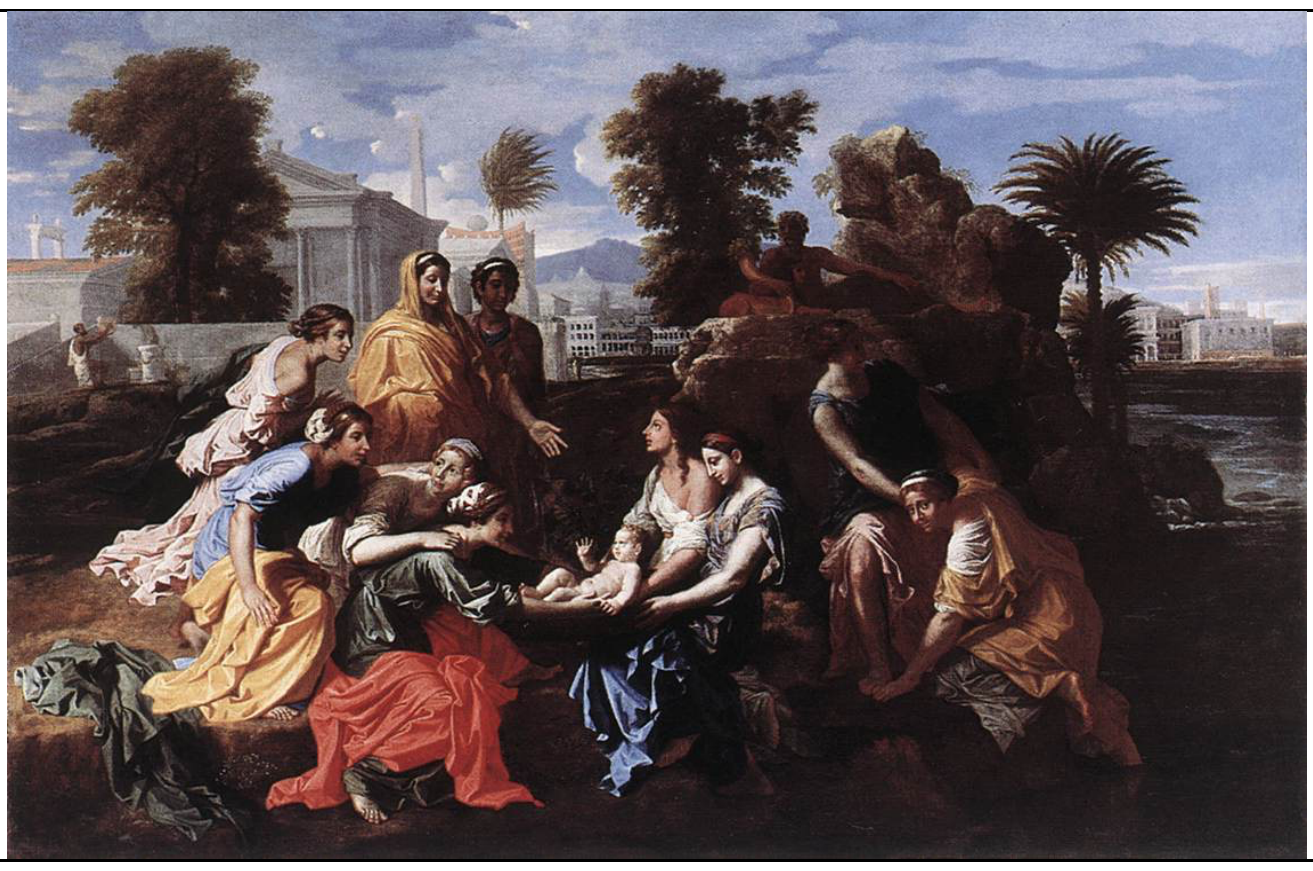

The Finding of Moses –Nicolas Poussin (1651)

By way of illustration, let's take one example from the paintings listed in section 3.3: Nicolas Poussin's The Finding of Moses,43 which he painted in 1651. Poussin produced almost twenty paintings of Moses,44 including three paintings of the episode of Moses Found. This is the last version. It is an interesting example of a representation of the episode in Classicist style.45 In the background we see buildings that are reminiscent of the grandeur of ancient Greece and Rome and a landscape that looks European rather than Egyptian. However, the Egyptian environment is indicated by the river god (symbolizing the Nile) sitting on the rock embracing a sphynx, palm trees at the river shore and an obelisk. In the background an Egyptian priest worships the dog-shaped god Anubis. This depiction of an Egyptian setting, blending it with elements from classical Europe, and the interest in pagan gods is characteristic of the Classicist style period to which Poussin belongs.

The painting resonates with a traditional symbolic Christian reading of the episode, which considered Moses as a type of Christ and the waters of the Nile in which the Israelite babies were drawn as a prefiguration of baptism. Various figures in the painting are reminiscent of figures in other paintings: the baby Moses resembles Christ in Poussin's Adoration of the Magi and his Adoration of the Shepherds. Moses' mother is clothed in blue, a color of the Holy Virgin. Interestingly the balanced composition with the grouping of figures in the foreground and a landscape with static buildings in the background is reminiscent of another painting by Poussin, which he painted one year earlier for the same patron, namely Christ healing the Blind Man. This also shows how this scene centered around Moses is modelled on scenes in which Christ takes central position.

In the 17th century in which Poussin lived, in addition to Moses' theological significance as a prefiguration of Christ, there was also another symbolic layer. Based on the claim that "Moses was educated in all the wisdom of the Egyptians" (Acts 7:22), he was considered a bearer of divine knowledge, who stood at the start of a lineage of transmission of occult knowledge from Hermes

[43] Our discussion of this painting is indebted to ihttps://www.wga.hu/html_m/p/poussin/4/07moses.html
[44] On Poussin's fascination for the figure of Moses see Alain Mérot, *Nicolas Poussin* (New York: Abbeville Press, 1990), pp. 168–169 and the discussion below.
[45] For Poussin's earlier paintings of the same period (painted 1638 and 1947 and a drawing of the same episode, see Suzanne-Claire Guillais, *Poussin*, Paris: Atlas, 1994), pp. 46, 48–49 and Mérot, *Nicolas Poussin*, p. 174.

Trismegistus to Plato. It has been argued that Poussin was sensitive to this aspect as well so much that he did not hesitate to associate Moses with Egyptian gods in his compositions.[46]

The symbolic presentation of the episode is set in a carefully crafted composition which is "based on sweeping concentric curves" and in which the figures are crowded "in a single animated movement around a central axis".[47] Every figure takes its own place: the princess, standing there "generous and commanding", her maidens "curious and delighted", and the "humbler figure in a white shift at Moses' head", most likely Mirjam.[48] The faces do not express much emotions, but the posture and the gestures do. Poussin "preferred to convey emotion by the action and gestures of their figures, rather than through facial expression".[49]

In contrast with the figures surrounding the healing of the blind man, all figures surrounding this painting are women. Their roles, including the commanding attitude of Pharao's daughter and the absence of any male figure except for the passive baby boy, deviates from the general tendency in art history that female biblical figures were often positioned in ways that emphasized passivity or divine selection rather than personal agency.[50] All women in the picture, especially Pharao's daughter and Moses' mother contribute to the symbolic depth of the painting.

As it is concluded on the Web Gallery of Art page of this picture:

> Bodies and colours, each distinct and separate, combine in ample rhythms across the picture surface, echoed by the rocks beyond. It is at once solemn and joyful, as befits a scene in which a child is rescued from death, and through him an entire people is saved.

A comparison of his paintings of this episode, as well as closely related episodes such as "Moses Abandoned" (Exodus 2:3), is beyond the scope of this article. It has been observed that in those themes that he painted repeatedly, such as the Holy Family or Moses Found "it is as if, with every new picture he painted, Poussin became increasingly aware of the richness of meaning inherent even in apparently hackneyed subjects".[51]

Symbolic depth, careful composition, expression of emotion, and, as it seems artistic or symbolic intentions behind every small detail on the painting can be found, mutatis mutandis in all the other paintings listed in section 3.3).

### 4.4.2 Emotional and symbolic layers

In both the AI-generated images and the works from art history we find elements that from the perspective of biblical interpretation are inaccurate, be it elements of Moses' age (in the AI images) or the landscape surrounding the river Nile (in Poussin's painting). Ignoring for the moment the question as to whether it is justified to label these elements as errors, we can say that the artworks do not score higher than AI-generated images. The background, the clothes or the people are often not what we expect from an Egyptian setting at the river Nile.

---

[46] *Poussin et Dieu: l'album de l'exposition* (Paris: Musée du Louvre, 2015), p. 28,

[47] Mérot, *Nicolas Poussin*, p. 174.

[48] Quotations taken from https://www.wga.hu/html_m/p/poussin/4/07moses.html

[49] Esther Sutro, *Nicolas Poussin* (London: J. Cape e Medici Society, 1923), p. 95 (the quote is part of a comparison with William Blake who, according to Sutro, had the same preference.

[50] Apostolos-Cappadona, *Biblical Women and the Arts*, "Introduction, 1. On gendered and cultural biases" see also J. Cheryl Exum, *Plotted, Shot, and Painted: Cultural Representations of Biblical Women (*Sheffield: Sheffield Academic Press, 2004)

[51] Mérot, *Nicolas Poussin*, p. 166.

The main difference is in the emotional and symbolic depths of those paintings. The masterpiece of the history of art excels in emotional storytelling since artists go to great lengths to detail facial expressions, gestures, and postures specifically to convey intense human feelings. This painstaking process allows for nuanced expressions of feeling so that the characters are living and psychologically complex. AI-created pictures, though, tend to fail at emotional complexity, sometimes producing generic or too stylized facial expressions that miss the depth and intentionality of hand-painted works. AI can create extraordinarily expressive faces, but these are often uniform or exaggerated, without the close-to-human touch found in traditional compositions.

The symbolic depth of the human artworks may be in objects, colors and compositions, as we have seen in the example of Poussin's painting. The symbolic and theological depth of a painting may be experienced so intensively that, for example, only one painting by Rembrandt, his famous interpretation of The Return of the Prodigal Son, inspired Henri Nouwen to write an entire book reflecting on its spiritual significance.[52]

In an episode like ours, where women are the main actors, the symbolic presentation of women plays an important role. Thus, Poussin's use of blue for Moses' mother has a deep symbolic significance. This agrees with Diane Apostolos-Cappadona's observation that classical depictions of biblical women often employ color to represent purity, nobility, or danger.[53]

Apostolos-Cappadona further argues that biblical women in Western art have often been visually coded through gestures, colors, and objects that carry symbolic theological meanings.[54] For example, Pharaoh's daughter is traditionally depicted in white or gold garments, symbolizing both royalty and divine favor. Classical artists structured their compositions to emphasize her status as an agent of salvation—often placing her in a central, illuminated position within the artwork, a technique that reinforces her role in God's providential plan.

In contrast, AI-generated images frequently disregard or modify these visual codes, introducing vibrant, fantastical hues that may enhance aesthetic appeal but detach the imagery from its original theological meaning. Sometimes it exaggerates elements like clothing or alters the lighting and positioning of figures in ways that disrupt the classical iconographic tradition. For instance, some AI renditions of Pharaoh's daughter depict her in excessively colorful, jewel-adorned attire, resembling Orientalist or fantasy-themed depictions rather than historically accurate biblical artwork.

### 4.4.4 Composition

The symbolic bearing of a picture is often strengthened by its composition. Classical paintings are defined by a commitment to conveying a coherent story as powerfully as possible, employing the principles of composition such as the golden ratio, perspective or depth of field, embedded in specific art historical periods such as the Renaissance or the Baroque. Artists intentionally arranged

---

[52] Nouwen, The Return of the Prodigal Son.

[53] Apostolos-Cappadona, *Biblical Women and the Arts*, Chapter 3, "Beauty and its Beholders: Envisioning Sarah and Esther,"

[54] Ibid.

figures in ways that guide the viewer toward critical moments of the story.[55] As we have seen above, the composition not only meets aesthetic criteria, but also serves conveying symbolic layers, such as the parallels between the the finding of Moses in the Nile and the Adoration scenes centered around baby Christ.

While MidJourney is more than capable of structured compositions, they are often plagued by a battle between perspective and narrative due to image generation being based on probabilities instead of artistic intent.[56] At some points, the generated images suggest a story's depth, as we indicated in section 4.2.2. For example, an exaggeration of the smallness of the basket against the vastness of the Nile seems to highlight themes of vulnerability and protection. This is, however, only a symbolic meaning that can be assigned to the image by the human observer. Given the many pictures that need to be generated before a human researcher found an image with such a symbolic expression suggests that it appears only randomly.

### 4.4.5 Authorial intent

The last sentence of the preceding section brings us to the question of authorial intent. All our observations above suggest that an essential difference between human art and AI art consists in a lack of authorial intention in the latter. Even in those cases where we perceive symbolic depth or careful composition in an AI-generated picture, we can only localize it in the observer, not in the author. It is not for nothing that, precisely in relation to AI-generated art, there is talk of "the death of the author".[57]

The role if authorial intent can be avoided if we locate the "intent" in the work of art rather than in the artist, Kostas Terzidis, Filippo Fabrocini and Hyejin Lee argue that, the intentionality of the artist does not have any relevance if we focus just on the outcome of the artistic process and that what matters is the artwork's "intent". They use the term "unintentional intentionality" to indicate a work's intent that is free of human intervention, therefore "unintentional".[58] As early as 1950 Alan Turing famously raised objections to defining intentionality as the difference between human and computer, we see here a crucial element that to this day distinguishes human art from AI-generated images.[59]

This contrasts with a theological view in which creativity is understood as inseparable from that of a creator (cf. above, section 1). Although we are aware of the objections raised by Turing as early as 1950 to defining intentionality as the difference between human and computer, we see here a crucial element that to this day distinguishes human art from AI-generated images.[60]

It should be noted, however, that authorial intention in human painters is also a complex concept. We cannot assume a priori that painters who produced deeply religious paintings were expressing their own spirituality, or that they sympathized with the emotions expressed in the paintings. However, we can speak of conscious attempts by human painters to make their paintings an access to deep symbols, religious feelings or spiritual experiences.

---

[55] Cf. Gombrich, *The Story of Art*, 299, on Rubens' "unrivalled gifts in arranging large colourful compositions".
[56] Hertzmann, "Can Computers Create Art?" 26.
[57] Cf. Kreminski, "Endless Forms Most Similar".
[58] Terzidis, Fabrocini and Lee, "Intentional intentionality" (thanks to Jody Hunt for this reference).
[59] Cf. Turing, "Computing Machinery and Intelligence".
[60] Cf. Turing, "Computing Machinery and Intelligence".

### 4.4.6 Materiality and immediacy

One important aspect of art should be added to the discussion, and that is the material, physical presence of the objects of the world. Till now we have compared AI-generated images with images of works from art history. But an image is not the artwork itself. Poussin's painting is not just an image, it is a big real painting (115.7 × 175.3 cm), exhibited in the National Gallery in London. In an essay from 1945, Walter Benjamin speaks of the perception of an object's aura, that is: its unique existence in time and space. For Benjamin, presence ('Das Hier und Jetzt des Originals') is a prerequisite of authenticity ('Echtheit').[61] In his essay 'Das Kunstwerk im Zeitalter seiner technischen Reproduzierbarkeit' ('The work of art in the age of mechanical reproduction'), he argues that the distance or unapproachability ('Unnahbarheit') of a work of art cannot be overcome by mechanical reproduction because an object's aura cannot be reproduced.[62] Paintings are not only images, they are also brush strokes, the materiality of the paint used, and the size. This size was decided upon by the artist, often related to the purpose of a painting, its intended location, or the effect that the painter wanted to achieve.

For the human encounter with human artwork, let's consider, for example, E. Sammut's description of his experience of seeing Carvaggio's *Beheading of Sint John*. This is the largest painting by Carvaggio, which can be found in St John's Co-Cathedral of Valletta, Malta. Much has been written about the finest details of brush strokes, the contrast created between the male and female figures in the picture, and the expression of emotion and drama. Sammut describes the experience of seeing this painting as follows:[63]

> What immediately leaps to the eye in this picture is its unusual composition (...) The five figures in this picture are placed in a well-knit group forming a perfect semi-circle at one side, with the oblique lighting leading the eye to the centre of attraction, the head of St. John held in the grip of the executioner. Nearly three quarters of the canvas, however, are left practically blank. To the left of the group is a barred window through which dim figures can be discerned peering at the gruesome scene; whilst behind it, serving as a dropscene for the compact but sordid drama, is a massive arched doorway which might easily be one of the side entrances of St John's itself. (...) It is evident that, having mastered the use of light in all its intricacies as one, and probably the most important, of his dramatis personae, he is here impressing the whole atmosphere of the scene into his service to enhance the psychological effect of his painting.

# 5. Conclusions

Our research has provided a critical exploration of AI-generated biblical imagery, particularly focusing on how MidJourney interprets the narrative of Moses being found in Exodus 2:5–9. By analyzing AI-generated images, classical paintings, and Google search results, this study has demonstrated the interplay between mimesis (imitation of traditional art) and poiesis (creative reimagining) in biblical visual storytelling.

---

[61] Benjamin, 'Das Kunstwerk', 352: "Das Hier und Jetzt des Originals macht den Begriff seiner Echtheit aus"; English translation (ed. by H. Arendt; transl. by H. Zohn, p. 220): 'The presence of the original is the prerequisite to the concept of authenticity'.

[62] See especially the Zweite Fassung in *Gesammelte Werken* 1/2, 480, note 7.

[63] Sammut, "Caravaggio in Malta", 85–86; see also Sutherland Harris, *Seventeenth-Century Art & Architecture*, 48.

The dataset analysis revealed that certain elements, such as Pharaoh's daughter and Moses in a basket, are consistently depicted, while others, such as the number of maidens, Pharaoh's court, and the setting details, vary greatly. We observed some distortions, anachronisms, and inaccuracies in AI depictions of Moses' discovery. Certain AI-generated images misrepresented historical settings, introduced inconsistent ethnic portrayals, or overemphasized decorative elements, altering the visual storytelling. Google search images revealed dominant artistic trends in widely circulated biblical art. Many results favored Eurocentric depictions influenced by classical Western religious paintings, limiting the diversity of artistic interpretations.

This project also examined AI's theological and artistic contributions, particularly how poiesis expands the boundaries of biblical interpretation. AI introduces futuristic and symbolic reinterpretations, blending historical, surreal, and modern elements into visually compelling compositions. For instance, MidJourney's ability to generate stylized, painterly, or photorealistic representations enables a more diverse and imaginative visual engagement with scripture than Google search images, which primarily return pre-existing religious art. However, AI also introduces inherent limitations, such as inaccuracies in character representation, anachronistic settings, and inconsistent cultural depictions.

In comparison to classical paintings, MidJourney exhibits some level of artistic flexibility but lacks the intentional theological symbolism embedded in classical religious art. Traditional biblical paintings were crafted with a deep theological purpose, using light, gestures, and color to convey divine action and emotion. AI-generated images, by contrast, prioritize aesthetic appeal over theological depth, leading to images that are visually captivating but sometimes theologically ambiguous.

Ultimately, this project underscores the transformative potential and challenges of AI in religious art. While AI-generated biblical imagery offers new possibilities for creative theological engagement, it also raises critical questions about historical accuracy, artistic bias, and the evolving role of AI in shaping biblical narratives. Moving forward, future studies could explore how AI's training datasets influence its biblical depictions, how AI-generated religious art is received by different faith communities, and how AI could be ethically integrated into theological and artistic traditions.

In the evolving landscape of digital biblical interpretation, AI functions not just as a replicator of sacred imagery but as a participant in the creative process, shaping new ways of seeing and understanding biblical text in the modern digital age.

# Appendices

## Appendix 1: AI-generated images

| Fig | Images | Visual Elements Present | Prompt & Observations |
|---|---|---|---|
| 1. | 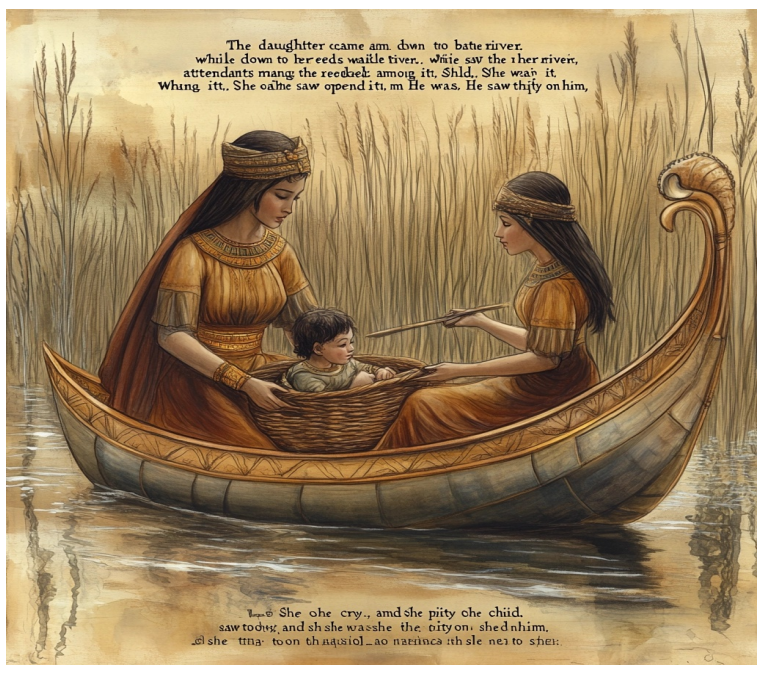 | ☑Pharaoh's Daughter Coming to the River<br>☐ The Basket in the Reeds<br>☐ The Maidens/Servants<br>☑ The Opening of the Basket<br>☐ The Child Crying<br>☐ Dialogue and Interaction<br>☑ The River and Reeds as Setting | Prompt: NIV<br>Evaluation: A tranquil scene with Pharaoh's daughter and another figure in a decorative boat surrounded by reeds. There seem to be no attendants in the picture and the other figure/woman present looks like a child, probably the sister of Moses. The detailed attire and gentle expressions highlight care and reverence, though the ornate boat introduces an artistic embellishment. |
| 2. | 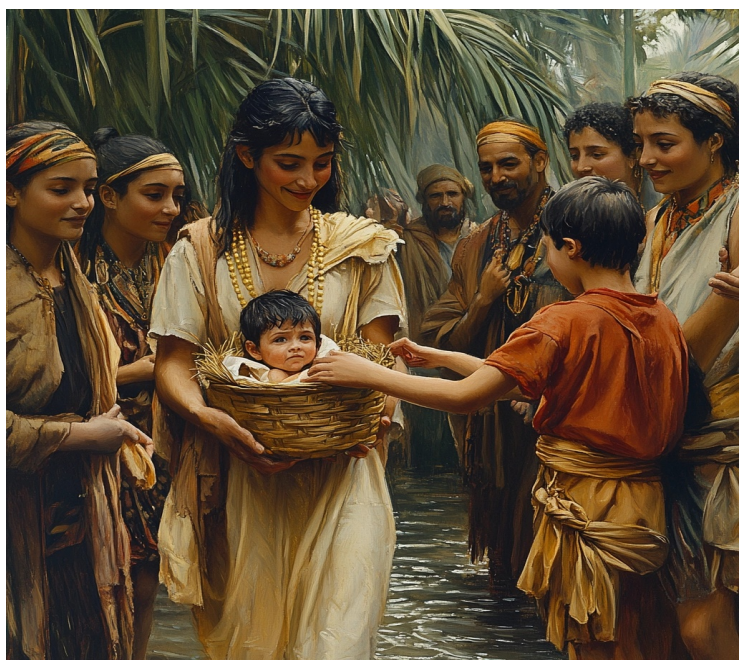 | ☑ Pharaoh's Daughter Coming to the River<br>☐ The Basket in the Reeds<br>☐ The Maidens/Servants<br>☑ The Opening of the Basket<br>☐ The Child Crying<br>☑ Dialogue and Interaction<br>☐ The River and Reeds as Setting | Prompt: NIV<br>Evaluation: A vibrant, joyous composition with richly dressed figures holding a child. The palm-filled background and dynamic interaction between characters highlight themes of rescue and celebration. This image has some men in the background. |
| 3. | 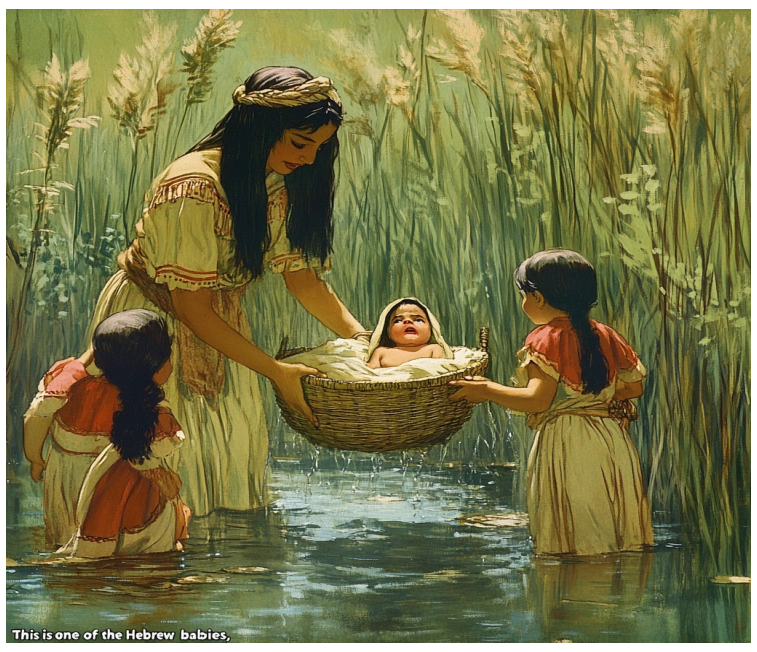  | ☐ Pharaoh's Daughter Coming to the River<br>☐ The Basket in the Reeds<br>☐ The Maidens/Servants<br>☐ The Opening of the Basket<br>☑ The Child Crying<br>☑ Dialogue and Interaction<br>☑ The River and Reeds as Setting | Prompt: KJV<br>Evaluation: A vibrant and lively depiction of Pharaoh's daughter taking Moses from a young girl in a lush wetland. The scene's bright colors and expressive figures capture a sense of nurturing and wonder, aligning with the narrative's themes of rescue and divine providence. |
| 4. | 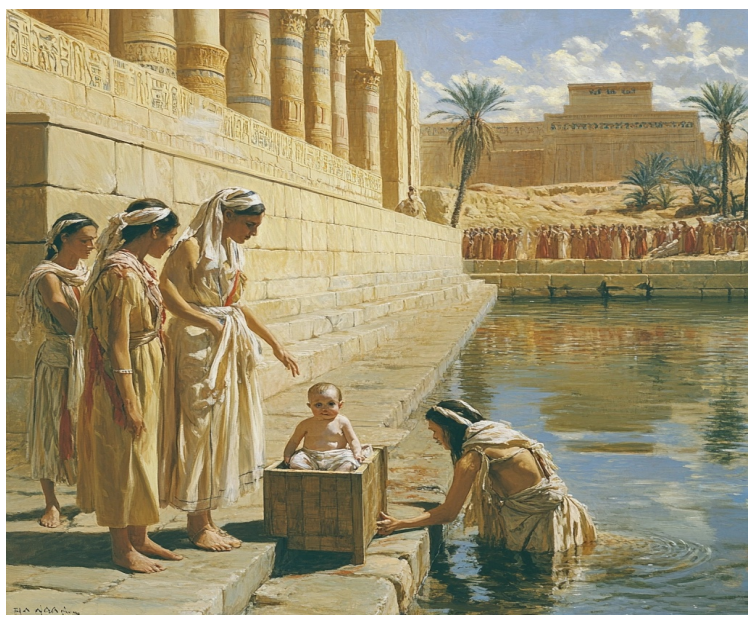 | ☐ Pharaoh’s Daughter Coming to the River<br>☐ The Basket in the Reeds<br>☐ The Maidens/Servants<br>☑ The Opening of the Basket<br>☐ The Child Crying<br>☑ Dialogue and Interaction<br>☑ The River and Reeds as Setting | Prompt: KJV<br>Evaluation: A detailed portrayal of Moses being retrieved from the river, set against the backdrop of grand Egyptian architecture. The vibrant colors and precise depiction of robes evoke a historical atmosphere, though the lighting lends a cinematic quality. |
| 5. | 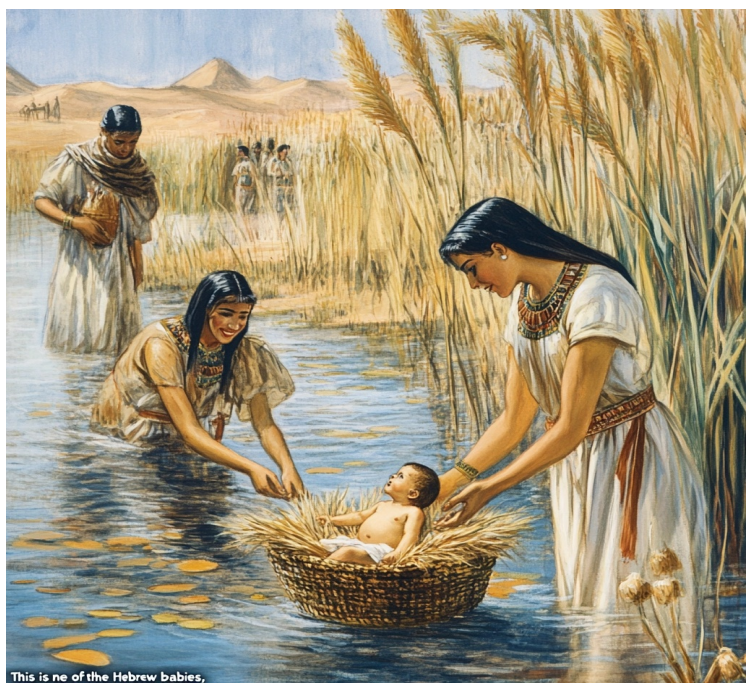 | ☑ Pharaoh’s Daughter Coming to the River<br>☑ The Basket in the Reeds<br>☑ The Maidens/Servants<br>☑ The Opening of the Basket<br>☑ The Child Crying<br>☑ Dialogue and Interaction<br>☑ The River and Reeds as Setting | Prompt: KJV<br>Evaluation: A soft depiction of Moses in a straw-lined basket being tended to by women. The naturalistic backdrop of reeds and water creates a peaceful, intimate mood, though the narrative feels simplified compared to more elaborate compositions. |
| 6. | 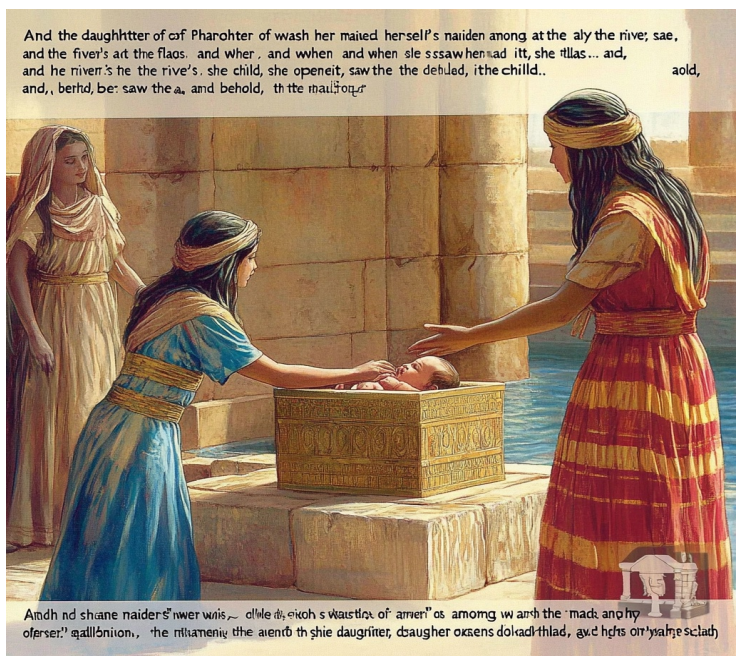 | ☐ Pharaoh’s Daughter Coming to the River<br>☐ The Basket in the Reeds<br>☐ The Maidens/Servants<br>☑ The Opening of the Basket<br>☐ The Child Crying<br>☑ Dialogue and Interaction<br>☐ The River and Reeds as Setting | Prompt: KJV<br>Evaluation: A serene composition featuring Pharaoh’s daughter and her attendants discovering baby Moses in an ornate basket. The architectural setting and warm tones emphasize a sense of royalty and sacredness, though the modern styling of the characters slightly detracts from historical accuracy. |
| 7. | 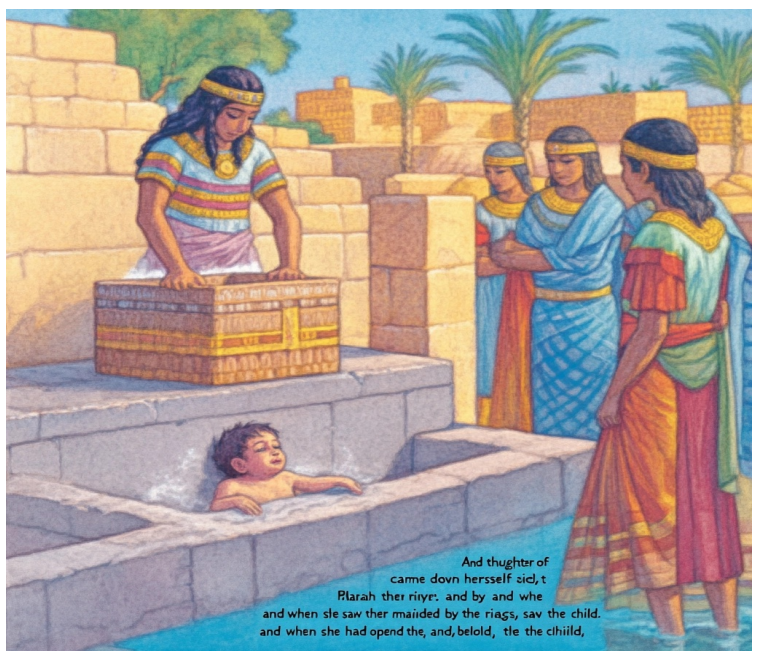 | ☑ Pharaoh’s Daughter Coming to the River<br>☐ The Basket in the Reeds<br>☑ The Maidens/Servants<br>☐ The Opening of the Basket<br>☐ The Child Crying7<br>☐ Dialogue and Interaction<br>☐ The River and Reeds as Setting | Prompt: KJV<br>Evaluation: A brightly colored illustration of Pharaoh’s daughter retrieving Moses from a river within a structured stone setting. The vibrant palette and simplified character expressions make it feel more illustrative and storybook-like than realistic. |
| 8. | 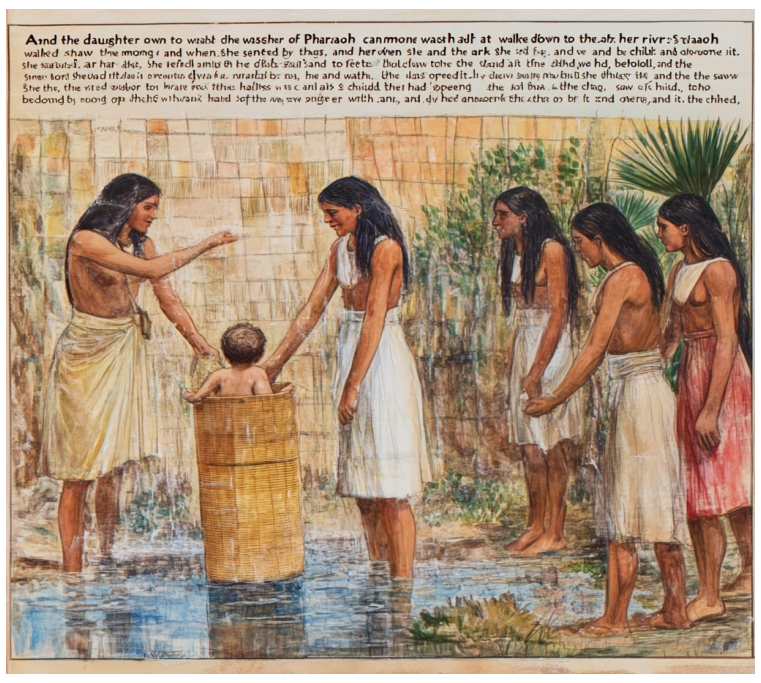 | ☑ Pharaoh's Daughter Coming to the River<br>☐ The Basket in the Reeds<br>☑ The Maidens/Servants<br>☑ The Opening of the Basket<br>☐ The Child Crying<br>☑ Dialogue and Interaction<br>☑ The River and Reeds as Setting | Prompt: KJV<br>Evaluation: A communal moment of interaction where women retrieve Moses from a cylindrical basket. The layered robes and subdued tones create a sense of calm reverence, though the scene feels slightly repetitive in its arrangement. |
| 9. | 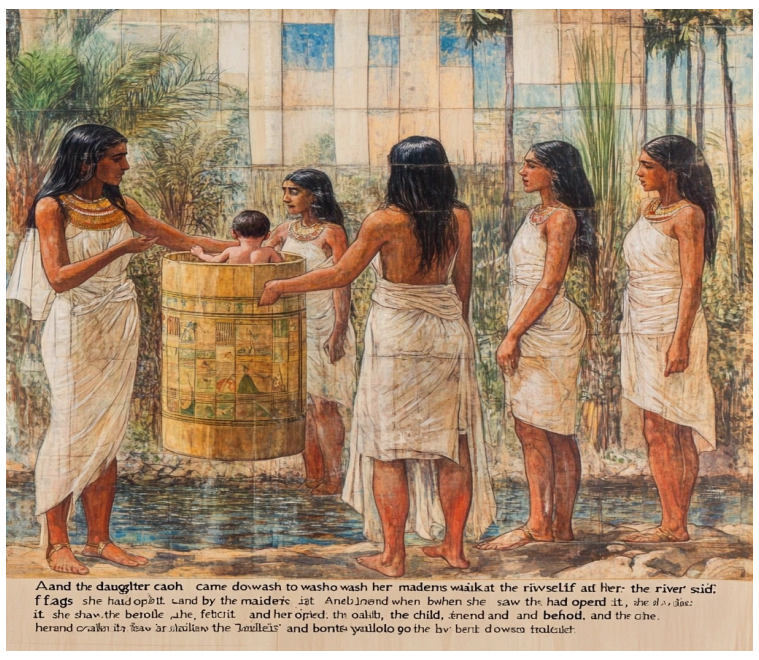 | ☑ Pharaoh's Daughter Coming to the River<br>☐ The Basket in the Reeds<br>☑ The Maidens/Servants<br>☑ The Opening of the Basket<br>☐ The Child Crying<br>☑ Dialogue and Interaction<br>☑ The River and Reeds as Setting | Prompt: KJV<br>Evaluation: A detailed and textured depiction of women lifting Moses from a tall, ornate basket. The attention to fabric and posture highlights the ceremonial nature of the moment, while the background suggests a symbolic connection to Egyptian royalty. |
| 10. | 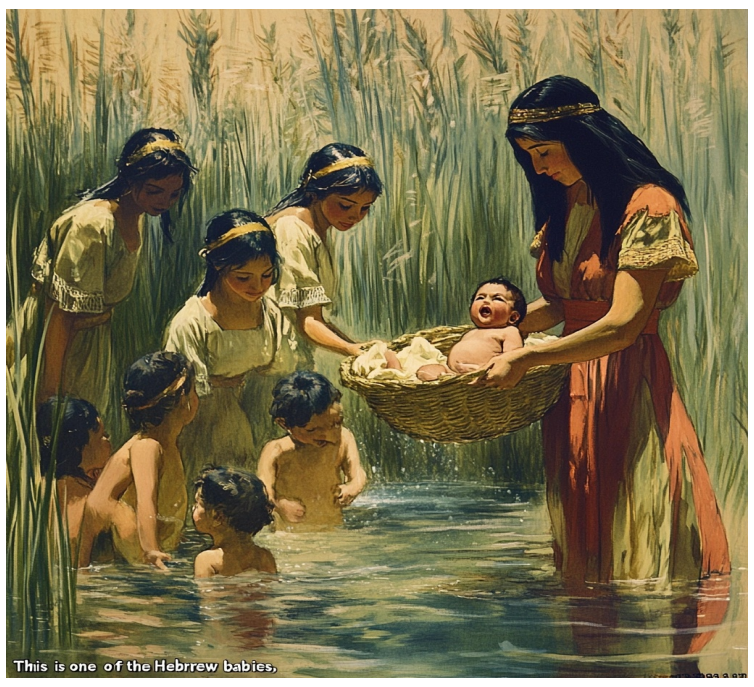 | ☐ Pharaoh's Daughter Coming to the River<br>☑ The Basket in the Reeds<br>☐ The Maidens/Servants<br>☐ The Opening of the Basket<br>☑ The Child Crying<br>☐ Dialogue and Interaction<br>☑ The River and Reeds as Setting | Prompt: KJV<br>Evaluation: A vibrant scene with women and children bathing in a lush river, surrounding Moses' basket. The joyous expressions and interaction suggest a communal celebration, though the stylistic emphasis on greenery and water lends a pastoral, almost idyllic quality. |
| 11. | 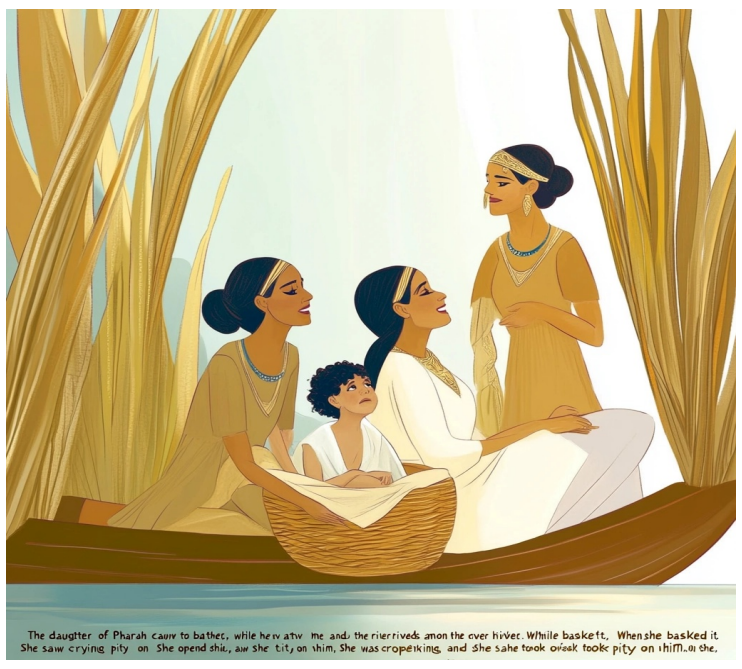 | ☐ Pharaoh's Daughter Coming to the River<br>☑ The Basket in the Reeds<br>☑ The Maidens/Servants<br>☑ The Opening of the Basket<br>☑ The Child Crying<br>☑ Dialogue and Interaction<br>☑ The River and Reeds as Setting | Prompt: NSRV<br>Evaluation: A modern illustration with clean lines and bright colors. The simplistic depiction focuses on the interaction of the women with baby Moses, creating a storybook-like aesthetic that is visually accessible but lacks historical depth. |
| 12. | 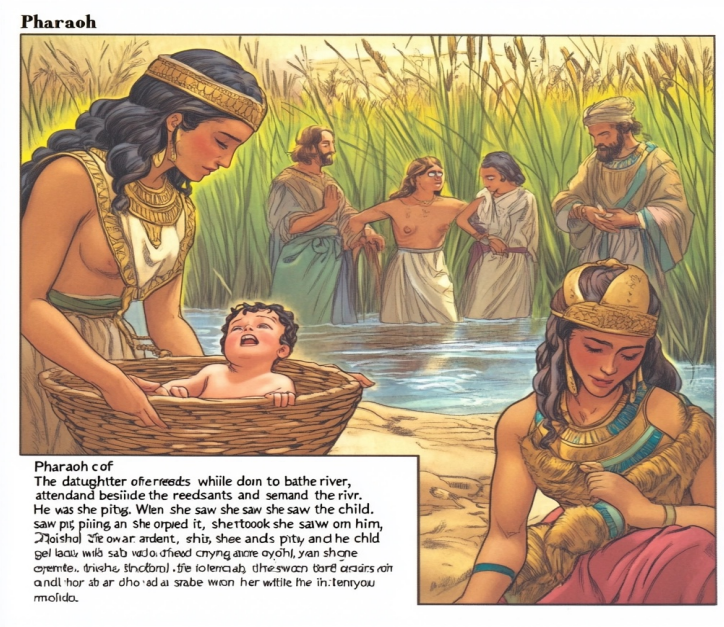 | ☑ Pharaoh's Daughter Coming to the River<br>☐ The Basket in the Reeds<br>☑ The Maidens/Servants<br>☐ The Opening of the Basket<br>☑ The Child Crying<br>☐ Dialogue and Interaction<br>☑ The River and Reeds as Setting | Prompt: NSRV<br>Evaluation: A vibrant, semi-realistic depiction of Pharaoh's daughter holding Moses in a basket, with attendants in the background. The use of Egyptian clothing and hierarchical composition evokes a sense of cultural and historical context. Also in the background are multiple men. |
| 13. | 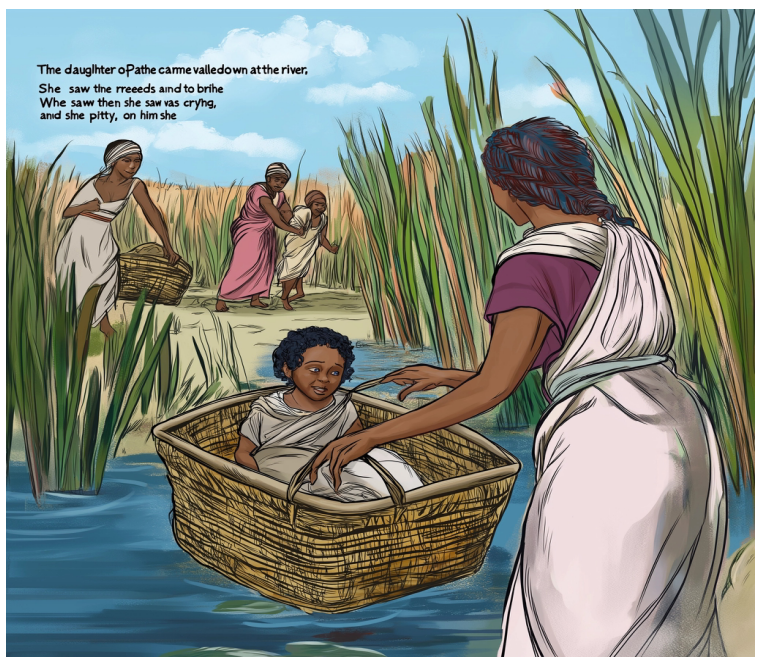 | ☐ Pharaoh's Daughter Coming to the River<br>☑ The Basket in the Reeds<br>☐ The Maidens/Servants<br>☑ The Opening of the Basket<br>☐ The Child Crying<br>☐ Dialogue and Interaction<br>☑ The River and Reeds as Setting | Prompt: NSRV<br>Evaluation: A lively riverside scene emphasizing communal care. The dynamic composition captures multiple figures engaging with Moses, but the clothing and setting reflect a more illustrative than historical approach. |
| 14. | 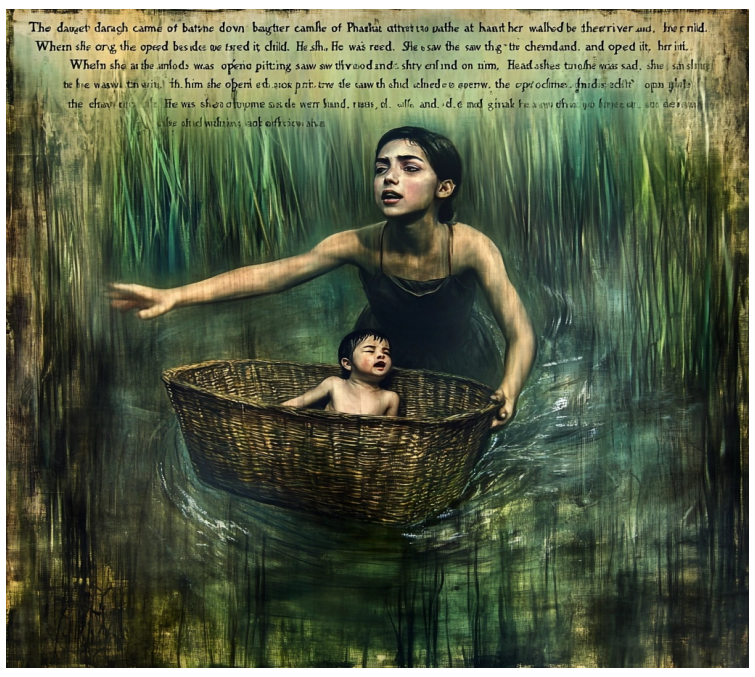 | ☐ Pharaoh's Daughter Coming to the River<br>☑ The Basket in the Reeds<br>☐ The Maidens/Servants<br>☐ The Opening of the Basket<br>☑ The Child Crying<br>☐ Dialogue and Interaction<br>☑ The River and Reeds as Setting | Prompt: NSRV<br>Evaluation: The style is reminiscent of Symbolism. A darker, dramatic portrayal emphasizing urgency and emotional intensity. The expression of the central figure and the use of shadow creates a poignant and almost haunting atmosphere. |
| 15. | 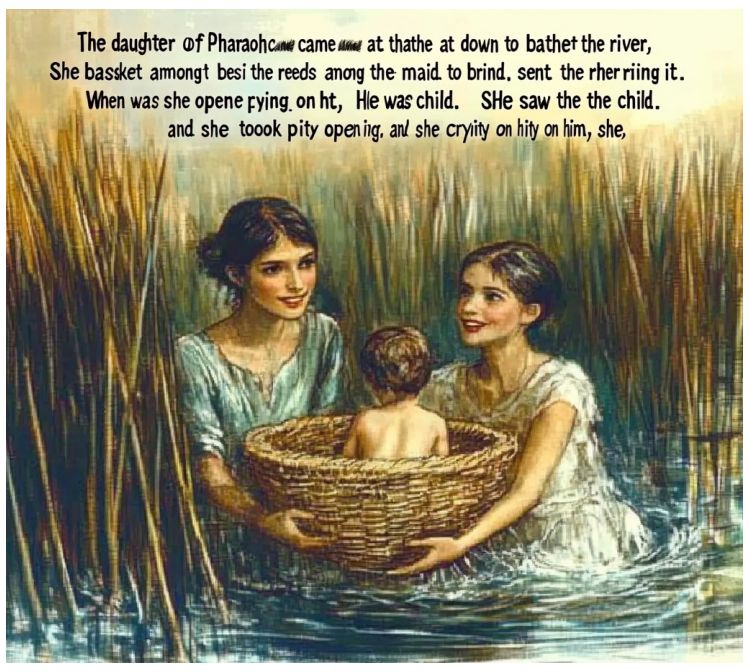 | ☐ Pharaoh's Daughter Coming to the River<br>☑ The Basket in the Reeds<br>☐ The Maidens/Servants<br>☑ The Opening of the Basket<br>☐ The Child Crying<br>☑ Dialogue and Interaction<br>☑ The River and Reeds as Setting | Prompt: NSRV<br>Evaluation: A soft, pastoral scene with women in flowing garments holding Moses. The golden lighting and natural setting suggest peace and reverence, though the simplicity may downplay the historical significance. |
| 16. | 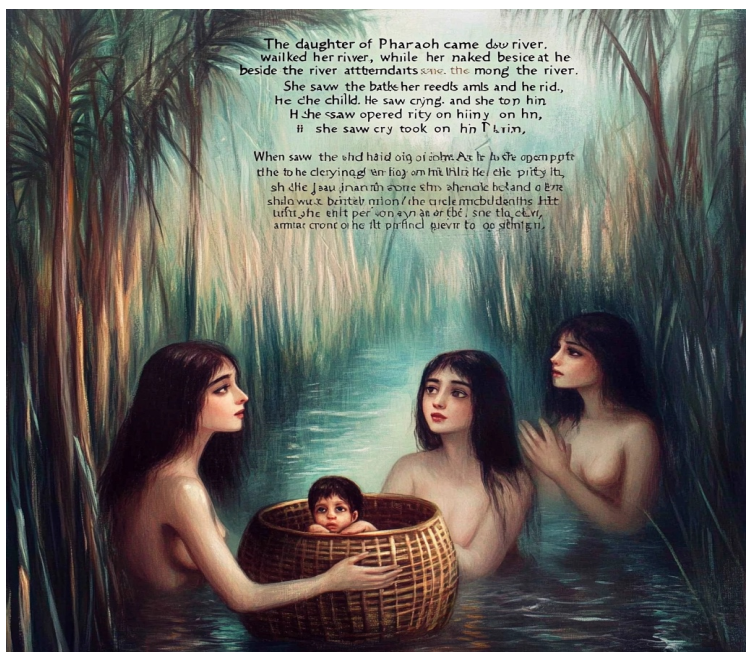 | ☐ Pharaoh’s Daughter Coming to the River<br>☑ The Basket in the Reeds<br>☐ The Maidens/Servants<br>☐ The Opening of the Basket<br>☐ The Child Crying<br>☐ Dialogue and Interaction<br>☑ The River and Reeds as Setting | Prompt: NSRV, higher scale weirdness<br>Evaluation: Another style is reminiscent of Symbolism. Like Fig. 11, with the key difference being the depiction of complete nudity as opposed to partial nudity in Fig. 11. It is unclear if the figure represents Pharaoh’s daughter bathing with her maidens, but the scene aligns with the description of Pharaoh’s daughter going to bathe. |
| 17. | 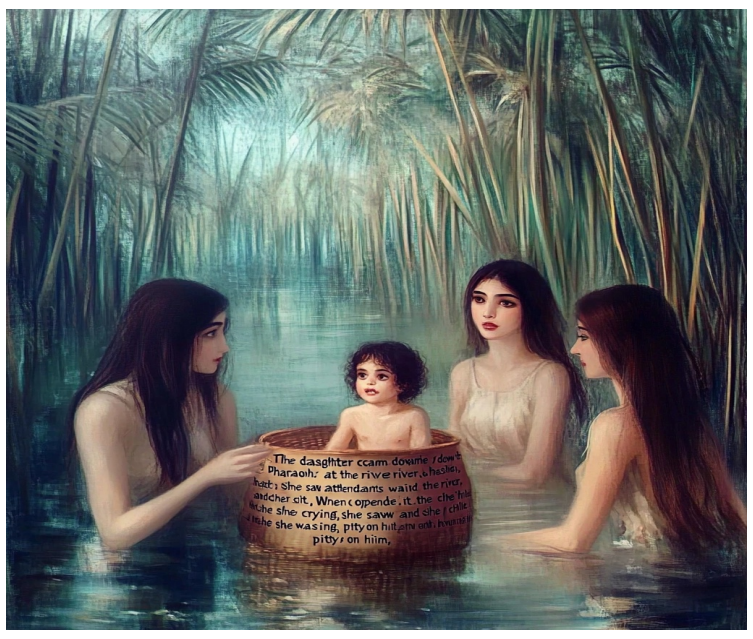 | ☑ Pharaoh’s Daughter Coming to the River<br>☑ The Basket in the Reeds<br>☑ The Maidens/Servants<br>☑ The Opening of the Basket<br>☐ The Child Crying<br>☐ Dialogue and Interaction<br>☑ The River and Reeds as Setting | Prompt: NSRV, 700 scale weirdness<br>Evaluation: The style is reminiscent of Symbolism. A dreamy, painterly depiction with muted tones and ethereal lighting. The reeds and soft textures evoke intimacy, though the figures appear idealized rather than culturally specific. See also Fig 10. |
| 18. | 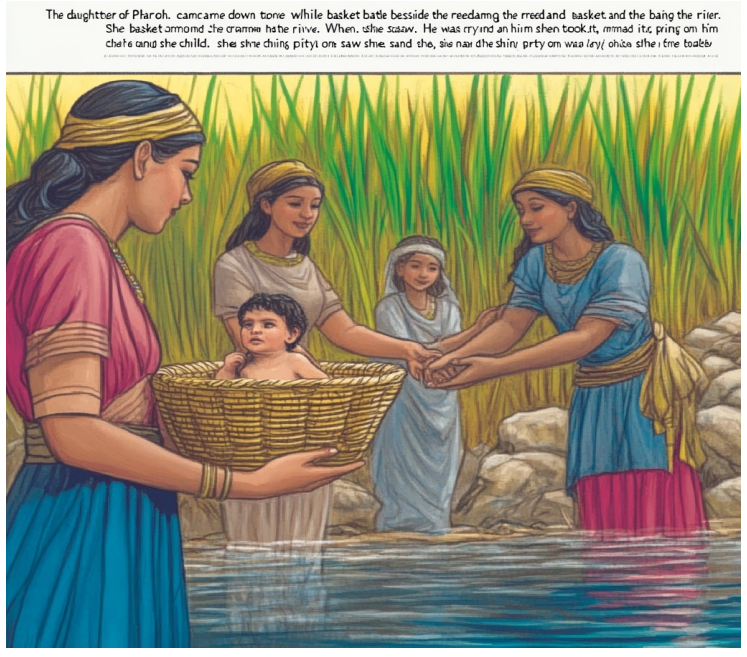 | ☑ Pharaoh’s Daughter Coming to the River<br>☑ The Basket in the Reeds<br>☑ The Maidens/Servants<br>☐ The Opening of the Basket<br>☐ The Child Crying<br>☑ Dialogue and Interaction<br>☐ The River and Reeds as Setting | Prompt: NSRV<br>Evaluation: A vibrant depiction of Pharaoh's daughter and her attendants surrounding Moses in the basket. The lush reeds and colorful attire create a lively atmosphere, though the modern styling softens the historical setting. |
| 19. | 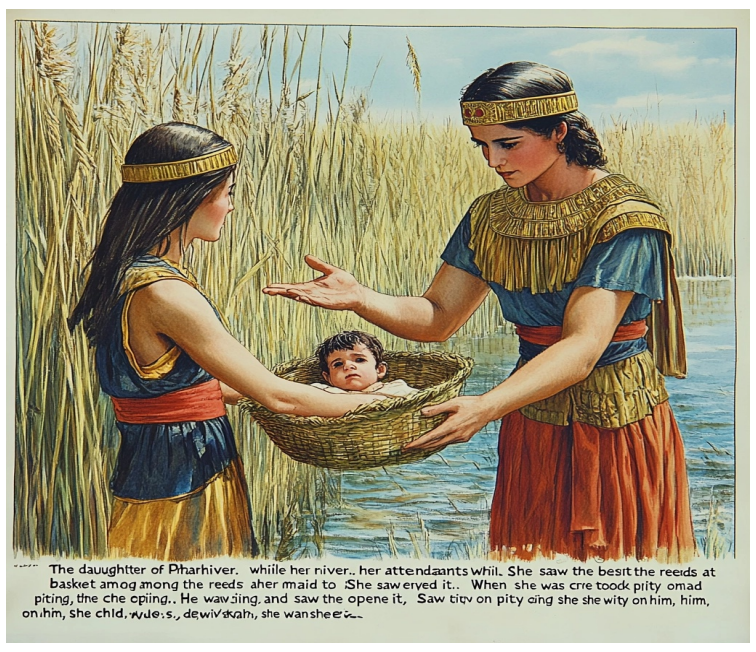 | ☐ Pharaoh’s Daughter Coming to the River<br>☐ The Basket in the Reeds<br>☐ The Maidens/Servants<br>☑ The Opening of the Basket<br>☐ The Child Crying<br>☑ Dialogue and Interaction | Prompt: NSRV<br>Evaluation: Pharaoh’s daughter is shown passing Moses in a basket to a companion. The golden accents and ornate clothing highlight status and opulence, while the composition |
| | | ☑ The River and Reeds as Setting | captures the emotional exchange with tenderness. |
| 20. | 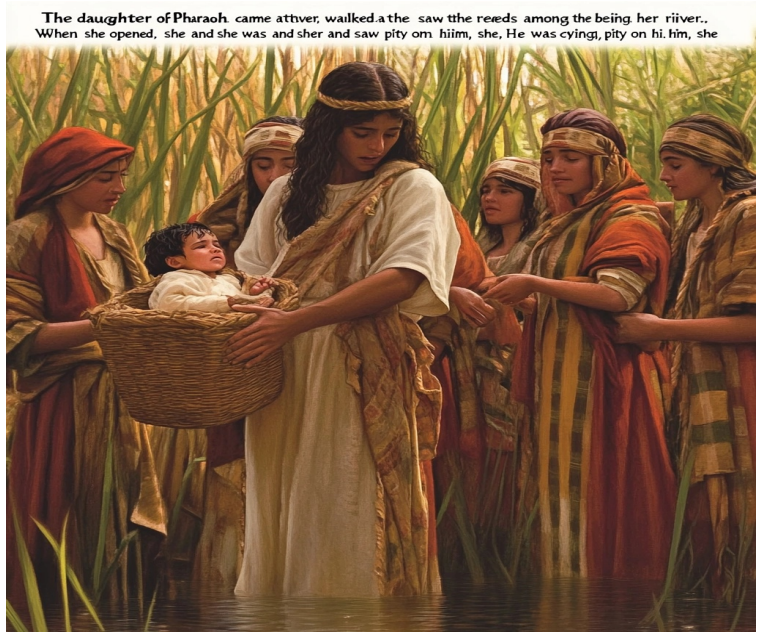 | ☑ Pharaoh's Daughter Coming to the River<br>☑ The Basket in the Reeds<br>☑ The Maidens/Servants<br>☐ The Opening of the Basket<br>☑ The Child Crying<br>☐ Dialogue and Interaction<br>☑ The River and Reeds as Setting | Prompt: NSRV<br>Evaluation: A muted and solemn scene, with Pharaoh's daughter cradling Moses surrounded by richly dressed attendants. The subdued lighting emphasizes a reflective and sacred mood, while the intricate fabrics add historical depth. |
| 21. | 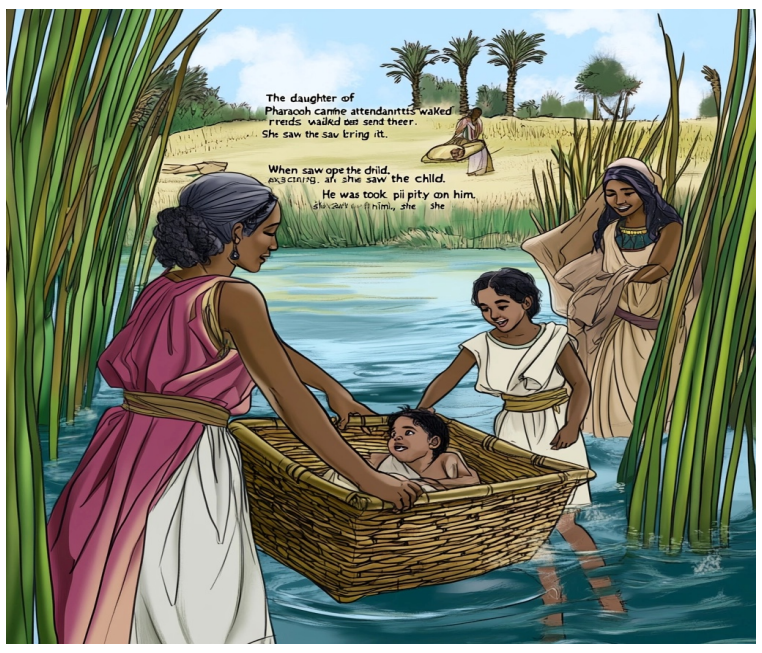 | ☑ Pharaoh's Daughter Coming to the River<br>☑ The Basket in the Reeds<br>☐ The Maidens/Servants<br>☐ The Opening of the Basket<br>☑ The Child Crying<br>☑ Dialogue and Interaction<br>☑ The River and Reeds as Setting | Prompt: NSRV<br>Evaluation: A bright and playful composition featuring Moses in a basket. The surrounding figures exhibit joy and nurturing care, while the vibrant reeds and distant palms give the scene a pastoral and idyllic quality. |
| 22. | 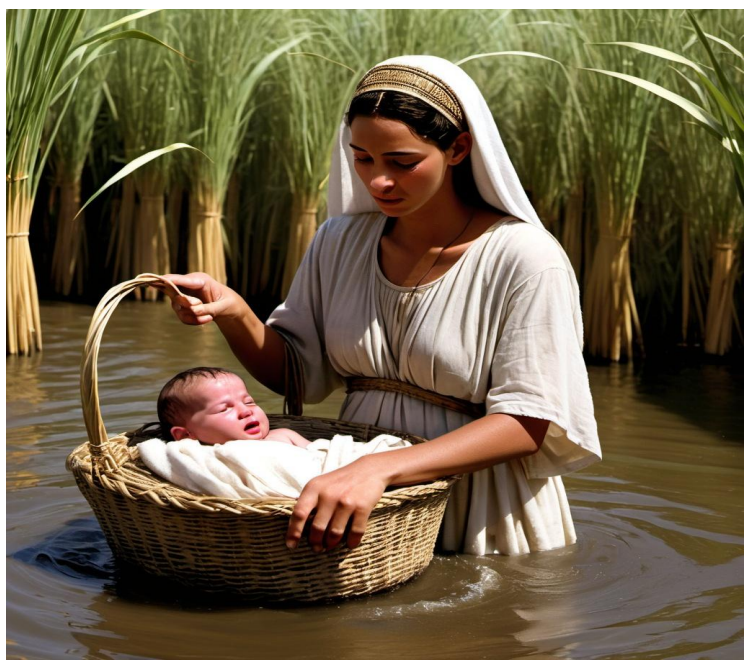 | ☐ Pharaoh's Daughter Coming to the River<br>☑ The Basket in the Reeds<br>☐ The Maidens/Servants<br>☑ The Opening of the Basket<br>☑ The Child Crying<br>☐ Dialogue and Interaction<br>☑ The River and Reeds as Setting | Prompt: NRSV<br>Evaluation: Photorealistic depiction captures a woman in flowing ancient-style garments standing in a river, holding a wicker basket with a baby, surrounded by lush reeds. The setting evokes the Nile River, aligning with the story of Moses' rescue. Though the attire and basket reflect modern interpretations, the scene's emotional depth and symbolism resonate strongly. |
| 23. | 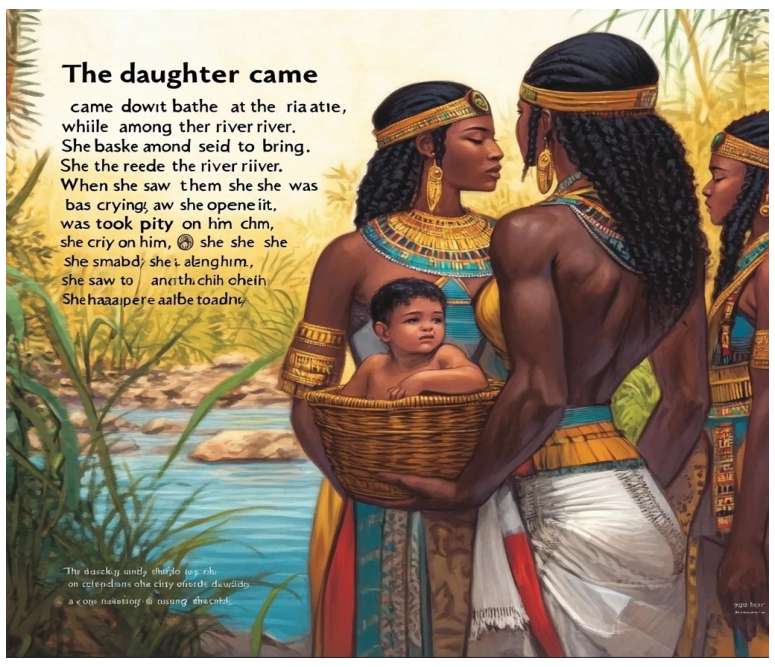 | ☑ Pharaoh's Daughter Coming to the River<br>☑ The Basket in the Reeds<br>☑ The Maidens/Servants<br>☑ The Opening of the Basket<br>☐ The Child Crying<br>☐ Dialogue and Interaction<br>☑ The River and Reeds as Setting | Prompt: NSRV, 3000 scale weirdness<br>Evaluation: A vibrant, semi-realistic depiction of a dark-skinned Pharaoh's daughter holding Moses in a basket, with attendants in the background. A richly detailed image with Egyptian-inspired jewelry and textiles, and hierarchical composition evokes a sense of cultural and historical context. |
| 24. | 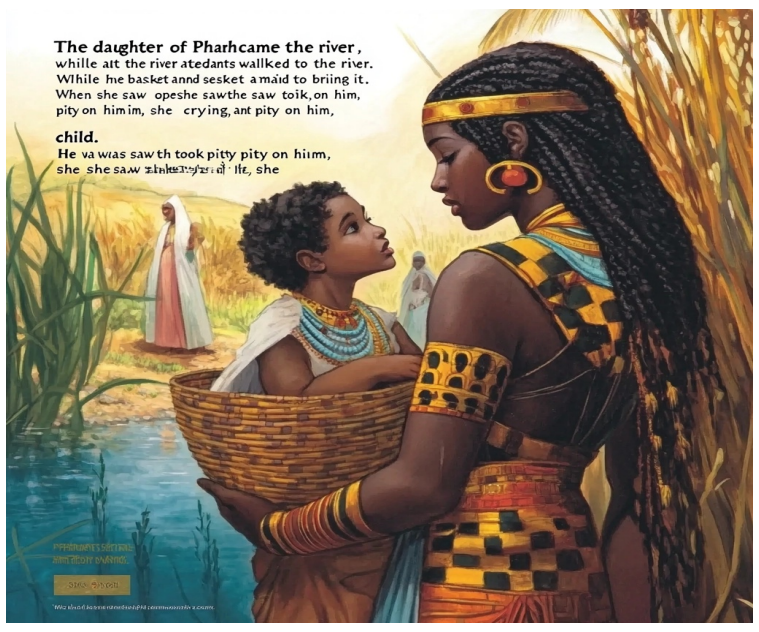 | ☑ Pharaoh's Daughter Coming to the River<br>☐ The Basket in the Reeds<br>☐ The Maidens/Servants<br>☐ The Opening of the Basket<br>☐ The Child Crying<br>☐ Dialogue and Interaction<br>☑ The River and Reeds as Setting | Prompt: NSRV, 3000 scale weirdness, rerun<br>Evaluation: A bold and colorful depiction with traditional African-inspired patterns and adornments. The vibrant palette highlights cultural richness, though it reimagines the story in a more symbolic than historical manner. |
| 25. | 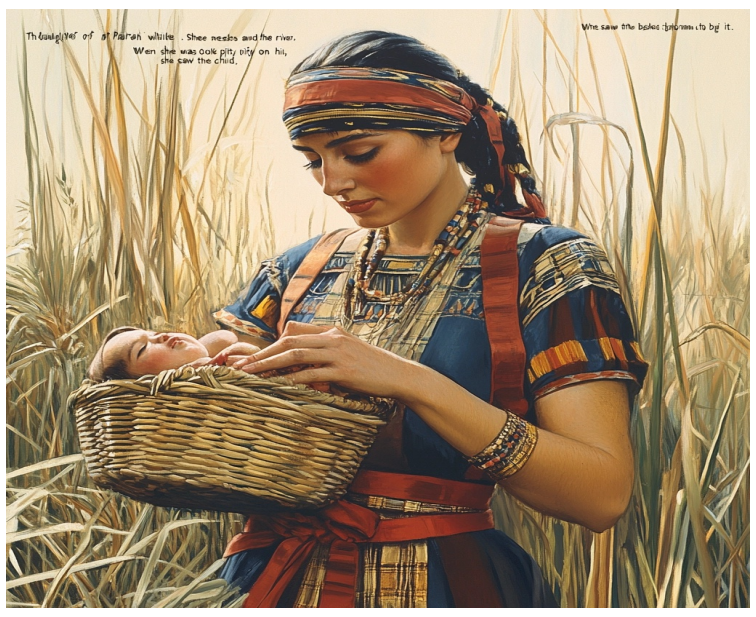 | ☐ Pharaoh's Daughter Coming to the River<br>☑ The Basket in the Reeds<br>☐ The Maidens/Servants<br>☑ The Opening of the Basket<br>☐ The Child Crying<br>☐ Dialogue and Interaction<br>☐ The River and Reeds as Setting | Prompt: Moses found in Dali style.<br>Observation: Although this image does not reflect Dali's art style it is a subdued, intimate moment of a motherly figure with a child in a basket, surrounded by reeds. The earthy tones and painterly texture create a timeless, reflective mood. |
| 26. | 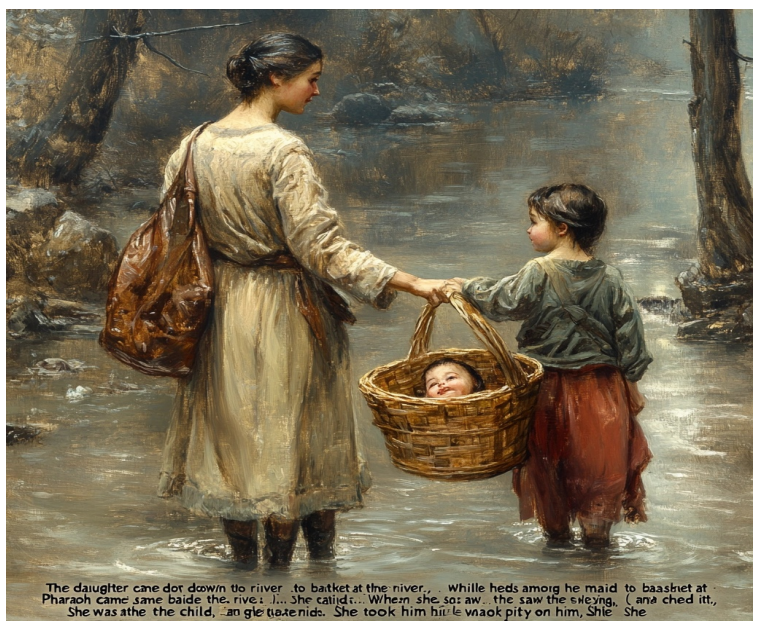 | ☐ Pharaoh's Daughter Coming to the River<br>☐ The Basket in the Reeds<br>☐ The Maidens/Servants<br>☐ The Opening of the Basket<br>☐ The Child Crying<br>☑ Dialogue and Interaction<br>☑ The River and Reeds as Setting | Prompt: Moses found in Dali style.<br>Evaluation: A painterly scene of two people (a woman and a child) and Moses in a basket, rendered with atmospheric lighting and soft textures. The reeds and water blend seamlessly with the figures, |
| | | | evoking a dreamlike quality of care and discovery.<br>The image, however, does not look like a Dali style |
| 27. | 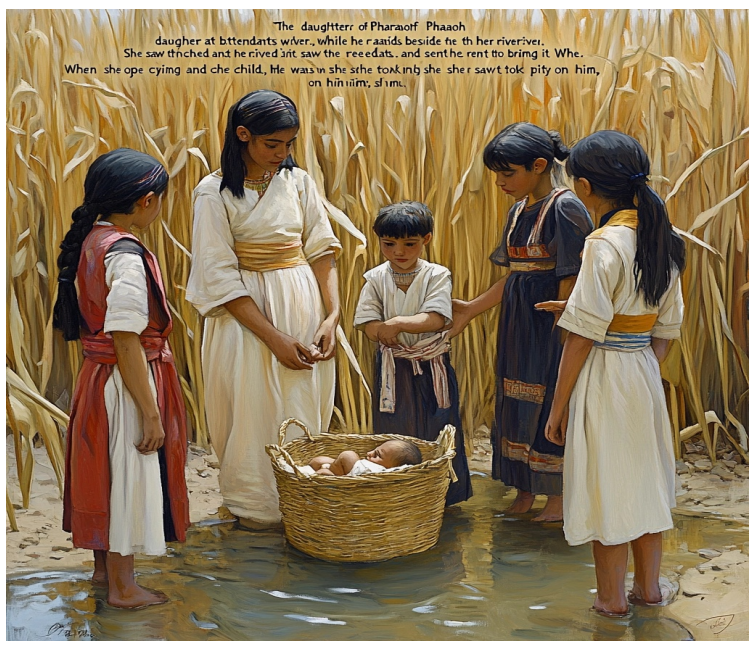 | ☐ Pharaoh's Daughter Coming to the River<br>☑ The Basket in the Reeds<br>☐ The Maidens/Servants<br>☑ The Opening of the Basket<br>☐ The Child Crying<br>☐ Dialogue and Interaction<br>☑ The River and Reeds as Setting | Prompt: Moses found in Dali style.<br>Evaluation: The image does not look like a Dali style but presents a warm, community-focused depiction of discovery, featuring a group of women or children in flowing garments surrounding a basket. The realistic reeds and subtle expressions emphasize a shared moment of care and curiosity. |
| 28. | 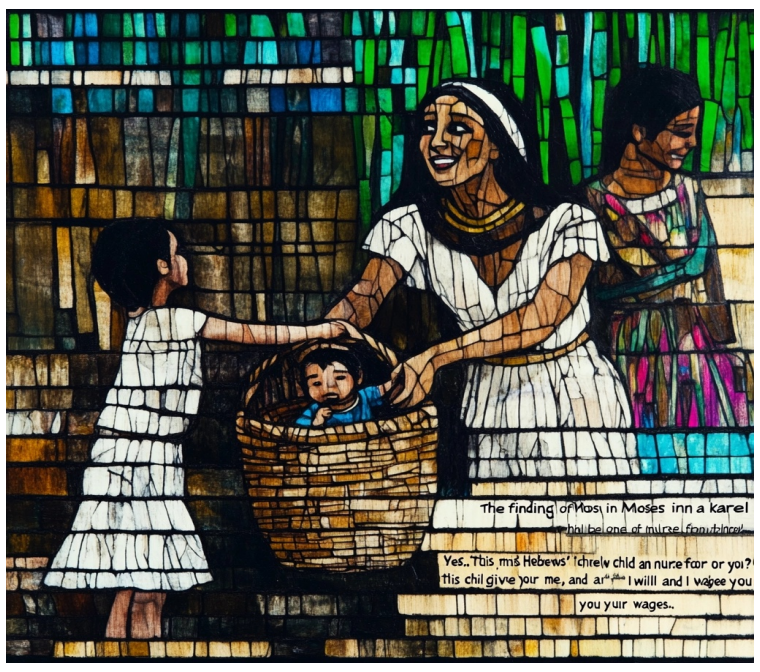 | ☐ Pharaoh's Daughter Coming to the River<br>☐ The Basket in the Reeds<br>☐ The Maidens/Servants<br>☑ The Opening of the Basket<br>☐ The Child Crying<br>☑ Dialogue and Interaction<br>☐ The River and Reeds as Setting | Prompt: Moses found in Karel Appel style.<br>Evaluation: This image mimics a stained-glass aesthetic, giving a mosaic-like structure to the biblical scene. The vibrant colors and segmented design evoke religious art found in churches, reinforcing a sacred tone. The joyful expressions contrast with the gravity of the moment, suggesting a celebratory rather than dramatic interpretation. |
| 29. | 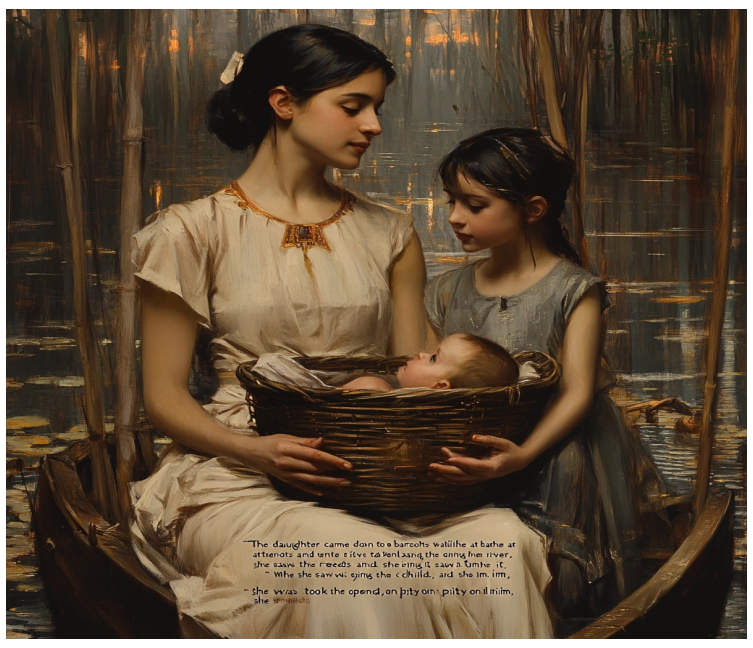 | ☐ Pharaoh's Daughter Coming to the River<br>☑ The Basket in the Reeds<br>☐ The Maidens/Servants<br>☑ The Opening of the Basket<br>☐ The Child Crying<br>☐ Dialogue and Interaction<br>☑The River and Reeds as Setting | Prompt: Moses found in Karel Appel style.<br>Evaluation: This image does not look like Karel Appel's style of art. However, here is a soft, naturalistic portrayal of a woman with a girl beside her, and a child in a basket by a serene river. The delicate brushstrokes and muted tones convey an intimate, peaceful moment, but the historical |
| | | | elements are less pronounced. |
| 30. | 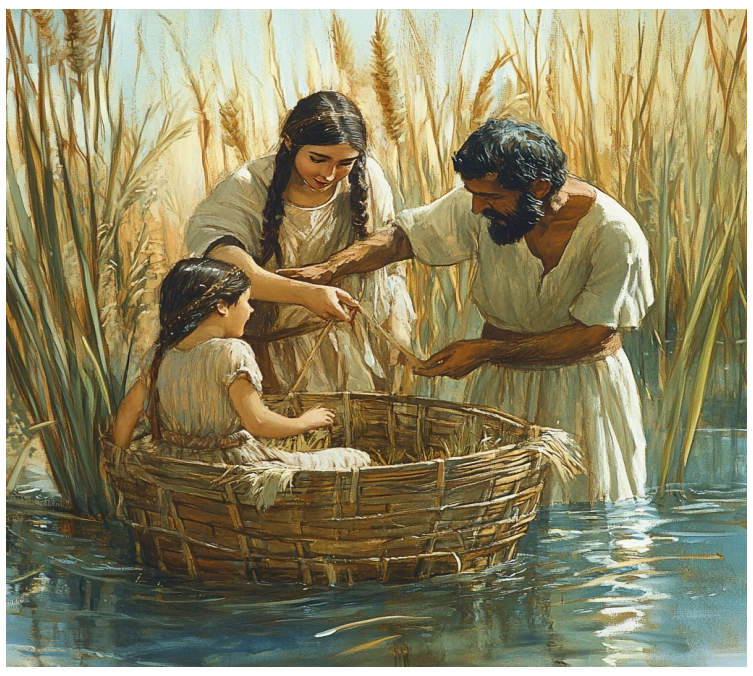 | ☐ Pharaoh's Daughter Coming to the River<br>☑ The Basket in the Reeds<br>☐ The Maidens/Servants<br>☐ The Opening of the Basket<br>☐ The Child Crying<br>☐ Dialogue and Interaction<br>☑ The River and Reeds as Setting | Prompt: using the visual elements<br>Evaluation: A familial moment depicted in a pastoral setting, with warm lighting emphasizing intimacy and care. The figures, likely depicting Moses' biological family, prepare the basket with visible tenderness, blending cultural authenticity with emotional resonance. The is a man in this image. |
| 31. | 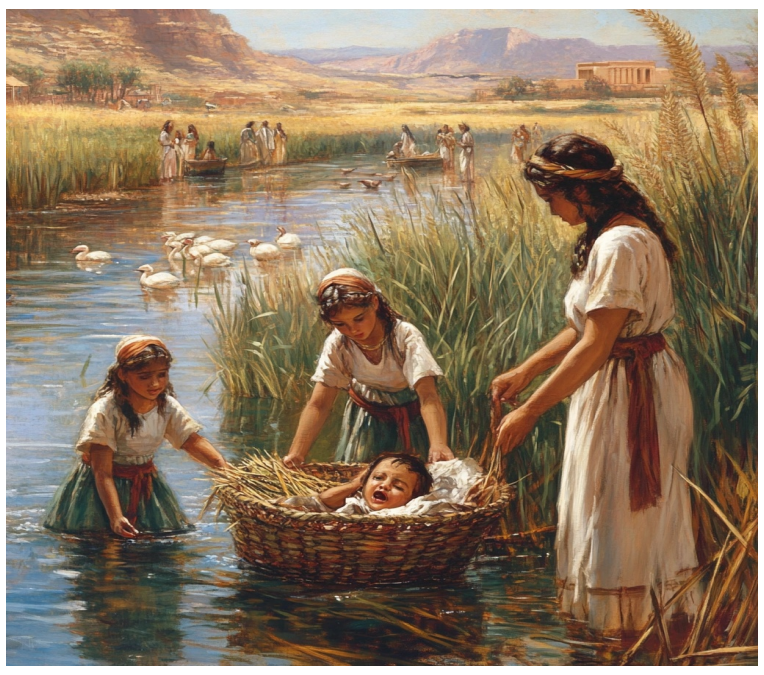 | ☐ Pharaoh's Daughter Coming to the River<br>☑ The Basket in the Reeds<br>☐ The Maidens/Servants<br>☐ The Opening of the Basket<br>☑ The Child Crying<br>☐ Dialogue and Interaction<br>☑ The River and Reeds as Setting | Prompt: using the visual elements<br>Evaluation: A tranquil landscape with vivid greenery and soft lighting, where women gently retrieve Moses' basket. The detailed environment enhances the realism and emotional warmth. |
| 32. | 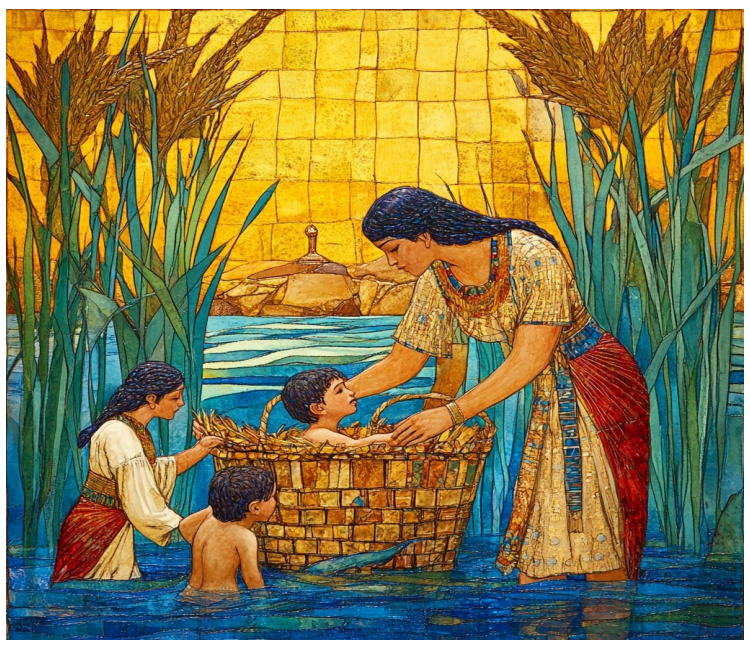 | ☑ Pharaoh's Daughter Coming to the River<br>☑ The Basket in the Reeds<br>☐ The Maidens/Servants<br>☐ The Opening of the Basket<br>☐ The Child Crying<br>☐ Dialogue and Interaction<br>☑ The River and Reeds as Setting | Prompt: using the visual elements.<br>Evaluation: A colorful, stylized rendering with a golden background reminiscent of stained glass. The geometric arrangement and vivid tones offer a symbolic, sacred interpretation of the story. |
| 33. | 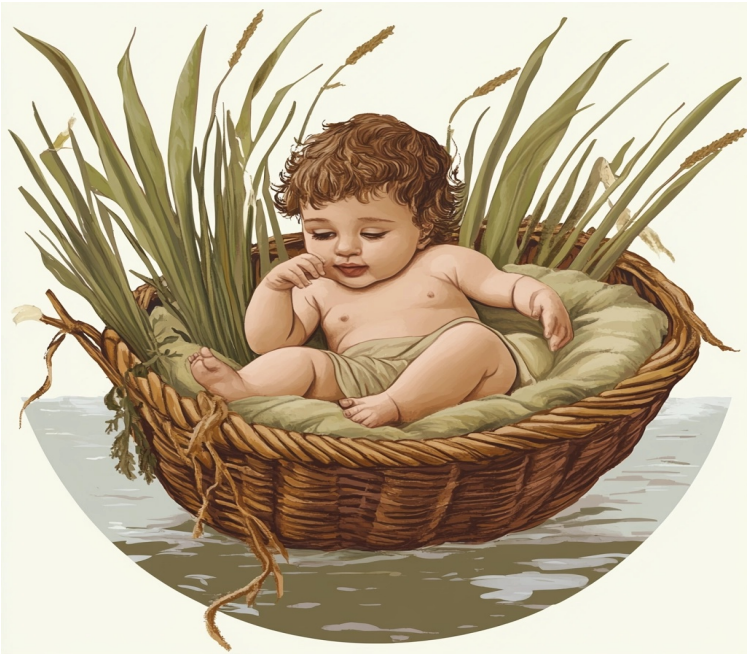 | ☐ Pharaoh's Daughter Coming to the River<br>☑ The Basket in the Reeds<br>☐ The Maidens/Servants<br>☐ The Opening of the Basket<br>☐ The Child Crying<br>☐ Dialogue and Interaction<br>☑ The River and Reeds as Setting | Prompt: Images of baby Moses found according to Exodus 2:5-9<br>Evaluation: It doesn't fit neatly into historical art movements like the Renaissance (focused on realism, anatomy, and classical themes) or Baroque |
| | | | (dramatic, emotional, and richly detailed).<br>The style leans toward contemporary illustration, possibly influenced by children's book art or digital art trends. |
| 34. | 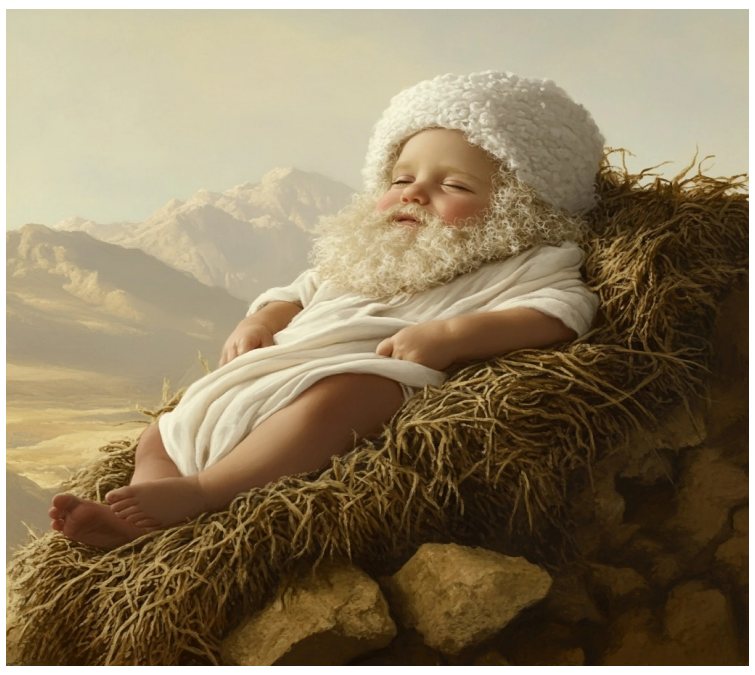 | ☐ Pharaoh's Daughter Coming to the River<br>☐ The Basket in the Reeds<br>☐ The Maidens/Servants<br>☐ The Opening of the Basket<br>☐ The Child Crying<br>☐ Dialogue and Interaction<br>☐ The River and Reeds as Setting | Prompt: Images of baby Moses found according to Exodus 2:5-9<br>Evaluation: This image portrays a baby lying in a naturalistic setting of straw and rocks, against a backdrop of mountains. The child is dressed in white fabric and features an exaggerated, humorous addition: a white curly beard and a woolly hat. |
| 35. | 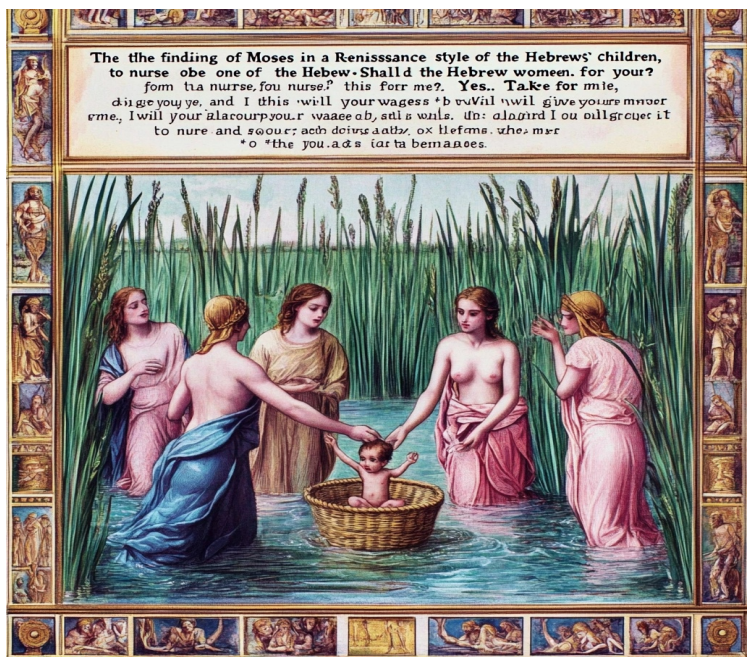 | ☐ Pharaoh's Daughter Coming to the River<br>☑ The Basket in the Reeds<br>☑ The Maidens/Servants<br>☑ The Opening of the Basket<br>☐ The Child Crying<br>☑ Dialogue and Interaction<br>☑ The River and Reeds as Setting | Prompt: the finding of Moses in a Renaissance style" + NRSV<br>Evaluation: A narrative-driven scene framed with intricate details and bordered panels. Figures in classical robes, partly nude interact with Moses, emphasizing storytelling through composition and layout. |
| 36. | 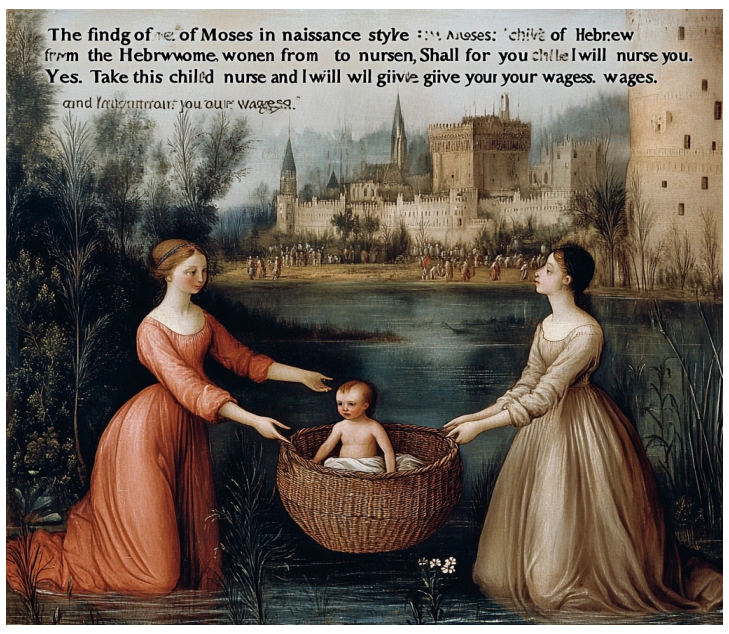 | ☐ Pharaoh's Daughter Coming to the River<br>☐ The Basket in the Reeds<br>☐ The Maidens/Servants<br>☑ The Opening of the Basket<br>☐ The Child Crying<br>☑ Dialogue and Interaction<br>☐ The River and Reeds as Setting | Prompt: the finding of Moses in a Renaissance style" + NRSV<br>Evaluation: A medieval European reinterpretation of Moses' story, featuring castle-like architecture in the background. The figures' poses and richly textured attire evoke an ethereal quality, but the historical accuracy of the setting is replaced by a Western aesthetic. |
| 37. | 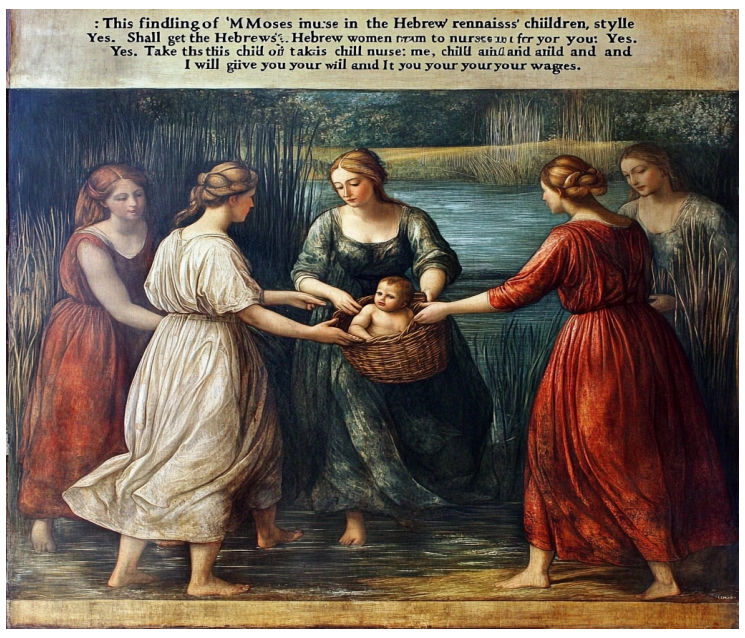 | ☐ Pharaoh's Daughter Coming to the River<br>☐ The Basket in the Reeds<br>☑ The Maidens/Servants<br>☐ The Opening of the Basket<br>☑ The Child Crying<br>☑ Dialogue and Interaction<br>☑ The River and Reeds as Setting | Prompt: the finding of Moses in a Renaissance style" + NRSV<br>Observations: A Renaissance-inspired scene with figures in flowing garments, reflecting classical composition and balance. The use of symmetry and detailed drapery elevates the sense of nobility and reverence, while the serene expressions and natural setting create a harmonious depiction of Moses' discovery. |
| 38. | 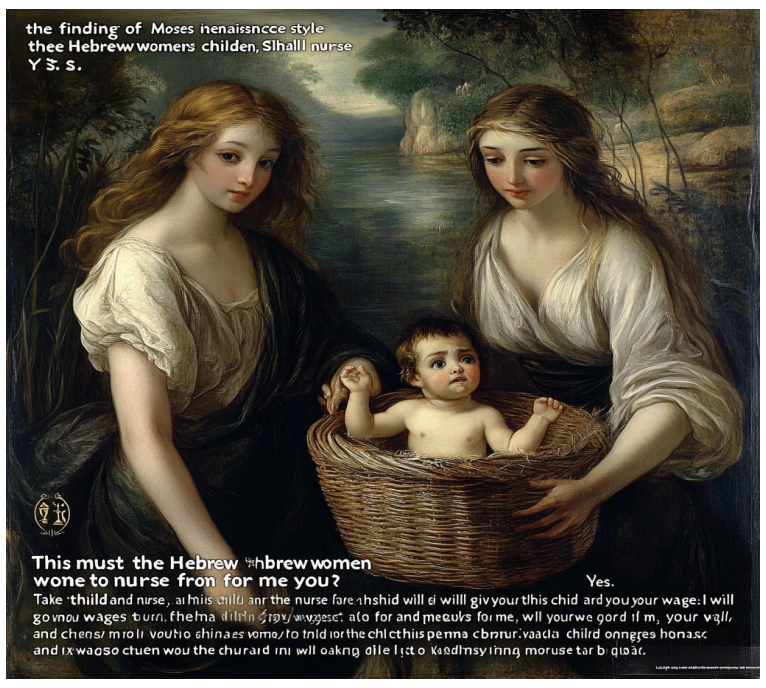 | ☐ Pharaoh's Daughter Coming to the River<br>☐ The Basket in the Reeds<br>☐ The Maidens/Servants<br>☑ The Opening of the Basket<br>☑ The Child Crying<br>☐ Dialogue and Interaction<br>☑ The River and Reeds as Setting | Prompt: the finding of Moses in a Renaissance style" + NRSV<br>Observations: A soft, pastoral depiction featuring elegant women and a baby in a serene river scene. The classical poses and muted palette reflect Renaissance sensibilities, focusing on beauty and grace. |
| 39. | 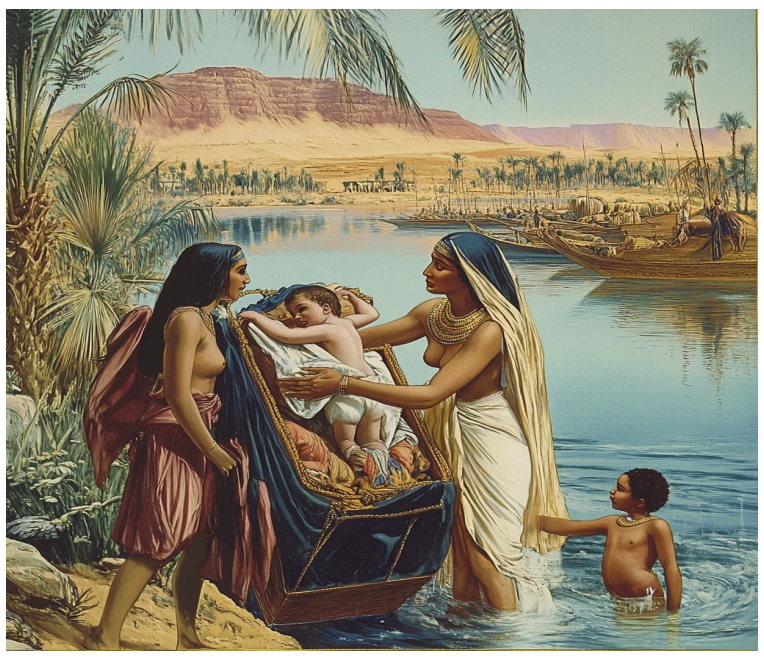 | ☑ Pharaoh's Daughter Coming to the River<br>☐ The Basket in the Reeds<br>☑ The Maidens/Servants<br>☐ The Opening of the Basket<br>☐ The Child Crying<br>☑ Dialogue and Interaction<br>☑ The River and Reeds as Setting | Prompt: from an African point of view<br>Evaluation: This image presents a romanticized take on Moses' discovery, with rich colors, ornate details, and a lush Nile backdrop. It idealizes Pharaoh's daughter and embellishes the basket, making it more regal than historically accurate. The scene prioritizes visual grandeur over realism, enhancing its dramatic storytelling. |
| 40. | 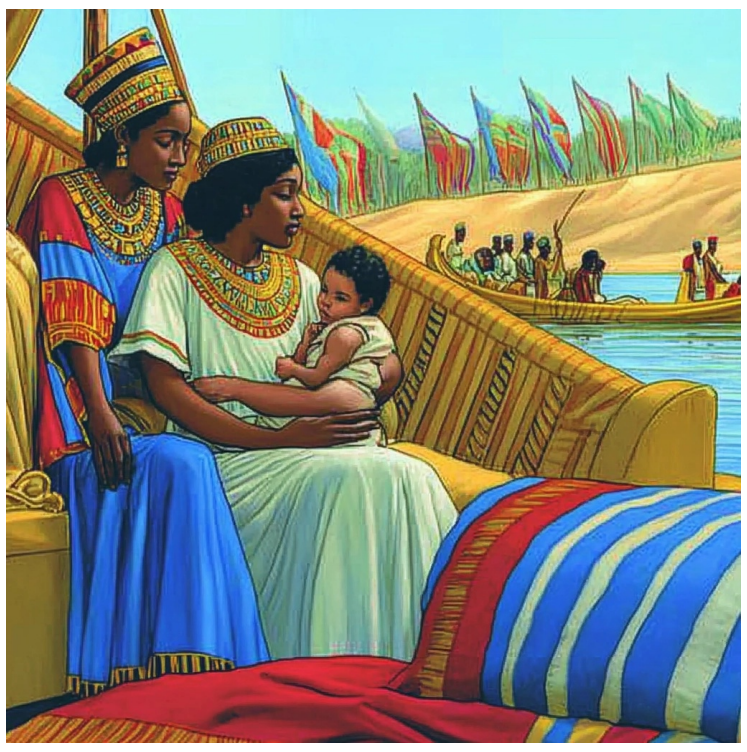 | ☑ Pharaoh’s Daughter Coming to the River<br>☐ The Basket in the Reeds<br>☐ The Maidens/Servants<br>☐ The Opening of the Basket<br>☐ The Child Crying<br>☐ Dialogue and Interaction<br>☐ The River and Reeds as Setting | Prompt: from an African point of view [stylize: 550, weirdness: 2900, chaos 40, style: raw]<br>Evaluation: This image presents an idealized depiction of Pharaoh’s court, with richly adorned figures in traditional Egyptian attire. The baby Moses is cradled with care, emphasizing a sense of protection and royalty. The vibrant flags and distant boats add a ceremonial atmosphere, suggesting a grand adoption scene. However, the stylization leans towards modernized historical fiction rather than an authentic ancient Egyptian portrayal. |
| 41. | 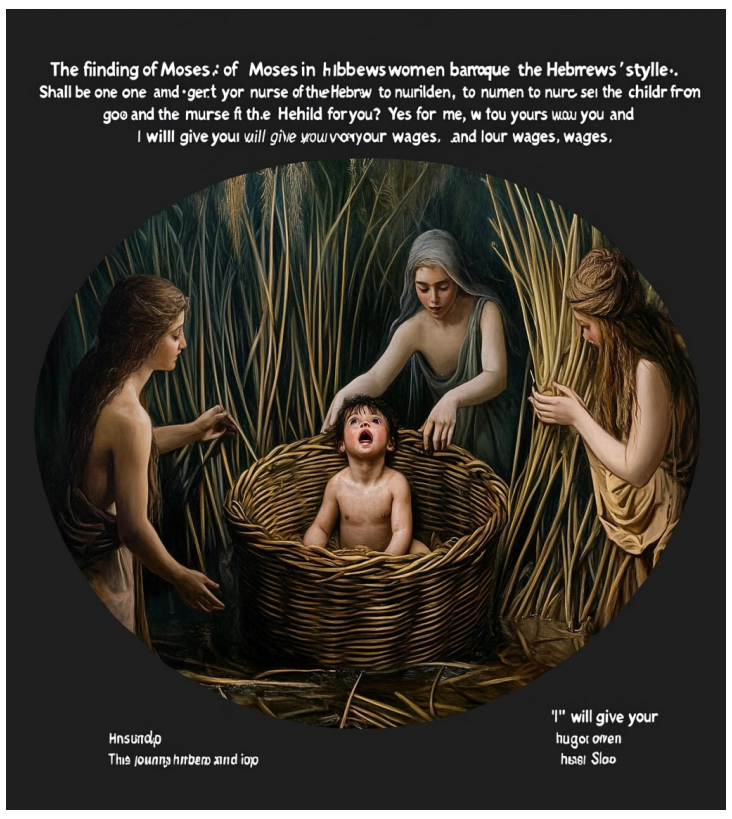 | ☐ Pharaoh’s Daughter Coming to the River<br>☑ The Basket in the Reeds<br>☐ The Maidens/Servants<br>☑ The Opening of the Basket<br>☑ The Child Crying<br>☐ Dialogue and Interaction<br>☑ The River and Reeds as Setting | Prompt: the finding of Moses in Baroque style<br>Evaluation: This image presents a dark, moody interpretation of the finding of Moses, heavily influenced by the baroque style. The chiaroscuro lighting adds depth and an almost sacred atmosphere, emphasizing the emotional intensity of the moment. The three ethereal women, reminiscent of classical depictions of nymphs, appear solemn as they gaze at the vulnerable child, who looks upwards in a desperate, almost divine plea. The oversized basket and intricate reeds enhance the mythical quality of the composition. |
| 42. | 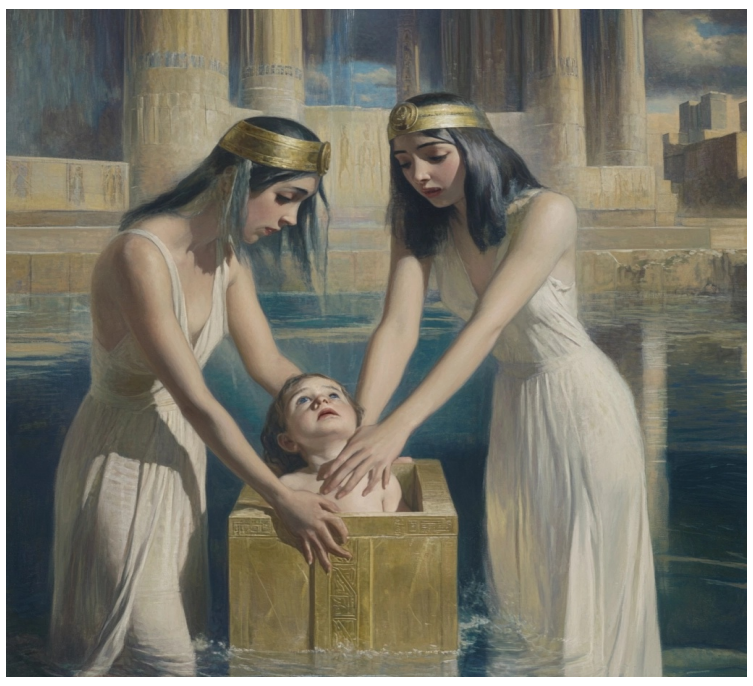 | ☐ Pharaoh's Daughter Coming to the River<br>☐ The Basket in the Reeds<br>☐ The Maidens/Servants<br>☑ The Opening of the Basket<br>☐ The Child Crying<br>☐ Dialogue and Interaction<br>☑ The River and Reeds as Setting | Prompt: symbolism style<br>Evaluation: This image portrays an elegant, ethereal take on the finding of Moses. Two women in flowing white dresses and gold headbands, resembling Egyptian royalty, lift a child from a golden box. The grand architectural backdrop and reflective water create a mystical atmosphere. The scene emphasizes divine intervention, with a focus on idealized beauty rather than strict historical accuracy. |
| 43. | 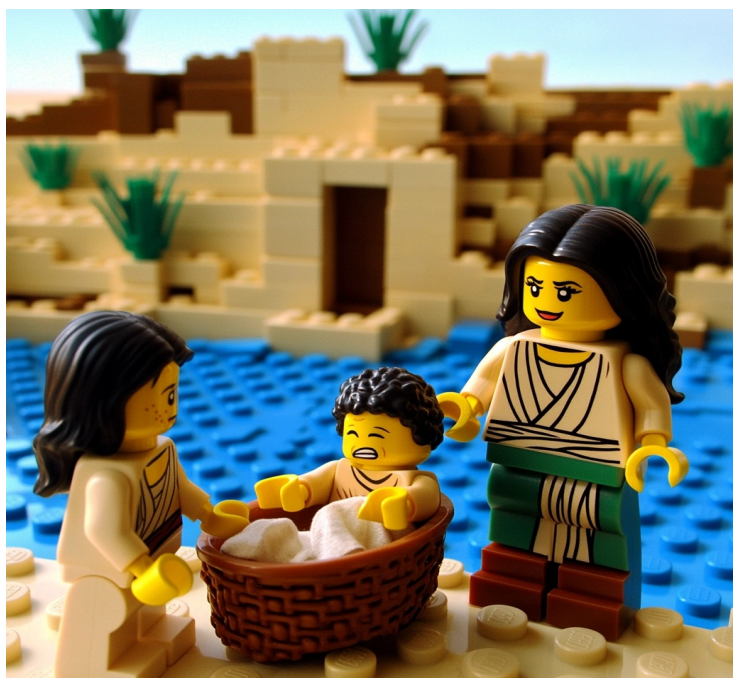 | ☑ Pharaoh's Daughter Coming to the River<br>☐ The Basket in the Reeds<br>☐ The Maidens/Servants<br>☑ The Opening of the Basket<br>☑ The Child Crying<br>☐ Dialogue and Interaction<br>☑ The River and Reeds as Setting | Prompt: using LEGO style<br>Evaluation: This image reimagines the biblical scene of Moses being found using LEGO-style figures. The playful, blocky aesthetic contrasts with the traditionally dramatic depictions of the story. Bright colors and expressive faces make it engaging, though it simplifies historical and cultural accuracy. The creative medium adds a lighthearted and accessible interpretation to a well-known narrative. |
| 44. | 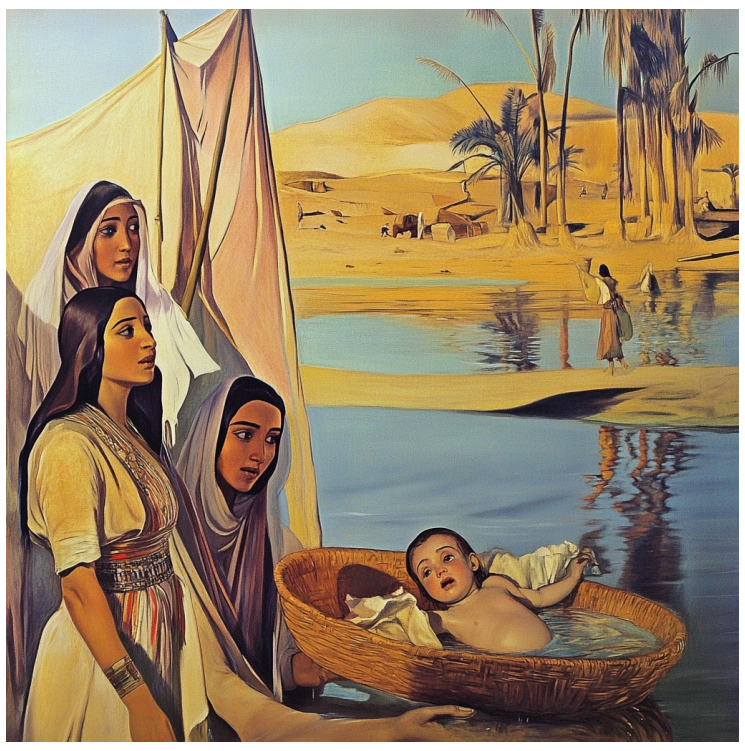 | ☐ Pharaoh's Daughter Coming to the River<br>☐ The Basket in the Reeds<br>☐ The Maidens/Servants<br>☑ The Opening of the Basket<br>☑ The Child Crying<br>☐ Dialogue and Interaction<br>☑ The River and Reeds as Setting | Prompt: in Deli Style<br>Evaluation: This image presents a stylized, painterly depiction. The warm, earthy color palette enhances the desert setting, while the figures are elegantly posed, exuding grace and reverence. The soft, flowing garments and serene expressions emphasize a sense of wonder and divinity. The composition directs attention toward the baby in the basket, while the |
| | | | background details, such as distant figures and reflections in the water, add depth. |
| 45. | 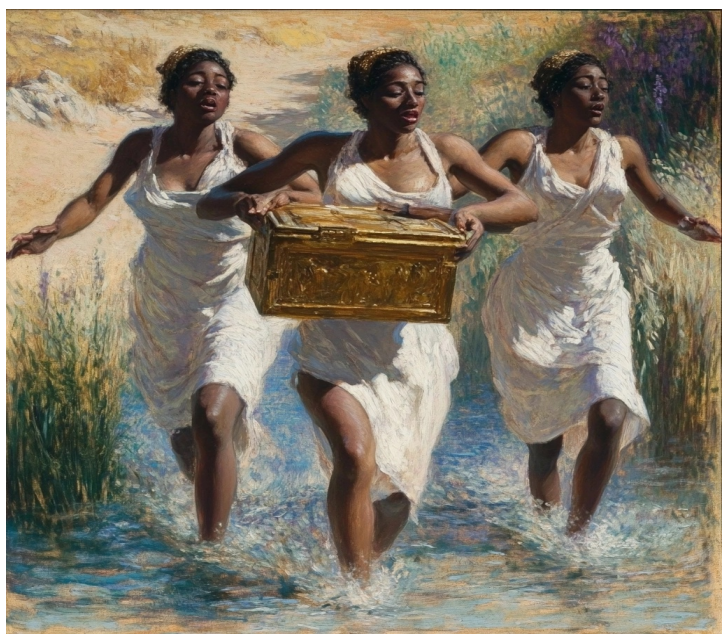 | ☐ Pharaoh's Daughter Coming to the River<br>☐ The Basket in the Reeds<br>☐ The Maidens/Servants<br>☐ The Opening of the Basket<br>☐ The Child Crying<br>☐ Dialogue and Interaction<br>☑ The River and Reeds as Setting | Prompt: from an African point of view [stylize: 550, weirdness: 2900, chaos 40, style: raw]<br>Evaluation: Featuring three women rushing through water carrying a golden chest, this image introduces an unusual interpretation of the event. The motion is dynamic, but the golden box replacing Moses' basket raises questions about accuracy. The expressive faces capture urgency but deviate from the expected biblical imagery. |
| 46. | 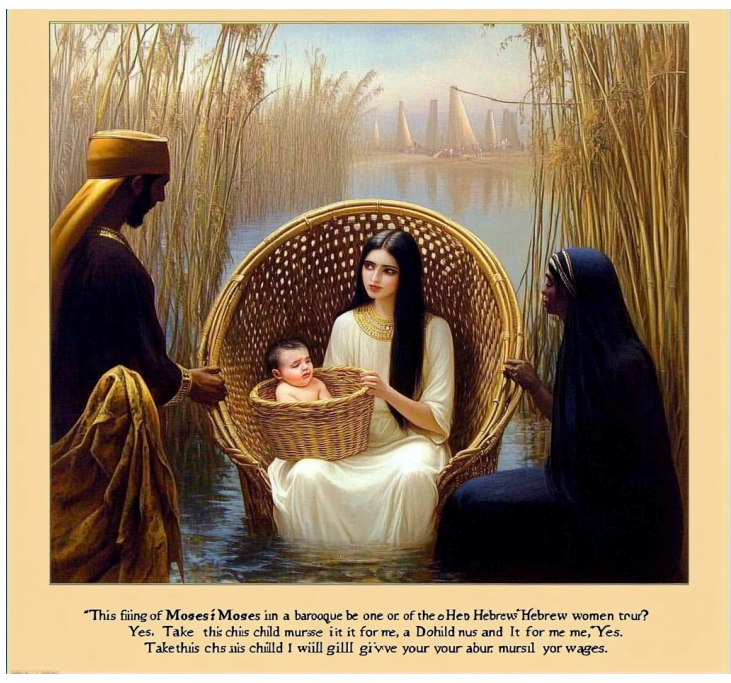 | ☑ Pharaoh's Daughter Coming to the River<br>☐ The Basket in the Reeds<br>☑ The Maidens/Servants<br>☑ The Opening of the Basket<br>☐ The Child Crying<br>☑ Dialogue and Interaction<br>☑ The River and Reeds as Setting | Prompt: the finding of Moses in Baroque style<br>Evaluation: This image blends realism with a mystical atmosphere. The central figure, possibly Pharaoh's daughter, is ethereal, with an almost divine presence, holding baby Moses in a woven basket. The intricate details of clothing and jewelry evoke a sense of wealth and nobility. The warm, golden lighting contrasts with the dark-clad figures, creating a visual hierarchy of power and servitude. |
| 47. | 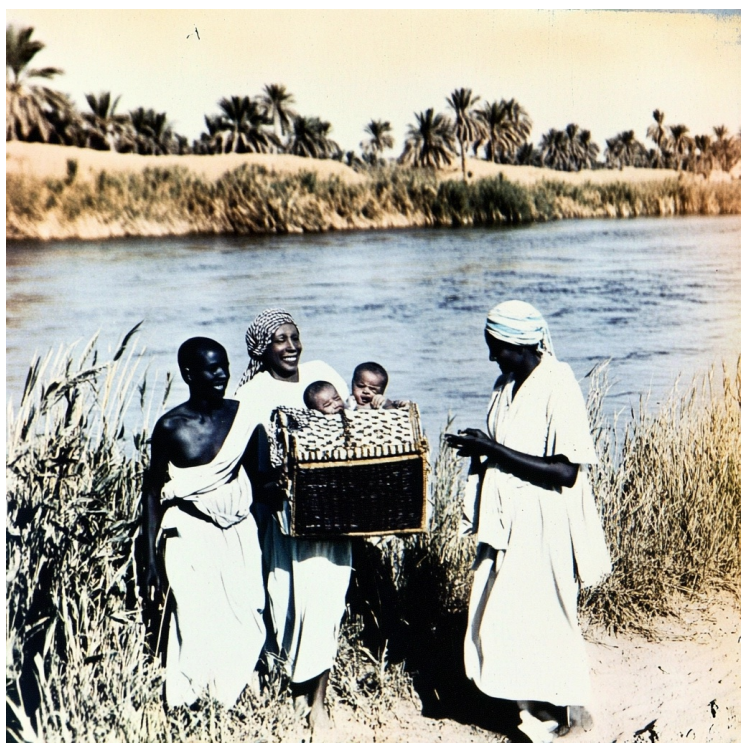 | ☐ Pharaoh's Daughter Coming to the River<br>☑ The Basket in the Reeds<br>☐ The Maidens/Servants<br>☑ The Opening of the Basket<br>☐ The Child Crying<br>☐ Dialogue and Interaction<br>☑ The River and Reeds as Setting | Prompt: in surrealism style + increased weirdness<br>Evaluation: This image mimics an aged, historical photograph, portraying a scene along the Nile where a group of traditionally dressed individuals joyfully carries a basket containing two infants. The sepia-toned, slightly faded effect adds an archival authenticity, evoking early ethnographic photography. The relaxed, smiling expressions contrast with the traditionally dramatic depictions of Moses' discovery, shifting the focus from distress to communal warmth. However, the presence of two babies in the basket deviates from the biblical narrative, introducing an unintended alteration to the story's core symbolism. |
| 48. | 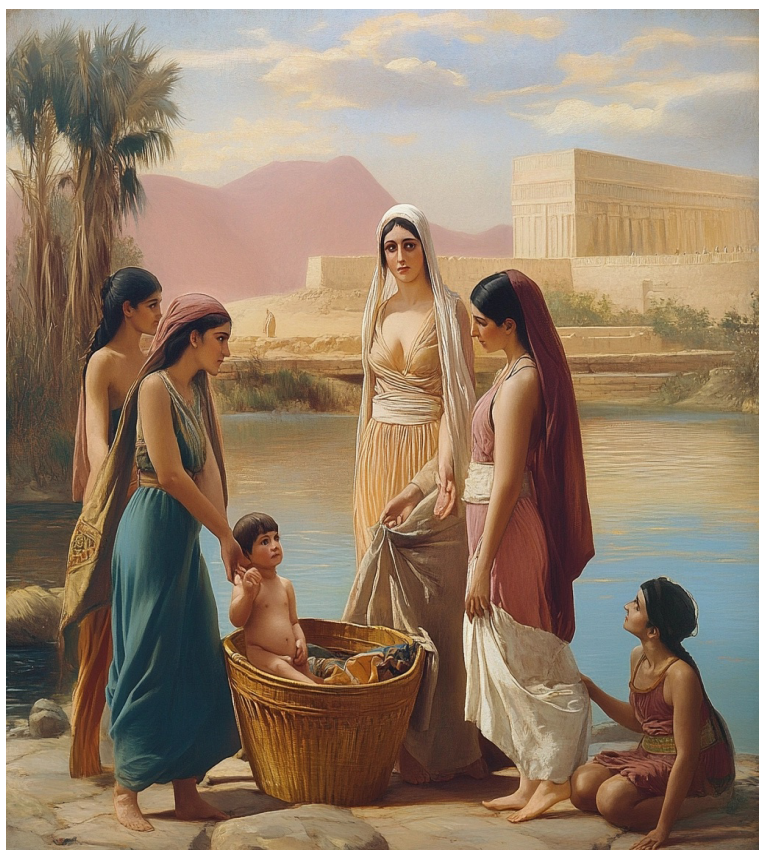 | ☑Pharaoh's Daughter Coming to the River<br>☐ The Basket in the Reeds<br>☐ The Maidens/Servants<br>☑ The Opening of the Basket<br>☐ The Child Crying<br>☑ Dialogue and Interaction<br>☐ The River and Reeds as Setting | Prompt: in a Michaelangelo style<br>Evaluation: This image portrays the discovery of baby Moses with a neoclassical aesthetic, featuring soft, realistic lighting and carefully composed figures. The women's flowing garments and serene expressions evoke a sense of reverence and awe. The architectural backdrop enhances the historical atmosphere, grounding the narrative in ancient Egypt. The baby's posture and engagement with the figures add emotional depth, though the idealized features and polished composition lean towards romanticized |
| | | | storytelling rather than strict historical accuracy. The warm color palette and delicate details create a visually harmonious and compelling interpretation of the biblical scene. |
| 49. | 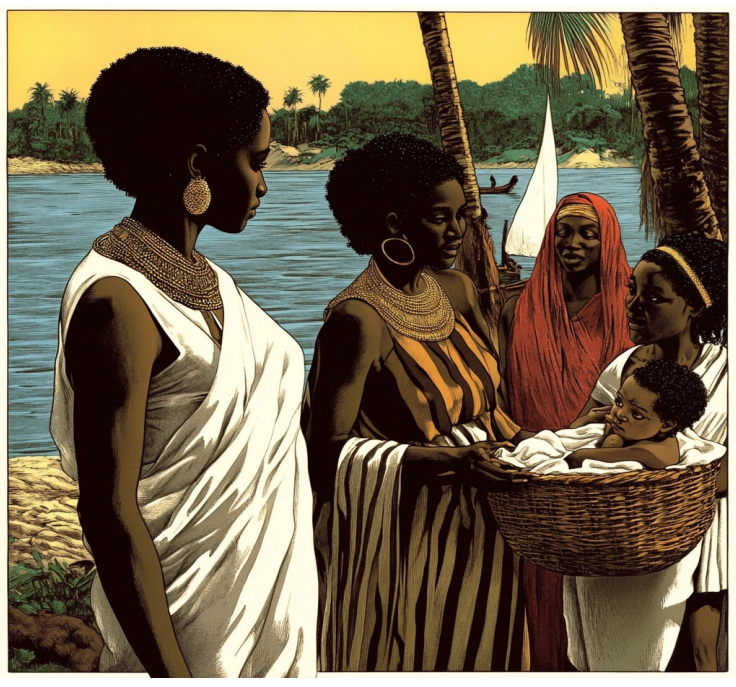 | ☑Pharaoh's Daughter Coming to the River<br>☐The Basket in the Reeds<br>☑The Maidens/Servants<br>☑The Opening of the Basket<br>☐The Child Crying<br>☑Dialogue and Interaction<br>☐ The River and Reeds as Setting | Prompt: in postcolonial African style<br>Evaluation: This image reimagines Moses' discovery in a bold, woodcut style with rich African cultural elements. Strong female figures, intricate jewelry, and a serene river scene evoke warmth and destiny. The earthy tones and traditional aesthetics offer a unique, historical perspective on the biblical narrative. |
| 50. | 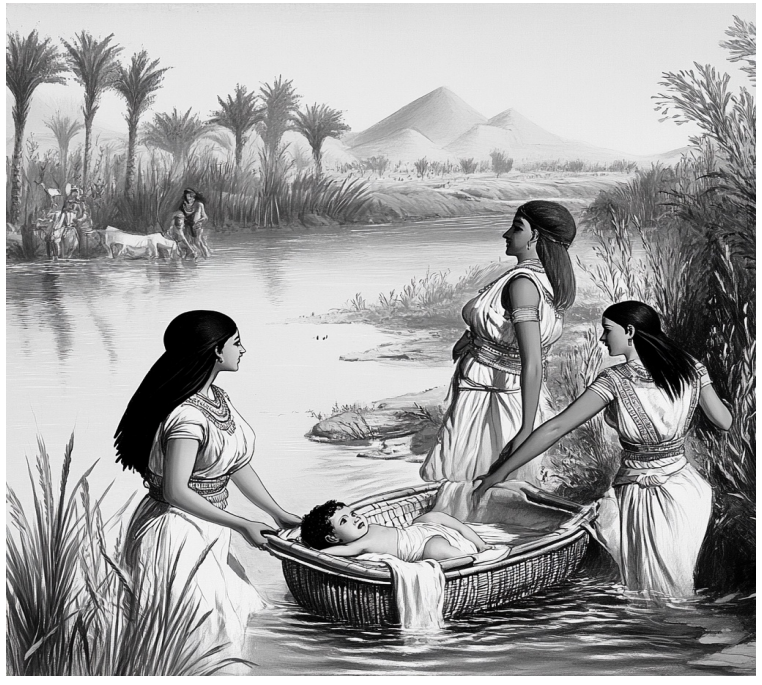 | ☑ Pharaoh's Daughter Coming to the River<br>☑ The Basket in the Reeds<br>☑ The Maidens/Servants<br>☑ The Opening of the Basket<br>☑ The Child Crying<br>☐ Dialogue and Interaction<br>☑ The River and Reeds as Setting | Prompt: In Egyptian style<br>Evaluation: This black-and-white AI-generated image evokes a historical documentary style, enhancing the timeless nature of Moses' discovery. The composition is balanced, with Egyptian figures gracefully positioned along the Nile. The soft shading and intricate details of clothing and landscape create a striking contrast, reinforcing a sense of reverence and destiny in the biblical scene. |

## Appendix 2: Random images from Google search

| | | | |
|---|---|---|---|
| 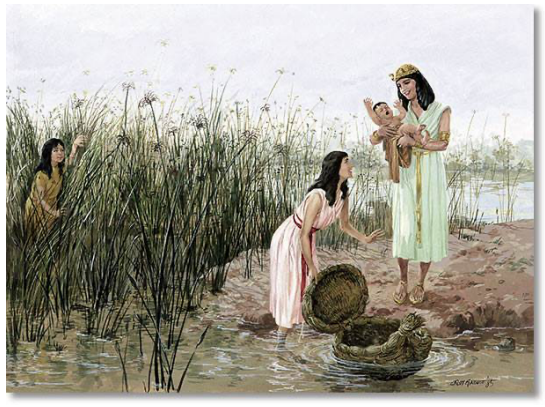 | 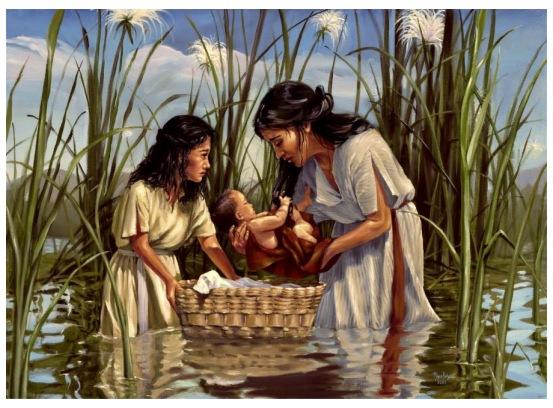 | 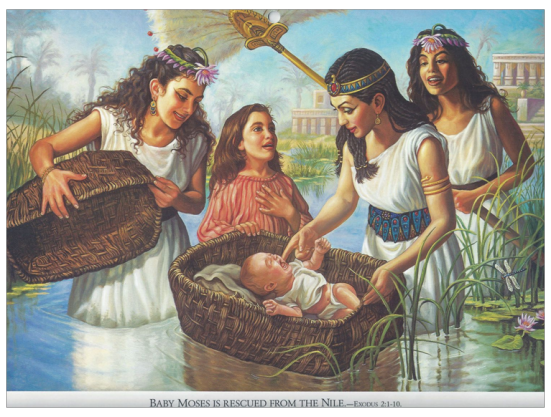 | 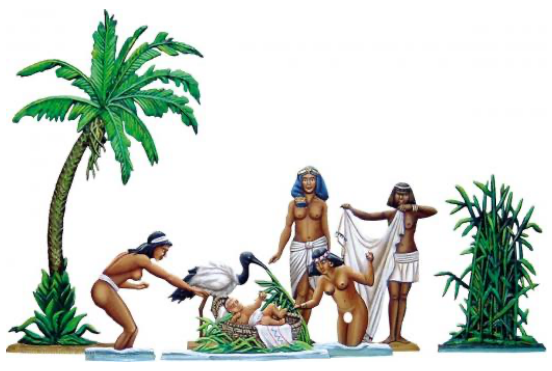 |
| https://ubdavid.org/bible/know-your-bible3/know3-6.html | https://www.womeninthescriptures.com/2015/08/the-women-who-delivered-moses.html | https://medium.com/@saharqovaizi60/the-life-of-moses-7cd131721be9 | https://www.zinnfigur.com/en/Flat-Figures/Unpainted-figures/Cultural-History/54-mm-Series/Finding-of-Moses-in-the-Nile.html |
| 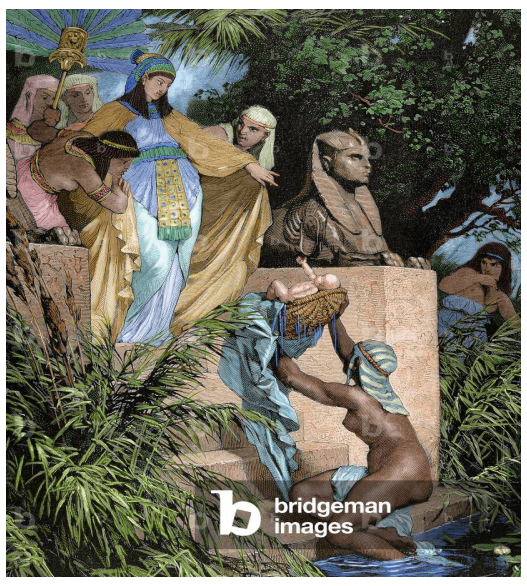  | 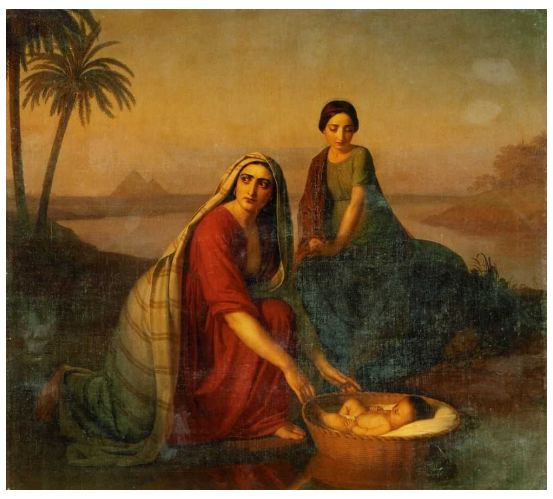 | 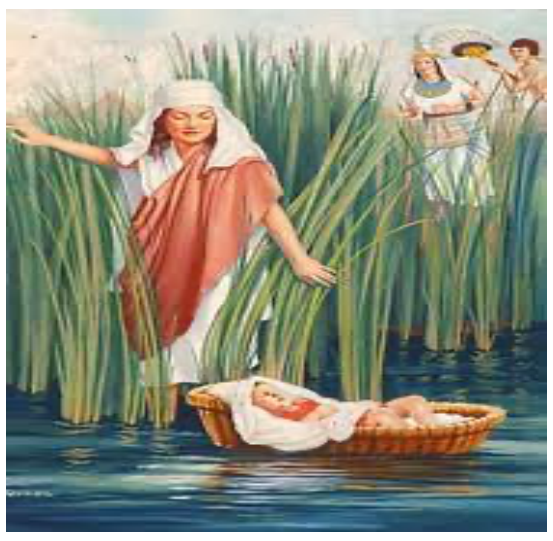 | 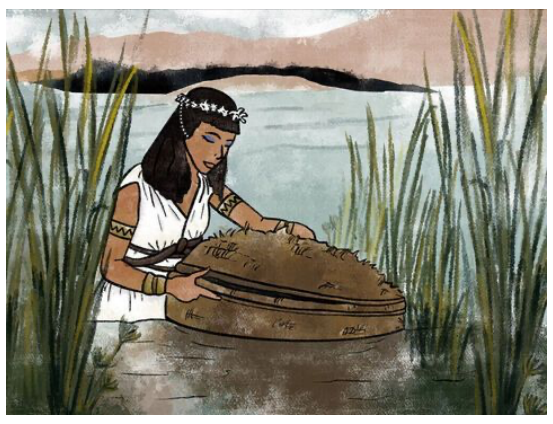 |
| https://www.bridgemanimages.com/en/noartistknown/moses-rescued-from-the-nile-by-the-daughter-of-pharaoh-of-egypt-exodus-engraving-by-a-closs-colored/nomedium/asset/5488911?offline=1 | https://signoftherose.org/2017/02/02/2527/ | https://ahefamily.org/2011/09/14/jochebeds-sunshine/ | https://pin.it/4wshnS87O |
| 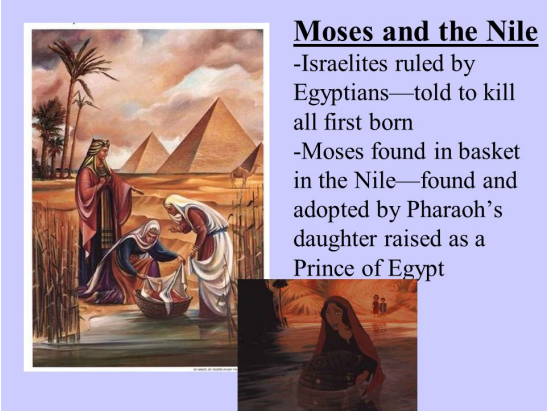  | 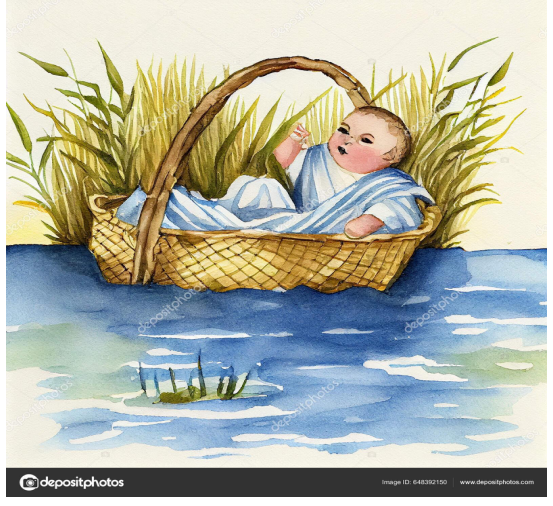  | 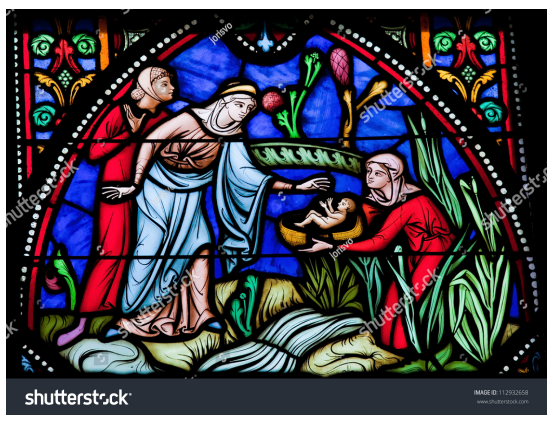  | 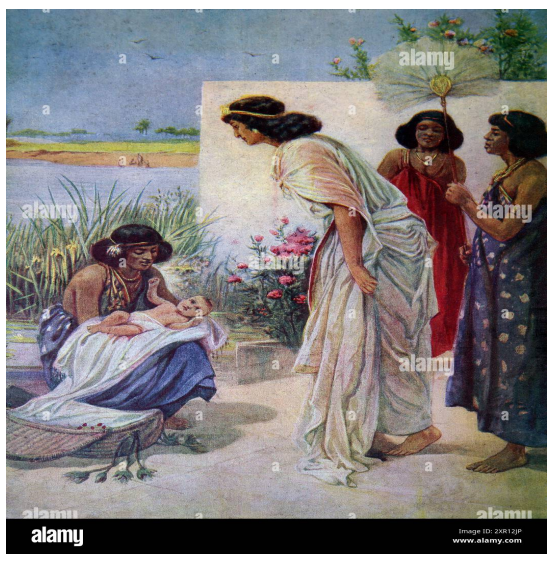  |
| https://slideplayer.com/slide/8241162/ | https://depositphotos.com/illustration/baby-moses-floating-reed-basket-river-nile-egypt-jewish-christian-648392150.html | https://www.shutterstock.com/image-photo/brussels-july-26-stained-glass-window-112932658 | https://www.alamy.com/illustration-of-of-the-finding-of-moses-by-the-pharaohs-daughter-from-antique-book-moses-bible-stories-retold-by-catharine-shaw-image616524014.html |
| 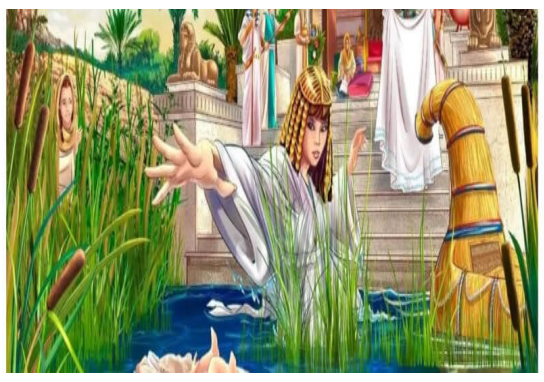 | 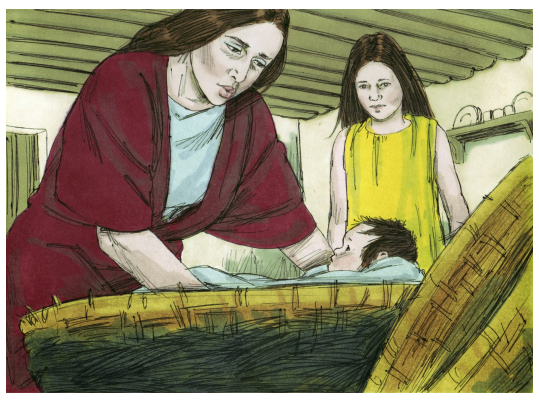 | 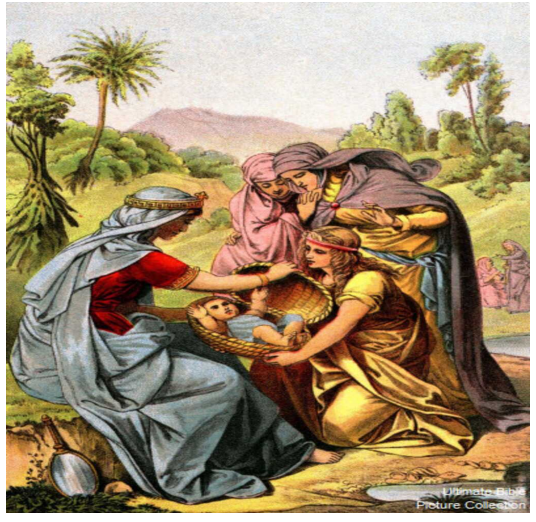 | 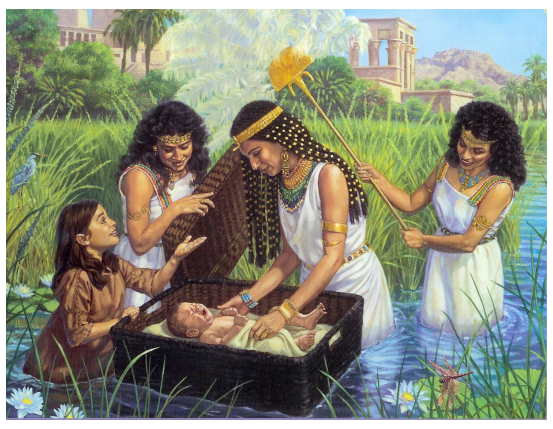 |
| https://rabbisacks.org/archive/every-child-even-a-child-of-your-enemies-is-holy/ | https://www.biblefunforkids.com/2017/05/21-moses-in-bulrushes.html | https://inspiredscripture.com/bible-studies/exodus-2 | https://theholyschmitz.wordpress.com/2015/02/21/what-we-can-learn-from-moses-part-1/ |
| 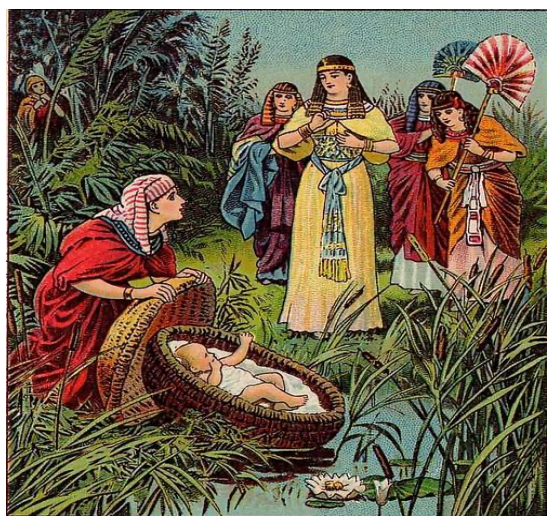 | 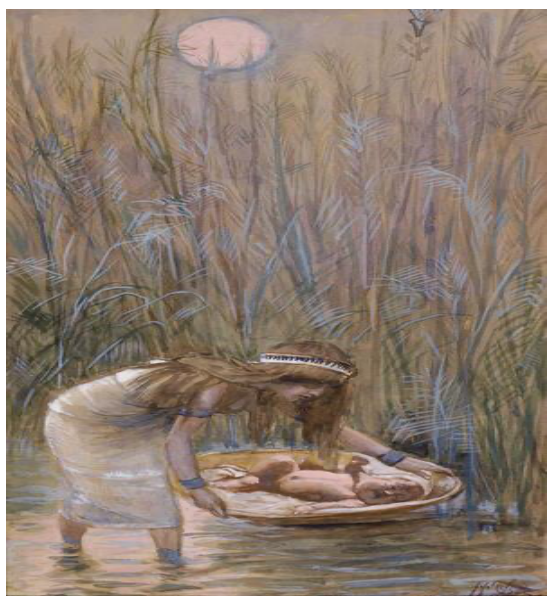 | 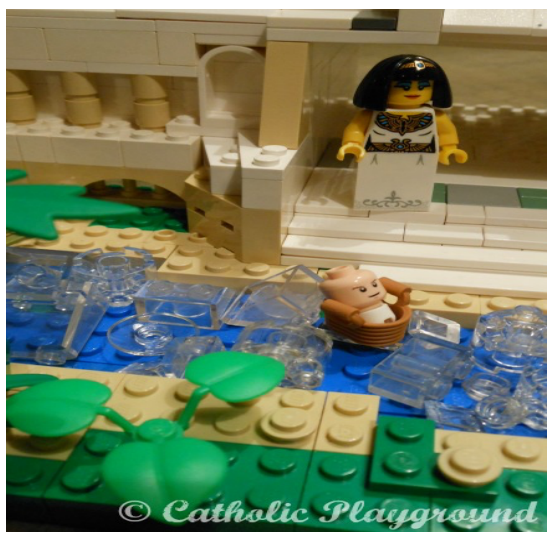 | 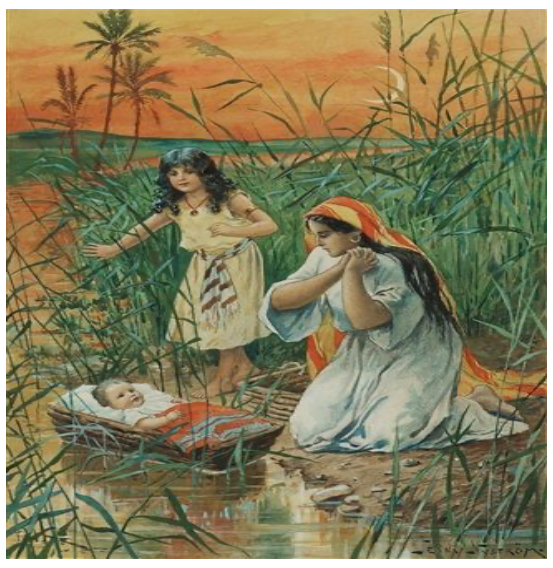 |
| https://www.worldhistory.org/Moses/ | https://jwa.org/encyclopedia/article/hebrew-women-in-egypt-midrash-and-aggadah | https://www.catholicplayground.com/baby-moses-nile-river-lego-blocks/ | https://www.pinterest.com/irisgallant0400/pharaoh-daughter-with-moses-in-the-river/ |

## Appendix 3: Abbreviated Prompts

1. The Holy Bible, King James Version (KJV). *see the table under "Objectives"*
2. The Holy Bible, New International Version (NIV). *see the table under "Objectives"*
3. The Holy Bible, New Revised Standard Version (NRSV). *see the table under "Objectives"*
4. In an Egyptian style + KJV *('ark' replaced with 'basket')* → In this case the prompt is the phrase followed by a colon and the Bible text. *(This will be repeated for prompts that appear with +).*
5. Moses found in Dali style + KJV *('ark' replaced with 'basket')*
6. Create an image in the style of Dali for the following text + NRSV
7. Create images of the following text using Salvador Dali's style of art + KJV
8. In a Lego style + KJV
9. In a Lego style + KJV ('ark' replaced with 'basket')
10. In a surrealism/symbolism style + KJV
11. In a surrealistic style + KJV *('ark' replaced with 'basket')*
12. From an African point of view + KJV
13. In a typical roman catholic style painting + KJV *('ark' replaced with 'basket')*
14. In a catholic saint art style + KJV *('ark' replaced with 'basket')*
15. In a catholic style + KJV ('ark' replaced with 'basket')
16. In a realistic style + KJV ('ark' replaced with 'basket')
17. In a post-colonial African style + KJV *('ark' replaced with 'basket')*
18. Create Michael Angelo-style images that accurately represent this passage in Exodus 2:5-9 + KJV
19. In a Michael Angelo stone art style + KJV *('ark' replaced with 'basket')*
20. In a Michael Angelo style + KJV *('ark' replaced with 'basket')*
21. Generate images accurately depicting "the finding of Moses in a karel appel style" based on this text" + NRSV
22. create realistic images to accurately represent this text + NRSV
23. Create an image in the style of Karl Appel for the following text + NRSV
24. Generate images accurately depicting "the finding of Moses in a Renaissance style" based on this text + NRSV
25. Create realistic images that accurately illustrate this biblical passage + NRSV
26. Create realistic images that accurately illustrate this biblical passage +NRSV
27. Create an image in the style of Surrealism for the following text + NRSV
28. Create images that accurately represent this passage in Exodus 2:5-9 + KJV
29. Create images of Moses being found in Exodus 2:5-9 taking into account all the following visual elements: Pharaoh's Daughter Coming to the River, The Basket in the Reeds, The Maidens/Servants, The Opening of the Basket, The Child Crying, Dialogue and Interaction, and The River and Reeds as Setting.
30. Create images of the Moses being found in Exodus 2:5-9
31. Create an image that accurately represents Moses being found by Pharoah's daughter
32. Create images of Moses being found by Pharoah's daughter
33. Create images of the *child* Moses found in Exodus 2:5-9
34. Create images of *baby* Moses found in the bible story according to Exodus 2:5-9
35. Create images of albino baby Moses being found in a basket at the Nile River by the royal daughter of the pharaoh and her maidens.

36. Create images of baby Moses being found in a basket at the Nile River by the Gen Z daughter of the pharaoh with her besties.
37. Create modern-day images of baby Moses being found in a basket at the Nile River by the Gen Z daughter of Pharaoh and her friends.
38. Depict the story of Moses found in the river Nile
39. Generate images accurately depicting "The Finding of Moses" in the River Nile from an African point of view
40. Generate realistic images of *the child* Moses found by the daughter of Pharoah in the River Nile based on the bible story Exodus 2:5-9.
41. generate realistic images of Moses being found by the daughter of Pharoah in the river Nile based on the bible story Exodus 2:5-9.

## Appendix 4: Core Parameters in Midjourney Prompts

- **--chaos**
    - Controls the unpredictability of image results.
    - Range: 0 (more consistent) to 100 (more surprising).
    - Example: --chaos 80 gives more unexpected compositions.
- **--weird**
    - Adds artistic or surreal distortion.
    - Range: 0–3000 (higher means stranger visuals).
    - Example: --weird 1000 might make a realistic scene look dreamlike.
- **--stylize (or --s)**
    - Controls how strongly MidJourney follows its artistic style.
    - Higher values = more stylization, less literal.
    - Example: --stylize 1000 makes images more decorative and abstract.
- **--seed**
    - Determines the randomness of image generation.
    - Using the same seed with the same prompt gives repeatable results.
    - Example: --seed 12345 ensures consistent output across variations.
- **--quality (or --q)**
    - Affects how much rendering time is used.
    - --quality 1 is standard, --quality 2 is higher detail (uses more credits).
    - Example: --quality 2 gives more refined results.
- **--aspect (or --ar)**
    - Sets the aspect ratio of the image.
    - Example: --ar 16:9 creates a widescreen image.
- **--no**
    - Excludes certain elements from the image.
    - Example: --no text avoids any text in the image.
- **--image-weight (or --iw)**
    - Sets how much influence an image has in a mixed prompt.
    - Higher value = image dominates over text.
- **--style**
    - Applies preset styles.
    - Example: --style raw gives more photo-realism.
- **--uplight**
    - Uses a light upscaling method (less contrast, smoother).
- **--upbeta**
    - Experimental upscale, often more detailed.